\documentclass[10pt,journal,compsoc]{IEEEtran}

\UseRawInputEncoding
\usepackage{amsmath,amsfonts}
\usepackage{algorithmic}
\usepackage{array}
\usepackage[caption=false,font=normalsize,labelfont=sf,textfont=sf]{subfig}
\usepackage{textcomp}
\usepackage{stfloats}
\usepackage{url}
\usepackage{verbatim}
\usepackage{graphicx}
\usepackage{bm}
\def\BibTeX{{\rm B\kern-.05em{\sc i\kern-.025em b}\kern-.08em
    T\kern-.1667em\lower.7ex\hbox{E}\kern-.125emX}}
\usepackage{balance}
\ifCLASSINFOpdf
\else
\fi
\usepackage{amsfonts}
\usepackage{graphicx}
\usepackage{enumerate}
\usepackage{tabu}
\usepackage{algorithm}
\usepackage{algorithmic}
\usepackage{multirow} 
\usepackage{amsmath}
\usepackage{xcolor}
\usepackage{amsmath}
\usepackage{latexsym}
\usepackage{amsmath}
\usepackage{amsthm}
\usepackage{graphicx}
\usepackage{amssymb}
\usepackage{algorithm}
\usepackage{algorithmic}
\usepackage{graphicx}
\usepackage{epstopdf}
\usepackage{float}
\usepackage{booktabs}
\usepackage{multirow}
\usepackage{indentfirst}
\usepackage{subfig}
\usepackage{longtable}
\usepackage{booktabs}
\usepackage{threeparttable}
\usepackage[pdftex,pdfnewwindow=true,linkcolor=blue,breaklinks=true, colorlinks=true,filecolor=blue,anchorcolor=blue]{hyperref}
\usepackage{url}
\usepackage{cases}

\newtheorem{theorem}{Theorem}
\newcommand{\Real}{\mathbb{R}}

\newcommand{\C}{\mathcal{C}}

\newcommand{\U}{\mathbf{U}}
\newcommand{\V}{\mathbf{V}}

\newcommand{\x}{\mathbf{x}}

\newcommand{\vv}{\mathbf{v}}

\ifCLASSOPTIONcompsoc

  \usepackage[nocompress]{cite}
\else
  \usepackage{cite}
\fi

\ifCLASSINFOpdf

\else

\fi

\begin{document}

\title{Multi-Prototypes Convex Merging Based K-Means Clustering Algorithm}

\author{Dong~Li,
        Shuisheng~Zhou,
        Tieyong~Zeng,
        and~Raymond H.~Chan.
\IEEEcompsocitemizethanks{\IEEEcompsocthanksitem D. Li, S. Zhou are with School of Mathematics and Statistics, Xidian University, Xi'an 710071, China  (E-mail: lidong$\_$xidian@foxmail.com; sszhou@mail.xidian.edu.cn).\protect
\IEEEcompsocthanksitem T. Zeng is with the Department of Mathematics, The Chinese University of Hong Kong, Shatin, Hong Kong. E-mail: zeng@math.cuhk.edu.hk
\IEEEcompsocthanksitem R.H. Chan is with the Department of Mathematics, City University of Hong Kong, 83 Tat Chee Ave, Hong Kong, and with the Hong Kong Centre for Cerebro-Cardiovascular Health Engineering, 19 W Ave, Science Park, Hong Kong. E-mail: raymond.chan@cityu.edu.hk}
\thanks{Manuscript received xxxx, 2022; revised xxxx, 2022. This work was supported by the National Natural Science Foundation of China under Grants No. 61772020; HKRGC Grants Nos. CUHK14301718, CityU11301120, and C1013-21GF; and CityU Grant 9380101 \emph{(Corresponding author: Shuisheng Zhou.)}}}

\markboth{IEEE TRANSACTIONS ON KNOWLEDGE AND DATA ENGINEERING,~Vol.~xx, No.~xx, Oct~2022}%
{Shell \MakeLowercase{\textit{et al.}}: Multi-Prototypes Convex Merging Based K-Means Clustering Algorithm}

\IEEEtitleabstractindextext{
\begin{abstract}
K-Means algorithm is a popular clustering method. However, it has two limitations: 1) it gets stuck easily in spurious local minima, and 2) the number of clusters $k$ has to be given a priori. To solve these two issues,  a multi-prototypes convex merging based K-Means clustering algorithm (MCKM) is presented. First, based on the structure of the spurious local minima of the K-Means problem, a multi-prototypes sampling (MPS) is designed to {{select}} the appropriate number of multi-prototypes for data with arbitrary shapes. A theoretical proof is given to guarantee that the multi-prototypes selected by MPS can achieve a constant factor approximation to the optimal cost of the K-Means problem. Then, a merging technique, called convex merging (CM), merges the multi-prototypes to get a better local minima without $k$ being given a priori.  Specifically, CM can obtain the optimal merging and estimate the correct $k$. By integrating these two techniques with K-Means algorithm, the proposed MCKM is an efficient and explainable clustering algorithm for escaping the undesirable local minima of K-Means problem without given $k$ first.  Experimental results performed on synthetic and real-world data sets have verified the effectiveness of the proposed algorithm.
\end{abstract}

}

\maketitle

\IEEEdisplaynontitleabstractindextext

\IEEEpeerreviewmaketitle

\ifCLASSOPTIONcompsoc
\IEEEraisesectionheading{\section{Introduction}\label{sec:introduction}}
\else

\fi

\IEEEPARstart{C}{lustering} analysis is one of the important branches in machine learning \cite{jordan2015machine, Extreme2016}, which has extensive applications in different fields, for example, artificial intelligence \cite{2003Artificial}, pattern recognition \cite{Statistical2000}, image processing \cite{Rezaee2000multiresolution}, etc. The goal of the clustering algorithm is to separate a data set into multiple clusters so that the objects in the same cluster are highly similar. Many types of clustering algorithms have been studied in the literature, see \cite{SAXENA2017} and the references therein.

As a popular clustering paradigm, partition-based methods  believe that data set can be represented by cluster prototypes. They require one to specify the number of clusters $k$ a priori and update the clusters by optimizing some objective functions. The most representative partition-based clustering algorithms is the K-Means algorithm \cite{Least1982, JAIN2010651}, which aims to divide the data set into $k$ clusters so that the sum of squared distances between each sample to its corresponding cluster center is the smallest. However, because the K-Means algorithm is NP-hard, it easily gets stuck in spurious local minima \cite{properties2018,Nie2022}. Besides, $k$ has to be given first.

To avoid bad local minima in the K-Means algorithm, numerous remedies have been proposed. Most of them can be classified into three strategies.
The first strategy focuses on initialization selection. Pena \emph{et al.} \cite{PENA19991027} concluded that the quality of the solution and running time  of the K-Means algorithm highly depends on the initialization techniques. A good initialization can find  better local minima or even global minima. K-Means++ \cite{vassilvitskii2006k} was proposed to initialize K-Means by choosing the centers with specific probabilities, and the result is $\mathcal{O}(\log k)$-competitive with the optimal result. K-Means$\mid\mid$ \cite{bahmani2012scalable} was presented to obtain a nearly optimal result by an over-sampling technique after a logarithmic number of iterations. An improved K-Means++ with local search \cite{lattanzi2019better} was developed to achieve a constant approximation guarantee to the global minima with $\mathcal{O}(k\log\log k)$ local search steps.

The second strategy focuses on theoretical innovations in the model frameworks. A relaxation method for K-Means \cite{peng2007approximating} was designed to construct the objective of K-Means into the so-called 0-1 semidefinite programming (SDP), and solve it by the linear programming and SDP relaxations. Then, a feasible solution is obtained by principal component analysis. Experimental results show that the 0-1 SDP for K-Means always find a global minima for $k=2$ (\cite{dasgupta2008hardness} also summarized similar results). Coordinate descent method for K-Means \cite{Nie2022} was provided to get better local minima by reformulating the objective of K-Means as a trace maximization problem and solving it with a coordinate descent scheme.

The third strategy focuses on adjustment to local minima based on various heuristics and empirical observations. Usually, the adjustment scheme is the splitting and merging of prototypes \cite{franti2006iterative, muhr2009automatic, lei2016robust, ismkhan2018ik, Capo2022}.

All the methods above have achieved better local minima or global minima on relatively uniform size and linearly separable data sets. This is not surprising as K-Means-type algorithms often produce clusters of relatively uniform size, even if the data sets have varied cluster sizes. This is called the "uniform effect" \cite{xiong2008}. The Euclidean distance squared error criterion of K-Means-type algorithms therefore tends to work well on relatively uniform size and linearly separable data sets. This limits the performance of the algorithms on data sets with special patterns, such as the non-uniform, non-convex and skewed-distributed data sets.

To address the aforementioned problem,  the over-parametrization learning framework \cite{dasgupta2007, buhai2020, Zhang2021}, as a promising and empirical approach, has been applied to the clustering algorithms.  In particular, multi-prototypes K-Means clustering algorithms \cite{wang2011, liang2012, wang2013,  nie2019k, Chen2021} were developed that can generate multi-prototypes that are much better suited for modeling clusters with arbitrary shape and size compared with single prototype. Then the methods iteratively merge the prototypes into a given number of clusters by some similarity measures.

However, most existing multi-prototypes methods simply use a predefined number of multi-prototypes and the selection skills lack theoretical guarantees.  Therefore, a convex clustering model was introduced in \cite{Lindsten2011} to overcome these two issues. The model is formulated as a convex optimization problem based on the over-parametrization and sum-of-norms (SON) regularization techniques. There are other variants of convex clustering models, see \cite{Zhu2014, panahi2017clustering} and the references therein. The optimization methods for solving convex clustering are generally the alternating minimization algorithm (AMA) \cite{tseng1991} and the alternating direction method of multipliers (ADMM) \cite{boyd2011}.

In convex clustering models, the number of over-parametrization is set to the number of samples, and then the samples are classified into different clusters by tuning the regularization parameter. Inevitably, its computational complexity is very high, where each iteration of the ADMM solver is of complexity $\mathcal{O}(n^2p)$. Here, $n$ is the number of samples and $p$ is the dimensionality of the samples. Recently,  a novel optimization method, called the semismooth Newton-CG augmented Lagrangian method \cite{Defeng2021}, was proposed to solve the large-scale problem for convex clustering. We emphasize that since these clustering models are convex, there are theoretical guarantee to recover their global minima.

The above approaches rarely analyzed the structures of the local minima, so there is a lack of explanation and understanding of the approaches. Recently, Qian \emph{et al.} \cite{2022Structures} investigated the structures under a probabilistic generative model and proved that there are only two types of spurious local minima of K-Means problem under a separation condition. More precisely, all spurious local minima can only be of two structures: (i) the multiple prototypes lie in a true cluster, and (ii) one prototype is put in the centroid of multiple true clusters.  In this paper, these two structures are called over-refinement and under-refinement of the true clusters, respectively. Naturally, to get better local minima or global minima, we should refine the prototypes such that one prototype lies in one true cluster. This inspires us to explore an efficient and explainable approach for finding better local minima.

Another line of research focuses on the cluster number $k$. Most methods require $k$ to be given a priori. In general, $k$  is unknown.  By adding an entropy penalty term to K-Means to adjust the bias, \emph{unsupervised} K-Means clustering algorithm (U-K-Means) \cite{Sinaga2020} can automatically find the optimal $k$  without giving any initialization and parameter selection. An over-parametrization learning procedure is established to estimate the correct $k$ in K-Means for the arbitrary shape data sets \cite{gorzalczany2017, lu2019}. We remark that the convex clustering models mentioned above, e.g., \cite{Lindsten2011}, can also find the correct $k$  by tuning the regularization parameter and the number of neighboring samples.

In this paper, we propose an efficient and explainable multi-prototypes K-Means clustering algorithm for recovering better local  minima without the cluster number $k$ being given a priori. It is called \textbf{MCKM} (\textbf{m}ulti-prototypes  \textbf{c}onvex merging based \textbf{K}-\textbf{M}eans clustering algorithm). It has two steps. The first step is guided by the aim that the final structure of the minima should have (i) at least one or more prototypes located in a true cluster, and (ii) no prototypes are at the centroid of multiple true clusters. Along this line, an efficient over-parametrization selection technique, called multi-prototypes sampling (MPS), is put forward to {{select}} the appropriate number of multi-prototypes. A theoretical guarantee of the optimality is given. Then in the second step, a  merging technique, called convex merging (CM), is developed to get better local  minima without $k$ being given.

The main contributions of this paper are as follows:
\begin{enumerate}
  \item  An appropriate number of multi-prototypes can be selected by MPS to adapt to data with arbitrary shapes. More importantly, we prove that the multi-prototypes so selected can achieve a constant factor approximation to the global minima.
  \item  CM  can get better local  minima without $k$ being given.  It obtains the optimal merging and estimates the correct $k$, because it treats the merging task as a convex optimization problem.
  \item The combined method \textbf{MCKM} is an efficient and explainable K-Means algorithm that can escape the undesirable local minima without given $k$.
    \item Experiments on synthetic and real-world data sets illustrate that MCKM outperforms the state-of-the-art algorithms in approximating the global minima of K-Means, and accurately evaluates the correct $k$.  In addition, MCKM excels in computational time.
\end{enumerate}

The paper is organized as follows. Section \ref{sec2} reviews the related works. The research motivation is described in Section \ref{sec3} and the new algorithm is presented in Section \ref{sec4}. The experimental results with discussion are reported in Section \ref{sec5} and Section \ref{sec6} concludes the paper.

\emph{Notations}: Let a data set be $\mathbf{X}=\{\x_1,\x_2, \cdots, \x_n\}$ with sample $\x_{j}\in\Real^p$, and the cluster centers be $\V=[\vv_1, \vv_2, \cdots, \vv_k]$, where $\vv_{i}\in\Real^{p}$ is the prototype of the cluster $\C_{i}$ for $i=1,2,...,k$. Denote $\|\cdot\|$ the vector $2$-norm or the Frobenius norm of a matrix.  The distances between $\x_{j}$ and the prototypes $\V$ are $d_{ij}=\|\x_{j}-\vv_{i}\| (i=1,\cdots,k)$ and the closest distance is denoted as $D(\x_{j})$. The membership grade matrix is denoted by $\U=[u_{ij}]\in\Real^{k\times n}$, where $u_{ij}$ represents the grade of the $j$th sample belonging to the $i$th cluster. The optimal cost of K-Means on data set $\mathbf{X}$  is denoted by ${J_{\mathbf{X}}}^{opt}$ and the corresponding optimal clusters are denoted by $\C^{opt}$.

\section{Related Work}\label{sec2}
In this section, some improved K-Means-type clustering algorithms and the clustering algorithms based on  over-parametrization learning are briefly recalled.

First of all, the K-Means problem aims to find the $k$ partitions of $\mathbf{X}$ by minimizing the sum of squared distances between each sample to its nearest center. The underlying objective function is expressed as follows:
\begin{equation}\label{eq_1}
\begin{aligned}
\min_{\U,\V} J_{\mathbf{X}}(\U,\V)=&\sum_{i=1}^k\sum_{j=1}^n u_{ij}\|\x_{j}-\vv_{i}\|^2,\\
    s.t. \quad &\sum_{i=1}^k u_{ij}=1, u_{ij} \in \{ 0 , 1\}.
    \end{aligned}
\end{equation}
To solve problem \eqref{eq_1}, iterative optimization algorithms are usually employed to approximate the global minima of the K-Means problem \cite{Clustering1987}. Among these algorithms, the most commonly used is the K-Means algorithm in \cite{Least1982}.

\subsection{K-Means and K-Means++ Algorithms}\label{subsec2-2}
The K-Means algorithm \cite{Least1982}, as the most popular clustering algorithm, is a heuristic method. First, $k$ initial cluster centers are set as initializations, and then an iterative algorithm, called Lloyd's algorithm, is implemented.  For an input of $n$ samples and $k$ initial cluster centers, Lloyd's algorithm consists of two steps:  the assignment step assigns each sample to its closest cluster:\\
\begin{align}
u^{(t+1)}_{ij}&=\left\{
\begin{array}{ll}
  1, & {d^{(t)}_{ij}}=\min_{1\leq c \leq k} {d^{(t)}_{cj}}\\
  0, & \textrm{otherwise}, \\
\end{array}
\right.\label{eq_2}
\end{align}
where $t$ is the iteration number.
Then, the update step replaces the $k$ cluster centers with the centroid of the samples assigned to the corresponding clusters:\\
\begin{align}
\vv^{(t+1)}_{i}&=\frac{\sum\limits_{j=1}^{n}\left(u^{(t+1)}_{ij}\right)\x_{j}}
{\sum\limits_{j=1}^{n}\left(u^{(t+1)}_{ij}\right)}.\label{eq_3}
\end{align}
The algorithm alternately repeats the two steps until convergence is achieved. The K-Means algorithm easily gets stuck in spurious local minima because of the non-convexity and non-differentiability of \eqref{eq_1}.

As studied in \cite{PENA19991027}, a good initialization makes Lloyd's algorithm perform well. Therefore, K-Means++ algorithm \cite{vassilvitskii2006k} was proposed as a specific way of choosing the prototypes $\V^{(0)}$ for K-Means. In the first step, K-Means++ selects  an initial prototype $\vv_{1}$ uniformly at random from the data set. In the second step, each subsequent initial centroid $\vv_{i}, i=2, 3,...,k$, is chosen by maximizing the following probability with respect to the previously selected set of prototypes:
\begin{align}
\frac{D(\x_{j})^2}{\sum_{\x \in \mathbf{X}} D(\x)^2}, \quad j=1, 2,..., n.
\label{eq_4}
\end{align}
Then, the algorithm repeats the second step until it has chosen a total of $k$ prototypes. The sampling skill used in the second step is called "$D^{2}$ sampling". We note that it achieves approximation guarantees, as stated in the following \textbf{Theorem} \ref{theorem1}.
\begin{theorem}\textrm{\cite{vassilvitskii2006k}}
For any data set $\mathbf{X}$, if the prototypes are constructed with K-Means++, then the corresponding objective function $J_{\mathbf{X}}$ satisfies $E[J_{\mathbf{X}}] \leq 8(\ln k+2) {J_{\mathbf{X}}}^{opt}$.
\label{theorem1}
\end{theorem}

Thus, K-Means++ algorithm is fast, simple, and $\mathcal{O}(\log k)$-competitive with the optimal result.

\subsection{Split-merge K-Means Algorithm}\label{subsec2-2-2}
Split-merge K-Means Algorithm (SMKM) \cite{Capo2022} was introduced to reduce the cost of K-Means problem \eqref{eq_1} by a new splitting-merging step, which is able to generate better approximations of the optimal cost of the K-Means problem. It consists of the following two steps:
\begin{itemize}
\item Splitting step: 2-Means is applied to each cluster $\C_{i}$ to get $\{\vv_{i_{1}}, \vv_{i_{2}}\}$, and then $\C_{i_{\textrm{split}}}$ is selected as the split cluster based on the following:
\begin{align}
i_{\textrm{split}}=\mathop{\arg \max}_{i \in \{1,2,...,k\} } [J_{\C_{i}}(\U,\vv_{i})-J_{\C_{i}}(\U, \{\vv_{i_{1}}, \vv_{i_{2}}\})].
\label{eq_5}
\end{align}
\item Merging step: the pair of clusters with the smallest merging error increment can be merged together. Specifically, $\C_{i}$ and $\C_{c}$ are merged if
\begin{equation}
\begin{aligned}
i, c=\mathop{\arg \min}_{i, c \in \{1,2,...,k+1\}, i \neq c}  f_{i, c}
\label{eq_6}
\end{aligned}
\end{equation}
where $\C_{i,c}=\C_{i} \cup \C_{c}$, $\vv_{i, c}=\frac{\lvert \C_{i} \rvert \cdot \vv_{i} + \lvert \C_{c} \rvert \cdot \vv_{c}}{\lvert \C_{i} \rvert +\lvert \C_{c} \rvert}$, and \\
$f_{i, c}=J_{\C_{i, c} }(\U,\vv_{i, c})-[J_{\C_{i}}(\U, \vv_{i})+J_{\C_{c}}(\U, \vv_{c})]$.
\end{itemize}
SMKM repeats alternately  the splitting and merging steps until convergence is achieved. In conclusion, the splitting step reduces the K-Means approximation error, while the merging step increases it. Hence, the quality of the local minima can be improved when the splitting-merging step reduces the cost of the K-Means.

\subsection{K-Multiple-Means Algorithm}\label{subsec2-4}
K-Multiple-Means \cite{nie2019k}, as an extension of K-Means, divides the samples into a  predefined $m > k$ sub-clusters, and get  the specified $k$ clusters for multi-means datasets. Its objective function is expressed as follows:
\begin{equation}\label{eq_8}
\begin{aligned}
\min_{\U,\V} {J}_{\mathbf{X}}(\U,\V)=&\sum_{i=1}^m\sum_{j=1}^n w_{ij}\|\x_{j}-\vv_{i}\|^2+\gamma \|W\|^2_{F},\\
    s.t. \quad &\sum_{i=1}^m w_{ij}=1, w_{ij} \geq 0,
    \end{aligned}
\end{equation}
where $W=[w_{ij}], i=1,2,...,m, j=1,2,...,n$, is the probability matrix. The regularization parameter $\gamma$ is used to control the sparsity of the connection of the samples to the multi-prototypes.

Using an alternating optimization strategy to solve \eqref{eq_8}, the partition of the data set is obtained with  $m$ prototypes. Then, $k$ clusters are achieved based on the partition and the connectivity of the bipartite graph.  K-Multiple-Means is a clustering algorithm based on {{over-parametrization}} learning.  Therefore, it is another efficient method for non-convex data. However, its performance  highly depends on the predefined $m$, and $k$ needs to be known in advance.

\subsection{Convex Clustering Algorithm}\label{subsec2-3}
Convex clustering model \cite{Lindsten2011} formulates the clustering task as a convex optimization problem by adding a sum-of-norms (SON) regularization to control the trade-off between the model  error and the number of clusters. To reduce the computational burden of evaluating the regularization terms, the weight, $W=[w_{ij}]$, is introduced. The objective function of convex clustering model is expressed as follows:
\begin{equation}\label{eq_7}
\begin{aligned}
\min_{\bm{\mu}_{1},..., \bm{\mu}_{n} \in \Real^{p}} \frac{1}{2} \sum_{j=1}^n \|\bm{\mu}_{j}-\x_{j}\|^2+\gamma \sum_{i < j} w_{ij}\|\bm{\mu}_{i}-\bm{\mu}_{j}\|_{p},\\
   \end{aligned}
\end{equation}
where $\gamma > 0$ is a tuning parameter and the $p$-norm with $p \geq 1$ ensures the convexity of the model.  Here, $w_{ij}$ is a nonnegative weight given by:\\
\begin{equation}\label{eq 16}
w_{ij}=\left\{
\begin{array}{ll}
  \textrm{exp}(-\kappa\|\x_{i}-\x_{j}\|{{^2}}), &\textrm{if} \quad (i,j) \in E;\\
  0, & \textrm{otherwise},
\end{array}
\right.
\end{equation}
where $E=\cup_{j=1}^{n}\{l=(i,j)| i \in \textrm{KNN}(j) \}$, $\textrm{KNN}(j)$ is the index set of the $q$-nearest neighbors of $\x_{j}$ for $j=1,2,...,n$, and $\kappa$ is a given positive constant.

After the optimal solutions $\bm{\mu}^{*}_{1},..., \bm{\mu}^{*}_{n}$ of \eqref{eq_7} are obtained, the samples are assigned to be in one cluster if and only if their  optimal solutions $\bm{\mu}^{*}$ are the same. Convex clustering, based on over-parametrization learning,  can avoid bad local minima, cluster arbitrary shape data sets, and get the cluster number \cite{Zhu2014, panahi2017clustering}.  However, its computational complexity is very high, which still remains challenging for  large-scale problems. {{Meanwhile, the number of neighboring samples $q$ is generally selected empirically. If it is too small,  convex clustering will not achieve the perfect recovery. Conversely, the computational burden cannot be reduced.}}

\section{Motivation}\label{sec3}
The above-mentioned algorithms can avoid the bad local minima of K-Means problem. However, their structure is rarely involved in the studies,  so that the recovery approaches are not well understood. For example, in \cite{Capo2022}, the splitting and merging steps are performed for selecting  better local minima. But only a prototype is split by 2-Means, and only a pair of prototypes are merged in each iteration. This lack of explanation of the structure of better local minima inspires us to come up with a new algorithm.

\subsection{Recover the Better Local Minima Based on Multi-Prototypes Technique}\label{subsec3-1}

In this subsection, an important theorem in \cite{2022Structures}, which describes the structure of spurious local minima of the K-Means problem under convex data,  is recalled for convenience.
\begin{theorem}
For well-separated mixture models, all spurious local minima solutions $\V=[\vv_{1}, \vv_{1},..., \vv_{k}]$ of $\mathbf{X}$ involves the following configurations: (i) multiple prototypes $\{\vv_{i}\}$ lie in a true cluster and
(ii) one prototype $\vv_{i}$  is put in the centroid of multiple true clusters.
\label{theorem2}
\end{theorem}

Note that the above configurations (i) and (ii) are referred to as the over-refinement and under-refinement of the true clusters, respectively. Importantly, \textbf{Theorem} \ref{theorem2}  gives the  general splitting-merging approaches an explanation and understanding for the better local minima. In detail, the splitting and merging steps remove  under-refinement and over-refinement of the true clusters, respectively.

Borrowing the precise characterization of local minima in \textbf{Theorem} \ref{theorem2}, the splitting technique is theoretically a good way to remove  under-refinement. However, it does not determine exactly how many prototypes to split into, so as to eliminate under-refinement in the clustering result. Therefore, in this paper, a multi-prototypes technique, as an over-parametrization approach, is exploited instead of implementing an unsatisfactory splitting step. The multi-prototypes technique can avoid under-refinement. In detail, by setting a number larger than $k$ as the number of clusters, the multi-prototypes technique aims to achieve that at least one prototype or multiple prototypes lie in a true cluster. Then when coupled with the K-Means algorithm, one expects that only exact refinement or over-refinement of the true clusters exist after the K-Means algorithm. Hence, a better local minimum can be recovered by simply merging the prototypes in this structures.  In conclusion, the multi-prototypes technique is a suitable alternative to the splitting step for removing under-refinement of the true clusters in the clustering result.

\subsection{{{Select}} the Appropriate Number of Multi-Prototypes}\label{subsec3-2}
\begin{figure}[t]
 \centering
 \subfloat[]{\label{fig2a}\includegraphics[width=0.15\textwidth]{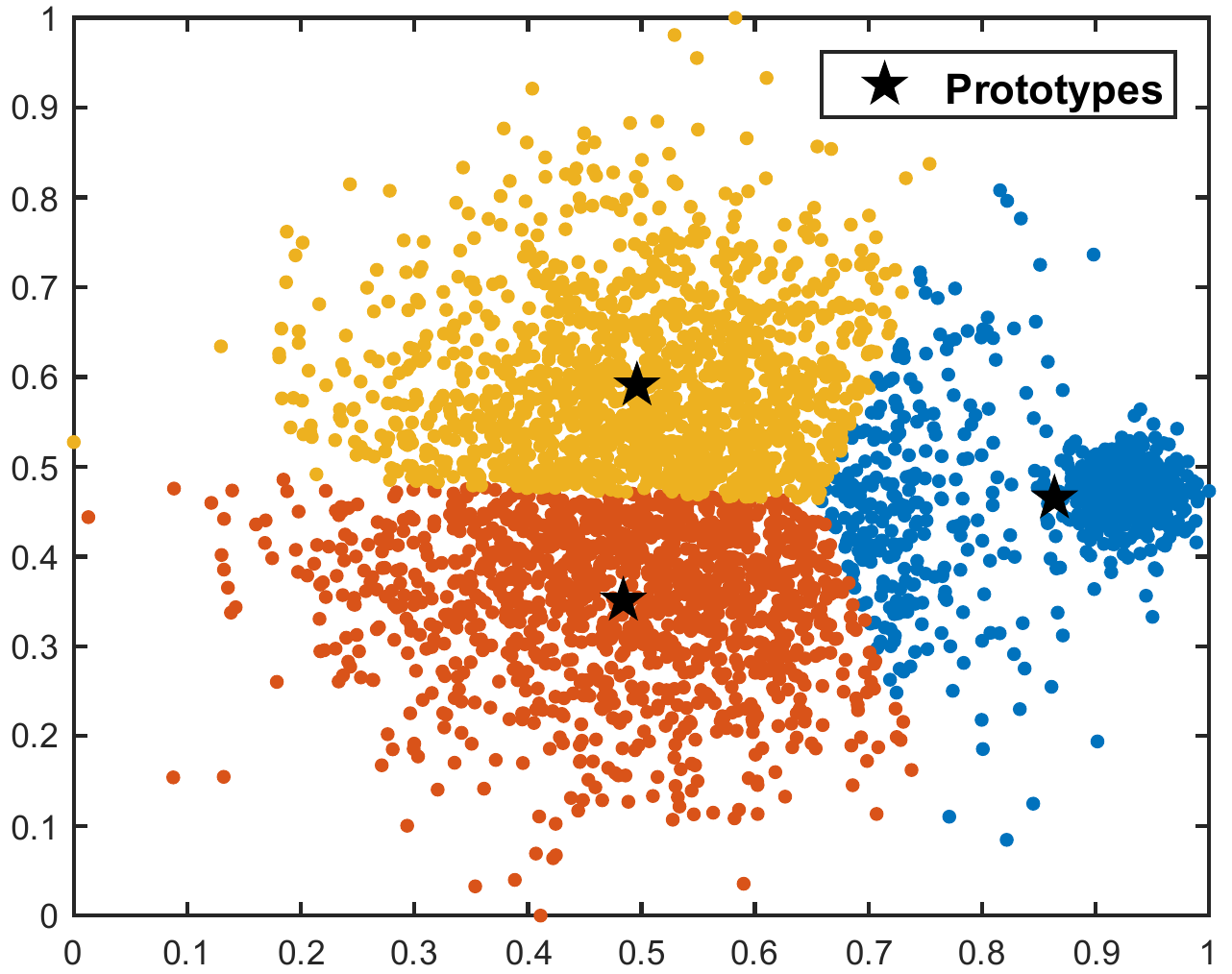}}~
 \subfloat[]{\label{fig2b}\includegraphics[width=0.15\textwidth]{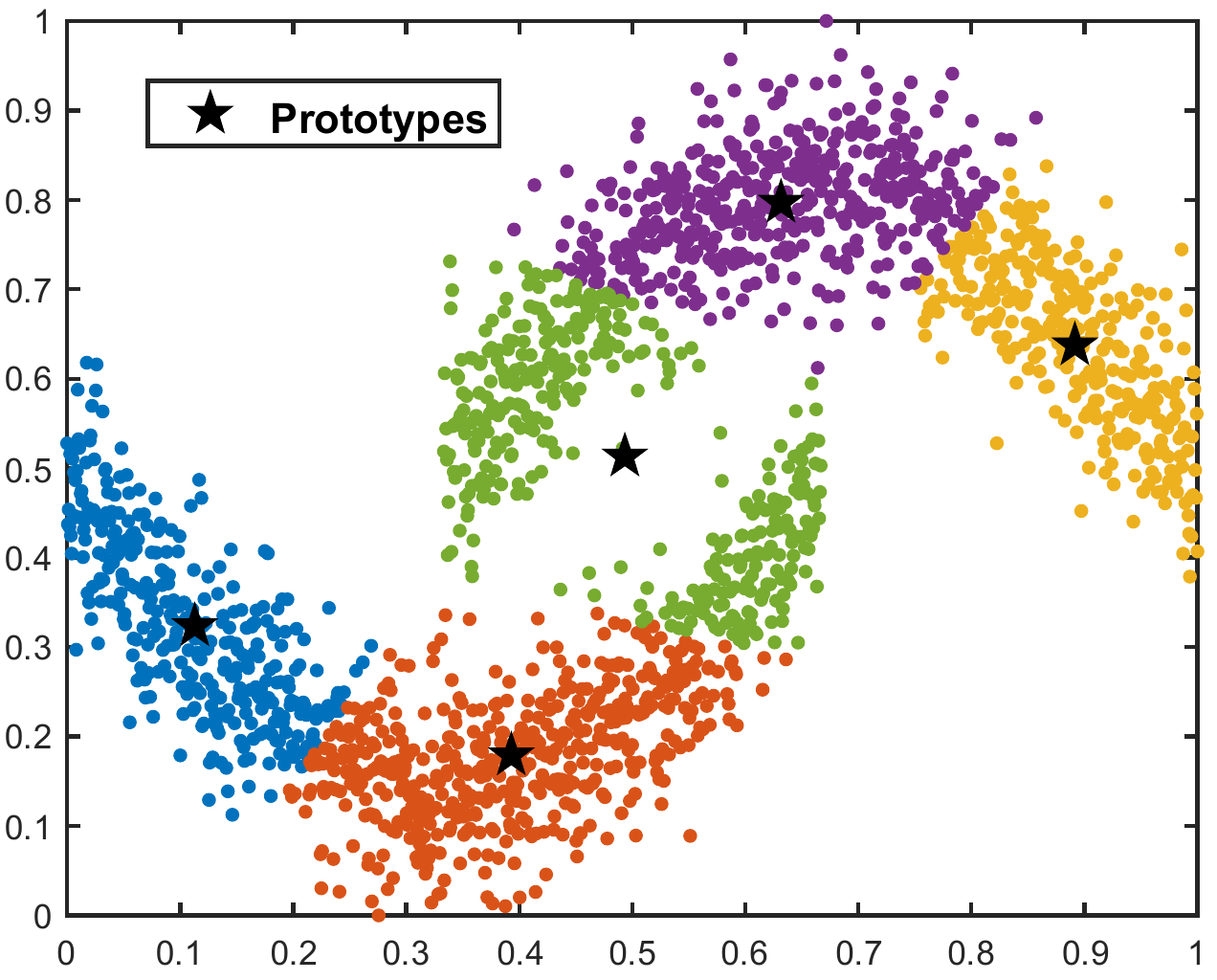}}~
 \subfloat[]{\label{fig2c}\includegraphics[width=0.15\textwidth]{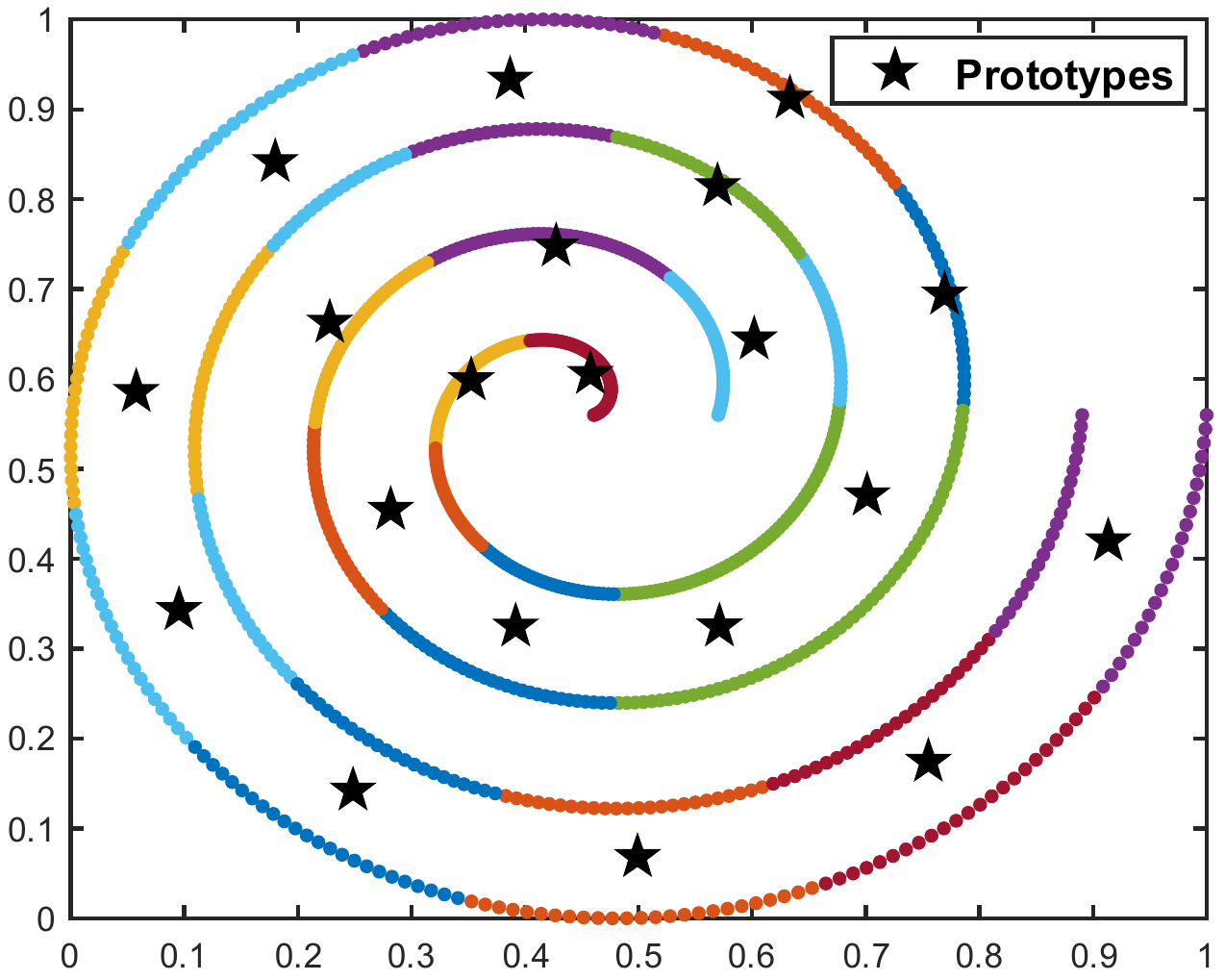}}\\
 \subfloat[]{\label{fig2d}\includegraphics[width=0.15\textwidth]{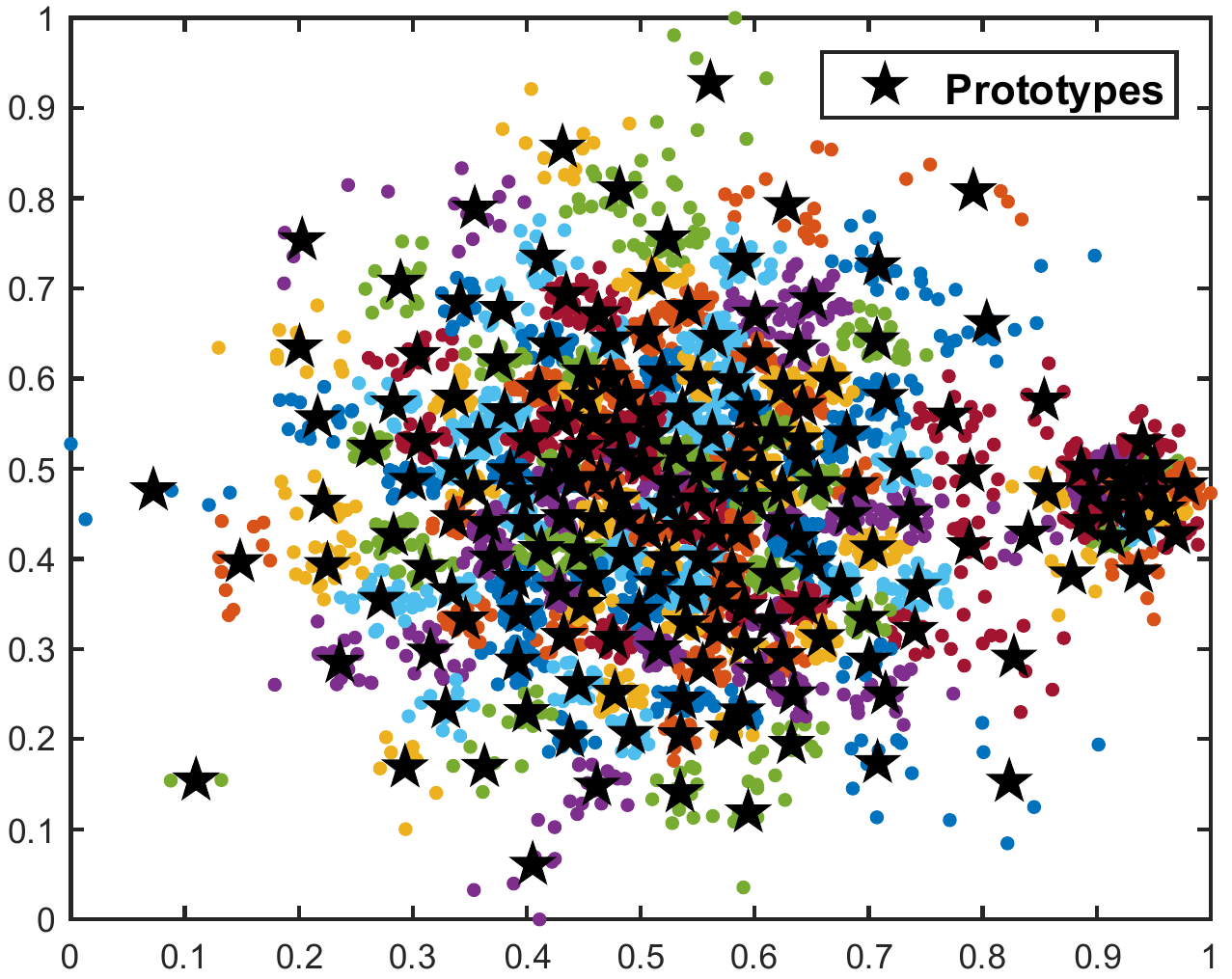}}~
 \subfloat[]{\label{fig2e}\includegraphics[width=0.15\textwidth]{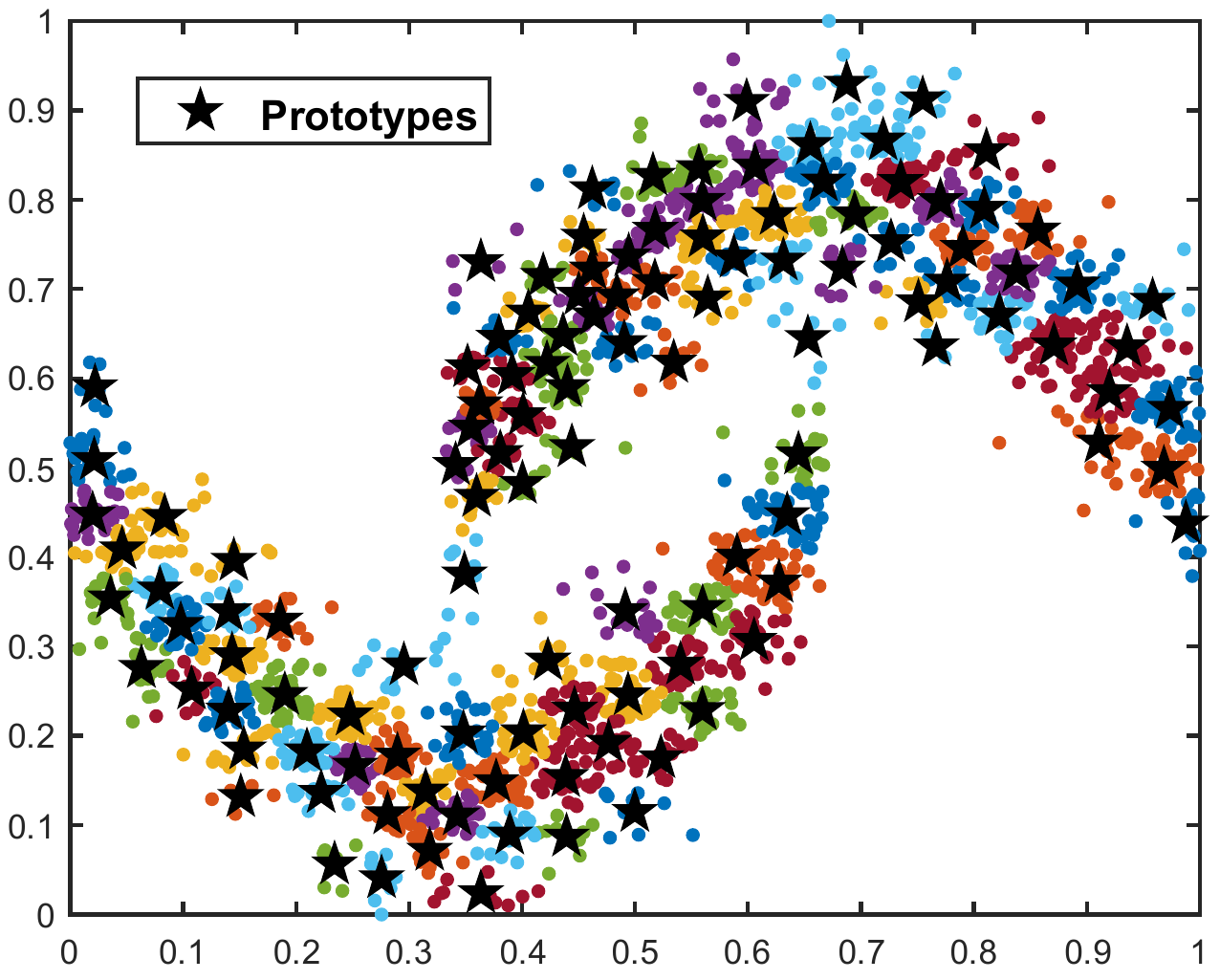}}~
 \subfloat[]{\label{fig2f}\includegraphics[width=0.15\textwidth]{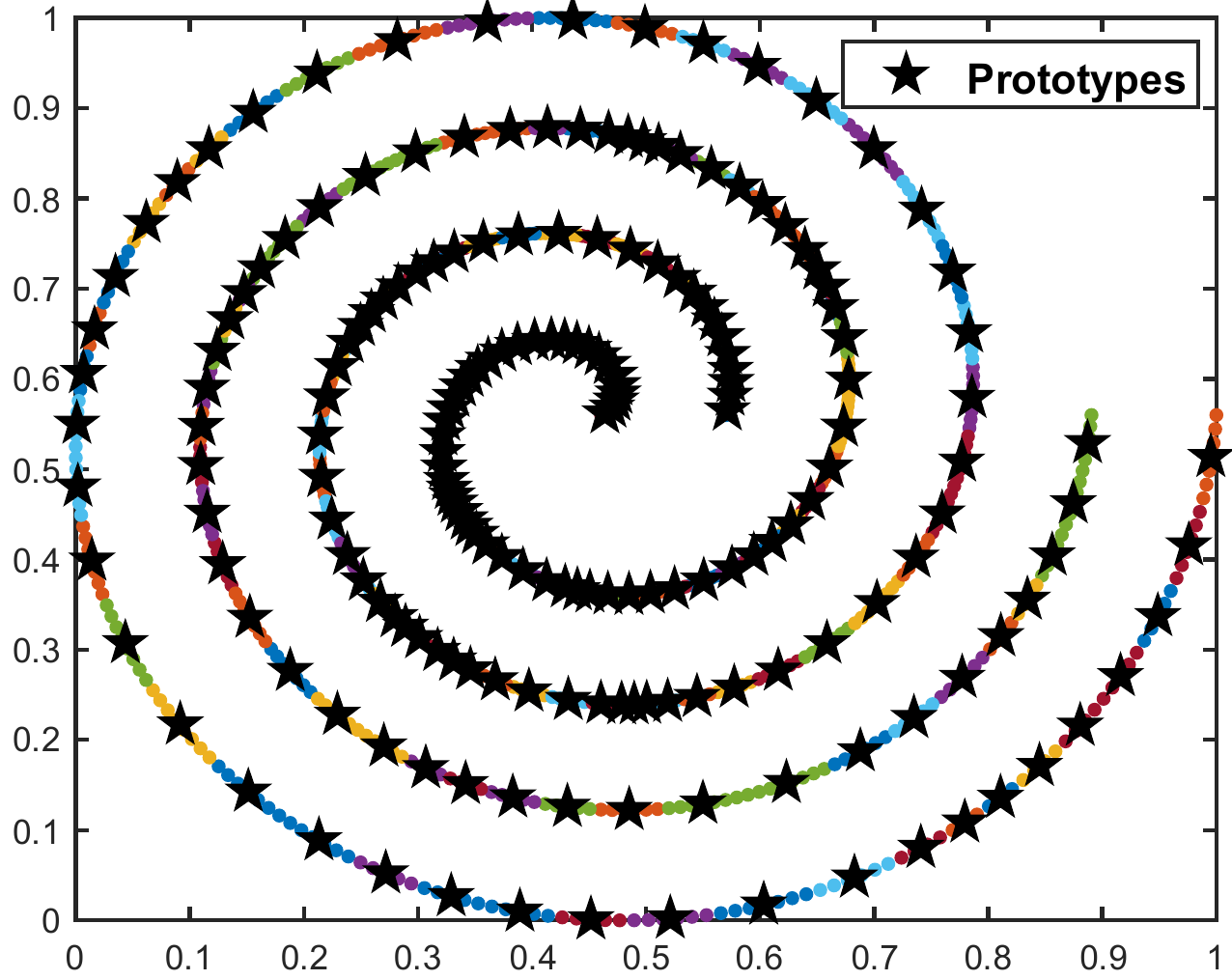}}
 \caption{Clustering results of K-Means with the multi-prototypes on three synthetic data sets.  In (a)--(c), the given number of multi-prototypes is too small; in (d)--(f), the given number of multi-prototypes is too large. The plots clearly show some final prototypes always lie in the overlapping area between the different true clusters, noises samples and outliers when the number of multi-prototypes is not selected properly.}\label{fig2}
\end{figure}

To  eliminate under-refinement of the true clusters in the clustering result, we need  to set a number of multi-prototypes larger than $k$. However, a predefined number of multi-prototypes may not work for different data sets. In order to illustrate these issues, a set of experiments is carried out on three synthetic data sets, as shown in  Fig. \ref{fig2}. See Section \ref{Synthetic} for the details of the data sets.

In Fig. \ref{fig2}, the clustering results of K-Means with different predefined numbers of  multi-prototypes on the three synthetic data sets are displayed, where the true number of clusters for the three data sets is $k=2$. In (a)--(c), the number of multi-prototypes is set to be small; in (d)--(f), the number of multi-prototypes is set to be large.

From the illustration, it can be summarized that if the given number of prototypes is too small,  the multi-prototypes for the different data distributions are inaccurate and some prototypes may lie in the centroid of multiple true clusters.  Conversely, if the  number  is too large, some prototypes  lie in the  overlapping area between the true clusters, noises samples and outliers. Clearly, the larger the given number of prototypes,  the better the representation of the multi-prototypes for the different data distributions. However, if the number of multi-prototypes exceeds a certain number,  the representation approximates density clustering and the computational complexity is too high. An appropriate number of multi-prototypes should be closely related to the data distribution. Therefore, we design a multi-prototypes selection technique to {{sample}} an appropriate number of multi-prototypes based on the data distribution, where the samples are gradually selected as  prototypes by $D^{2}$ sampling until the latest selected sample has little improvement in the data  representation. The details are presented in Section \ref{sec4}.

\section{Multi-Prototypes Convex Merging Based K-Means Clustering Algorithm} \label{sec4}
In this section, the multi-prototypes convex merging based K-Means clustering algorithm (MCKM) is proposed to recover better local minima without given the cluster number $k$ a priori. In the first step of MCKM,  a multi-prototypes sampling (MPS) first selects a suitable number of multi-prototypes for better data representation. It provides an explainable approach to refine or over-refine clusters based on the structure of the local minima.  Furthermore, a theoretical proof is given, which  guarantees that the multi-prototypes selected by MPS can achieve a constant factor approximation to the global minima of K-Means problem. Then in the second step, a merging technique, convex merging (CM), recovers the better local minima. CM can obtain the optimal merging and estimate the correct cluster number because it treats the merging task as a convex optimization problem. The overall process of MCKM is as follows:
\begin{equation*}
    \mathbf{X} \xrightarrow{\textrm{MPS (Alg. \ref{alg1})}} \{ \V_{\textrm{MPS}}, \C^{\textrm{MPS}}\}\xrightarrow{\textrm{CM (Alg. \ref{alg3})}} \C^{\textrm{MCKM}}.
  \end{equation*}
MPS and CM are explained in Subsections \ref{sec4-1} and \ref{sec4-2}, respectively.

\subsection{Multi-Prototypes Sampling (MPS)}\label{sec4-1}
In the K-Means algorithm, the multi-prototypes are constructed to represent the data structure.  To quantify the  representation ability of the multi-prototypes, a reconstruction criterion is introduced, see \cite{Enhancement2017, 2021Granular}:
\begin{equation}\label{eq_9}
R(s)=\sum_{j=1}^{n}\|\x_{j}-\hat{\x}_{j}(s)\|^2,
\end{equation}
where  $\hat{\x}_{j}(s)=\sum_{i=1}^{s} u_{ij} \vv_{i}/\sum_{i=1}^{s} u_{ij}$. It gives the reconstructed value with the current prototypes $\left\{ \vv_{1}, \vv_{2},..., \vv_{s} \right\}$ and assignment coefficients, $u_{ij}$, obtained by \eqref{eq_2}. Note that the lower value of the reconstruction criterion, the better the representation ability of the multi-prototypes.  Meanwhile, it can be inferred that the value of the reconstruction criterion decreases with the increasing number of multi-prototypes. However, it is not expected that the number of  multi-prototypes is very large, as analyzed in Subsection \ref{subsec3-2}. Therefore, a new ratio, called relative reconstruction rate with respect to the number of multi-prototype,  is defined as follows:
\begin{equation}\label{eq_10}
\frac{R(s-1)-R(s)}{R(s-1)}.
\end{equation}

The relative reconstruction rate can be utilized as a measure of the improvement of the representation ability of the new multi-prototypes set after adding a prototype to the multi-prototypes set.  If the new multi-prototypes set has little improvement in  the relative reconstruction rate after adding a prototype, the new prototype should not be added.  Hence, the number of multi-prototypes can be selected based on \eqref{eq_10}, where the quantization of little improvement is  equivalent to $\frac{R(s-1)-R(s)}{R(s-1)} \leq \varepsilon$ by setting a small threshold $\varepsilon$.

As the analysis in Subsection \ref{subsec3-2} shows, a predefined number of multi-prototypes is difficult to be set properly, and MPS can select a suitable number using the relative reconstruction rate. MPS randomly picks an initial sample into cluster $\C$, and then proceeds $D^2$ sampling, where the sample is selected with the probability \eqref{eq_4} and added to $\C$ in each iteration. MPS converges until the new multi-prototypes set has little improvement in representing the data set after adding the latest selected sample. Finally, the K-Means algorithm is performed with the selected prototypes on the data set, and the final result is obtained. The proposed MPS is presented  in Algorithm \ref{alg1}.

\begin{algorithm}[htb]
	\caption{Multi-Prototypes Sampling (MPS)}
	\label{alg1}
	\begin{algorithmic}[1]
        \REQUIRE Date set $\mathbf{X}=\{\x_1,\x_2, \cdots, \x_n\}$, the threshold $\varepsilon$;
		\ENSURE The multi-prototypes $\V_{\textrm{MPS}}$, the number of the multi-prototypes $s^{*}$.
        \STATE Pick a sample $\x^{(1)}$ randomly and $\V=\{\x^{(1)}\}$;
        \STATE Compute $R(1)$ based on $\V$, Eq. \eqref{eq_2} and Eq. \eqref{eq_9};
        \STATE Set $s:=2$ and $R=R(1)$;
        \WHILE{$s \leq n$}
            \STATE Sample $\x^{(s)}$ with probability $\frac{D(\x^{(s)})^2}{\sum_{\x \in \mathbf{X}} D(\x)^2}$ based on the current $\V$ and add it to $\V$;
            \STATE Compute $R(s)$ based on $\V$, Eq. \eqref{eq_2} and Eq. \eqref{eq_9};
            \IF {$\frac{R-R(s)}{R} \leq \varepsilon$}
            \STATE Break;
            \ELSE
            \STATE {$R=R(s)$};
            \STATE {$s=s+1$};
            \ENDIF
		\ENDWHILE
		\STATE Run K-Means with the selected prototypes $\V$ on $\mathbf{X}$;
		\STATE Obtain the multi-prototypes $\V_{\textrm{MPS}}$, the corresponding clusters $\C^{\textrm{MPS}}$ and the number of the multi-prototypes is $s^{*}=|\V_{\textrm{MPS}}|$.
	\end{algorithmic}
\end{algorithm}

We have the following comments for MPS:
\begin{itemize}
  \item As shown above, MPS is an unsupervised technique that does not require the number of clusters in advance. By introducing the relative reconstruction rate in \eqref{eq_10}, MPS has the ability to select the appropriate number of multi-prototypes.
  \item In MPS, $\varepsilon$  is a key parameter  to tune the number of multi-prototypes selected by the algorithm. Evidently, the smaller $\varepsilon$ is, the more number of multi-prototypes are sampled, and vice versa. Here,  $\varepsilon$ is empirically set as follows:
  \begin{equation}\label{eq_thred}
\varepsilon =\frac{1}{\rho\sqrt{n*p}}
\end{equation}
where $n$ is the number of samples, $p$ is the dimensionality of samples, and $\rho$ is a  positive constant. We show by experiments in Section \ref{sec5} that by changing $\rho$ appropriately, MPS allows the clustering results of K-Means to better adapt to the arbitrary shape data sets.

\item The computational complexity of MPS is $\mathcal{O}(np(s^{*}!)+nps^{*}t_{\textrm{K-Means}})$, where $s^{*}$ is the number of multi-prototypes by MPS, and $t_{\textrm{K-Means}}$ is the number of iterations of K-Means algorithm.

\item We can prove that  the multi-prototypes obtained by MPS achieve a constant factor approximation to the global minima of K-Means problem, see below.
  \end{itemize}

\begin{theorem}
For any data set $\mathbf{X}$, if the prototypes are constructed with MPS, then the corresponding objective function $J_{\mathbf{X}}$ satisfies:
\begin{center}
$E[J_{\mathbf{X}}] \leq 2(1-\varepsilon)(3{J_{\mathbf{X}}}^{opt}+2 n_{a}\bigtriangleup)$,
\end{center}
where $\varepsilon$ is the  termination threshold for MPS, $n_{a}=|\{\x| \|\vv(\x)-\vv^{*}(\x)\| \geq \|\x-\vv^{*}(\x)\|\}|$ and $\bigtriangleup=\varepsilon J_{\mathbf{X}} $.
\label{theorem3}
\end{theorem}

The proof is given in \textbf{Appendix} \ref{sec7-1}. Thus given a small $\varepsilon$,  the iterations of MPS can continuously optimize $\bigtriangleup$ to achieve the desired approximate upper bound on the global minima.  After Algorithm \ref{alg1}, the multi-prototypes need to be merged to recover better local minimum. In the next subsection, the  merging technique CM  is presented.

\subsection{Multi-Prototypes Convex Merging (CM)}\label{sec4-2}
In this part, a merging technique, called convex merging (CM), is proposed to recover better local minima in the case of unknown number of clusters.  CM, derived from the convex clustering paradigm \cite{Lindsten2011},  formulates the merging of the multi-prototypes task as a convex optimization problem by adding a sum-of-norms (SON) regularization to control the trade-off between the model  error and the number of clusters. The model is as follows:\\
\begin{equation}\label{eq_15}
\begin{aligned}
¡¡\min_{\bm{\mu}_{1},..., \bm{\mu}_{s^{*}} \in \Real^{p}} \frac{1}{2} \sum_{i=1}^{s^{*}} \|\bm{\mu}_{i}-\vv_{i}\|^2+\gamma \sum_{i < j} w_{ij}\|\bm{\mu}_{i}-\bm{\mu}_{j}\|,\\
    \end{aligned}
\end{equation}
where $\gamma > 0$ is a tuning parameter, $s^*$ is number of the multi-prototypes by MPS,  and the norms chosen ensure the convexity of the model. Here, $w_{ij}$  is chosen based on the number of neighboring samples $q$, and \eqref{eq 16}.

After solving \eqref{eq_15}, the optimal solutions, $\bm{\mu}^{*}_{1},..., \bm{\mu}^{*}_{s^{*}}$, are obtained. Then the multi-prototypes in $\V_\textrm{MPS}$  are assigned based on the following criteria:  for any $i$, $i^{'}$ $\in \{1, 2,..., s^{*}\}$, $\vv_{i}$ and $\vv_{i^{'}}$ can be assigned to the same cluster if and only if their optimal solutions $\|\bm{\mu}^{*}_{i}-\bm{\mu}^{*}_{i^{'}}\| \leq \eta$ for a given tolerance $\eta$. Otherwise, $i$ and $i^{'}$ are assigned to the different clusters.  Accordingly, the optimal clusters of the multi-prototypes, $\C^{\textrm{CM}}=\{\C_{1}^{\textrm{CM}}, \C_{2}^{\textrm{CM}},..., \C_{k^{*}}^{\textrm{CM}}\}$,  are formed, where  $k^{*}$ is  the estimated number of clusters.

Finally, the samples are merged into the clusters in $\C^{\textrm{CM}}$ and we get the final clusters of the data set $\C^{\textrm{MCKM}}$. In detail,  $\x_{j} \in \C^{\textrm{MCKM}}_{l}$, if $\x_{j} \in \C^{\textrm{MPS}}_{i}$ and $\vv_{i} \in \C^{\textrm{CM}}_{l}$ for $i=1, 2,..., s^{*}$, $j=1, 2,..., n$, $l=1, 2,..., k^{*}$.

In \eqref{eq_15}, $\gamma$ regulates both the assignment of the multi-prototypes and the number of clusters. When $\gamma=0$, each prototype occupies a unique cluster of its own. For a sufficiently large $\gamma$, all multi-prototypes are assigned to the same cluster.

The algorithm of CM is summarized in Algorithm \ref{alg3}, where the alternating direction method of multipliers (ADMM) \cite{Splitting2015} is used to solve \eqref{eq_15}. The details of the  optimization process can be referred to \textbf{Appendix} \ref{sec7-2} and the related studies \cite{boyd2011, Splitting2015}.
\begin{algorithm}[htb]
	\caption{Convex Merging (CM)}
	\label{alg3}
	\begin{algorithmic}[1]
        \REQUIRE The multi-prototypes $\V_\textrm{MPS}$ with $s^*$, the corresponding clusters $\C^{\textrm{MPS}}$, the weight $W$, the tuning parameter $\gamma$ and the termination $\eta$;
		\ENSURE The clusters of the data set $\C^{\textrm{MCKM}}$ and the estimated number of clusters $k^{*}$.
        \STATE Optimize \eqref{eq_15} via ADMM on the results of MPS, $\V_{\textrm{MPS}}$, and get the optimal solution $\bm{\mu}^{*}=\{\bm{\mu}^{*}_{1},..., \bm{\mu}^{*}_{s^{*}}$\};
        \STATE Form the optimal clusters of the multi-prototypes, $\C^{\textrm{CM}}=\{\C_{1}^{\textrm{CM}}, \C_{2}^{\textrm{CM}},..., \C_{k^{*}}^{\textrm{CM}}\}$, based on $\bm{\mu}^{*}$;
        \STATE Obtain the clusters of the data set $\C^{\textrm{MCKM}}$ based on the clusters $\C^{\textrm{MPS}}$ and $\C^{\textrm{CM}}$, and the estimated number of clusters $k^{*}=|\C^{\textrm{MCKM}}|$.
	\end{algorithmic}
\end{algorithm}

We have the following comments for CM:
\begin{itemize}
  \item Because the objective of CM is convex, the global minima of the merging of the multi-prototypes for a given suitable $\gamma$ is unique and is easier to obtain than the traditional merging techniques  \cite{Lindsten2011, Zhu2014}.
  \item By changing $\gamma$ in \eqref{eq_15}, the prototypes fusion path can be generated, which  enhances the explainable and comprehension of recovering the better local minima.
   \item  Originating from CC model \cite{Lindsten2011}, CM has the property that the value of $\gamma$ is inversely proportional to the estimation of the number of clusters $k^{*}$.  Based on monotonicity, a suitable $\gamma$ is sure to allow MCKM to evaluate the correct cluster number.
  \item The computational complexity of CM is $\mathcal{O}((s^{*})^2 p t_{\textrm{ADMM}})$, where $s^{*}\ll n$, and $t_{\textrm{ADMM}}$ the number of iterations of  ADMM solver.
\end{itemize}

In the next section, several experiments are performed to illustrate the performance of the proposed algorithm.

\section{Experimental results}\label{sec5}

To verify the effectiveness and efficiency of the proposed algorithm, experiments are carried out on synthetic and real-world data sets. We compare our method with four other clustering algorithms: 1) K-Means algorithm \cite{Least1982}; 2) Split-Merge K-Means algorithm (SMKM) \cite{Capo2022}; 3) Self-adaptive multiprototype-based competitive learning (SMCL) \cite{lu2019}; and 4) Convex clustering (CC) \cite{Lindsten2011}. These algorithms are chosen because they use different techniques to achieve the good approximation of the global minima. Specifically, K-Means algorithm  and SMKM  usually perform well on  relatively uniform size and linearly separable convex data sets. SMCL and CC can handle the clustering of non-convex and skewed data sets without given the cluster number. Moreover,  SMCL  and CC can estimate the correct cluster number by selecting appropriate hyper-parameters.

All experiments were run on a computer with an Intel Core i7-6700 processor and a maximum memory of 8GB. The computer runs Windows 7 with MATLAB R2017a. The experimental setup and the evaluation metrics used for clustering performance are described below. The termination parameter $\eta$ is set to $10^{-6}$ for all algorithms except SMCL which is empirically set at 0.001. The positive constant $\kappa$ in (\ref{eq 16}) is set at $0.9$ for MCKM and CC. The remaining parameters need to be fine-tuned in the experiments.

\subsection{Evaluation Metrics}
In order to evaluate the performances of the clustering algorithms, three metrics are used. They are: the F-measure ($\textbf{F}^{*}$), Normalized Mutual Information (\textbf{NMI}), and Adjusted Rand Index (\textbf{ARI}) \cite{parker2013accelerating, mei2016large, Hubert1985Comparingpartitions}. They measure the agreement with the ground truth and the clustering results. Let $n$ be the total number of samples, $\{\C_1, \C_2, \cdots,\C_{k}\}$ be the partition of the ground truth, and $\{\hat\C_1, \hat\C_2, \cdots,\hat\C_{\hat k}\}$ be the partition by an algorithm. Denote $|\cdot|$ the cardinality of a set. Let $\hat n_{i}=|\hat\C_i|$, $n_{l}=|\C_l|$, and $n_{i}^{l}=|\C_l\cap\hat\C_i|$, where $i=1,2,\cdots,\hat k$ and $l=1,2,\cdots, k$. Then the measure $F(l,i)=\frac{2 n_{i}^{l}}{n_{l}+\hat n_{i}}$ is the harmonic mean of the precision and recall of $\C_l$ and its potential prediction $\hat\C_i$. The overall F-measure $\textbf{F}^{*}$, \textbf{NMI} and \textbf{ARI} are defined as follows:
\begin{equation}
\textbf{F}^{*}=\sum_{l=1}^k \frac{n_{l}}{n}\max\{F(l,i)|i=1,\cdots,\hat k.\},
\end{equation}
\begin{equation}
\textbf{NMI}=\frac{\sum\limits_{i=1}^{\hat k}\sum\limits_{l=1}^{k} n_{i}^{l}\log(\frac{n\cdot n_{i}^{l}}{\hat n_{i}\cdot n_{l}})} {\sqrt{\left(\sum\limits_{i=1}^{\hat k}\hat n_{i}\log(\frac{\hat n_{i}}{n})\right)\left(\sum\limits_{l=1}^{k} n_{l}\log(\frac{n_{l}}{n})\right)}},
\end{equation}
\begin{equation}
\textbf{ARI}=\frac{\sum\limits_{i=1}^{\hat k}\sum\limits_{l=1}^{k}\tbinom{n_{i}^{l}}{2}-
\sum\limits_{i=1}^{\hat k}\tbinom{t_{i}}{2}\sum\limits_{l=1}^{k}\tbinom{s_{l}}{2}/\tbinom{n}{2}}
{\frac{1}{2}\left(\sum\limits_{i=1}^{\hat k}\tbinom{t_{i}}{2}+ \sum\limits_{l=1}^k\tbinom{s_{l}}{2}\right)-\sum\limits_{i=1}^{\hat k} \tbinom{t_{i}}{2}\sum\limits_{l=1}^{k}\tbinom{s_{l}}{2}/\tbinom{n}{2}},
\end{equation}
where $\tbinom{n}{i}=\frac{n!}{i!(n-i)!}$, $s_{l}=\sum_{i=1}^{\hat k} n_{i}^{l}$, and $t_{i}=\sum_{l=1}^k n_{i}^{l}$.

\subsection{Experiments on Synthetic Data Sets} \label{Synthetic}
\begin{figure}[htb]
 \centering
 \subfloat[D1]{\label{fig4a}\includegraphics[width=0.15\textwidth]{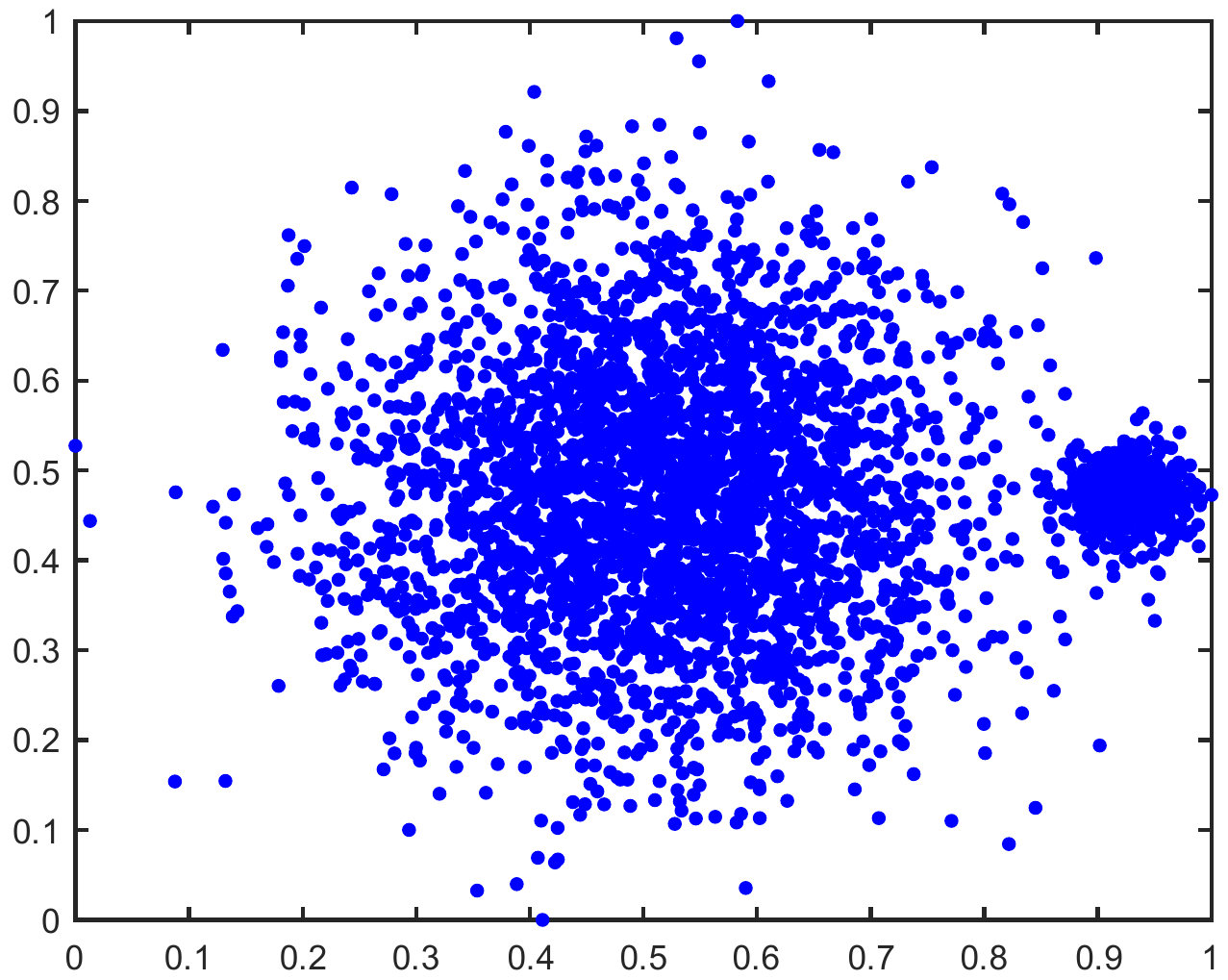}}~~
 \subfloat[D2]{\label{fig4b}\includegraphics[width=0.15\textwidth]{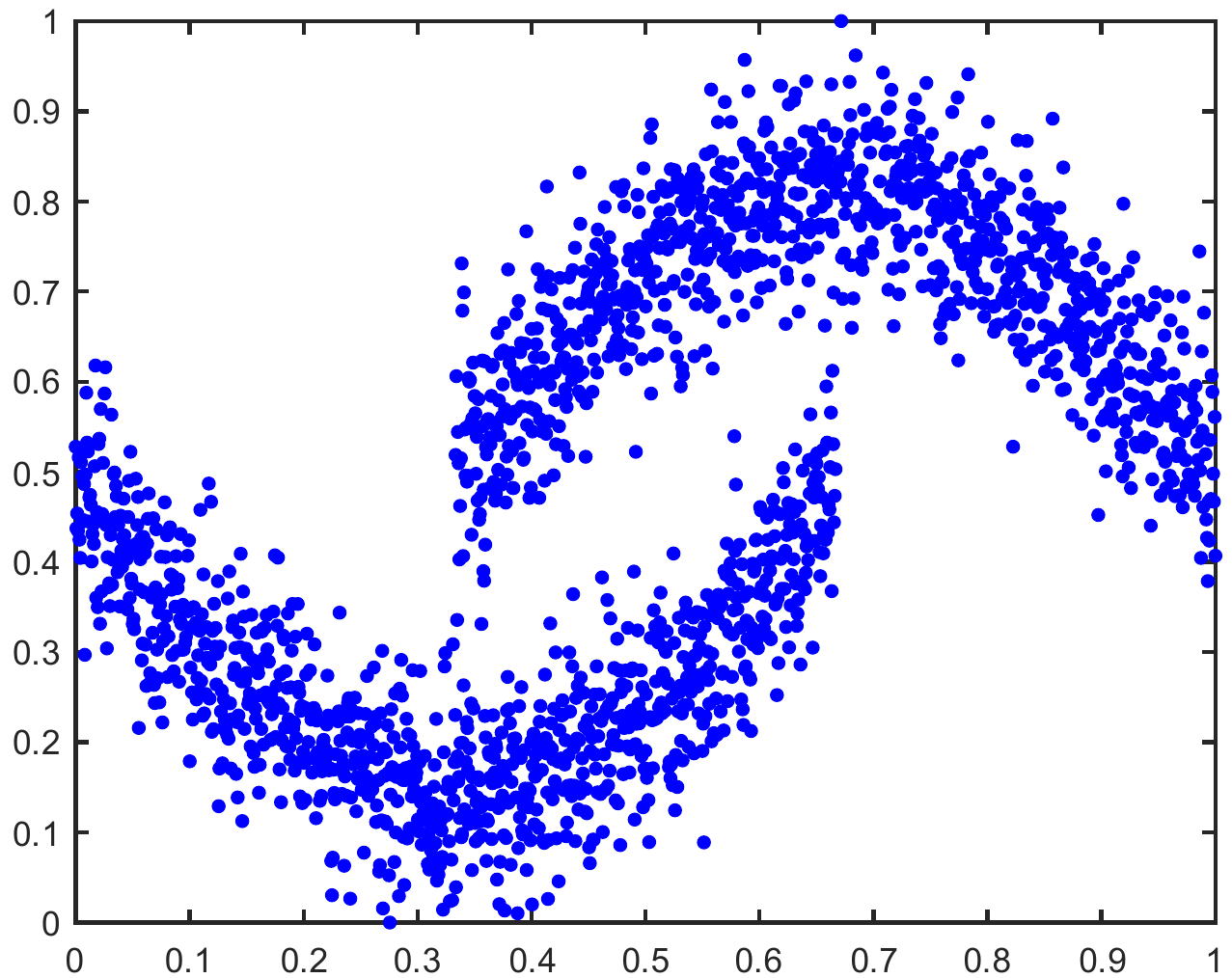}}~~
 \subfloat[D3]{\label{fig4c}\includegraphics[width=0.15\textwidth]{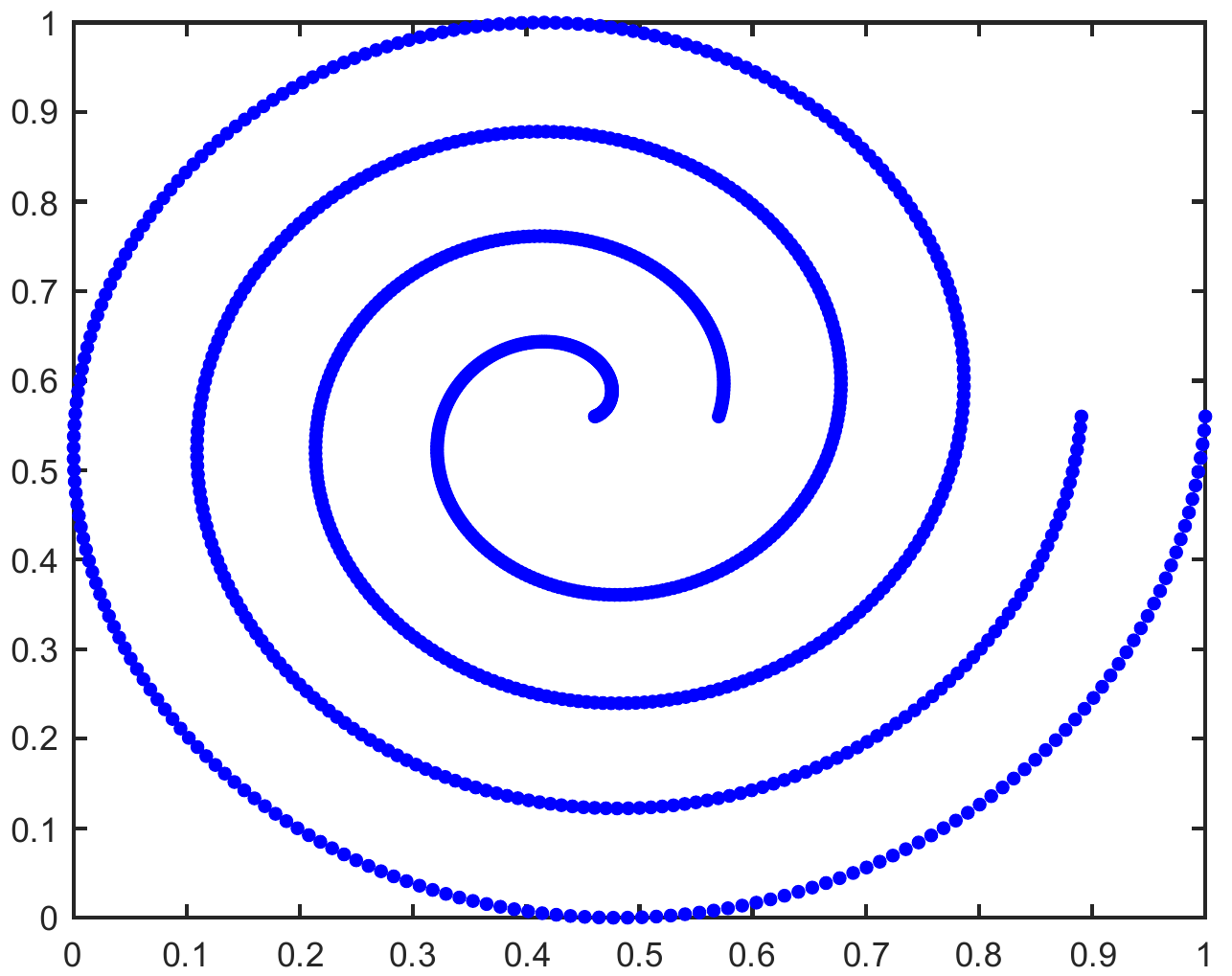}}\\
 \subfloat[D4]{\label{fig4d}\includegraphics[width=0.15\textwidth]{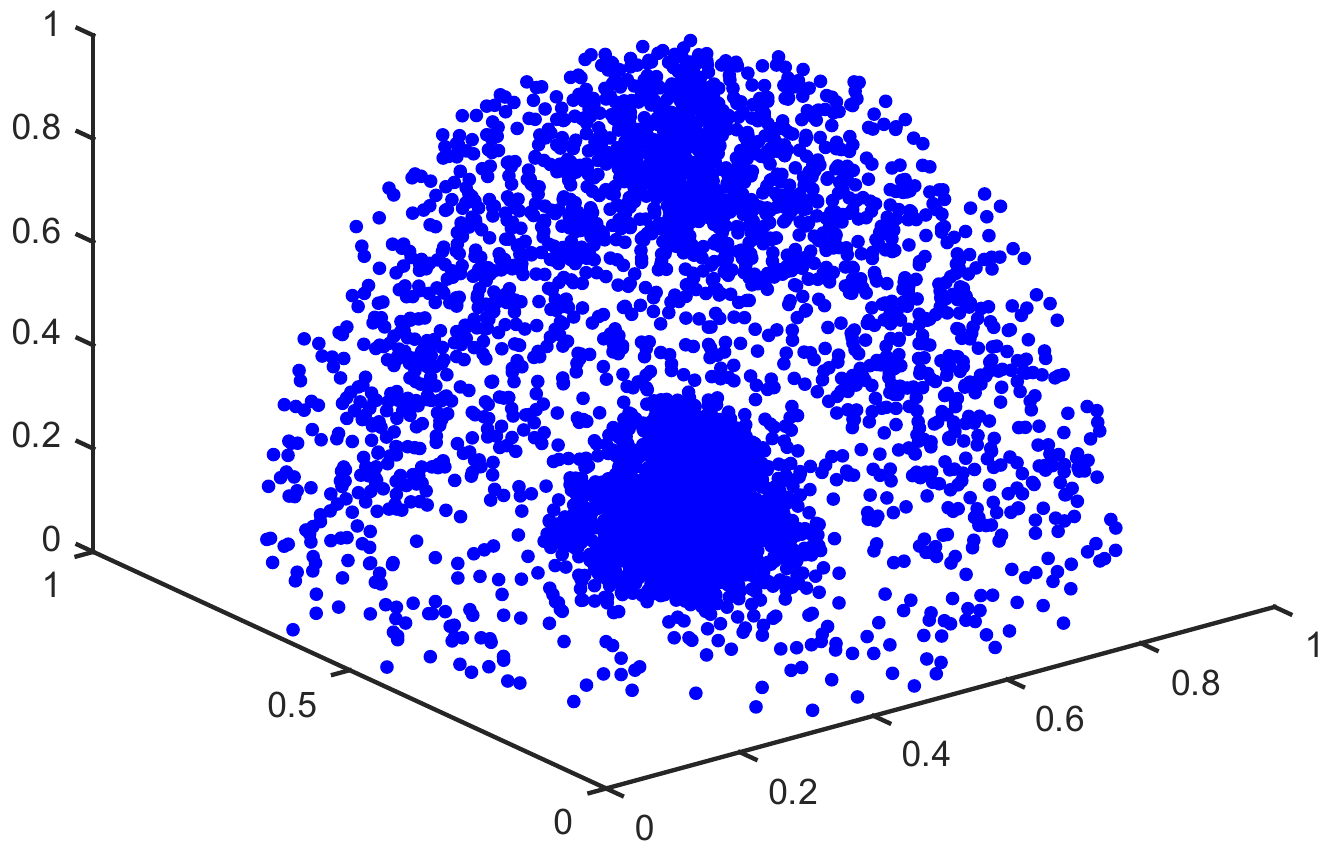}}~~
 \subfloat[D5]{\label{fig4e}\includegraphics[width=0.15\textwidth]{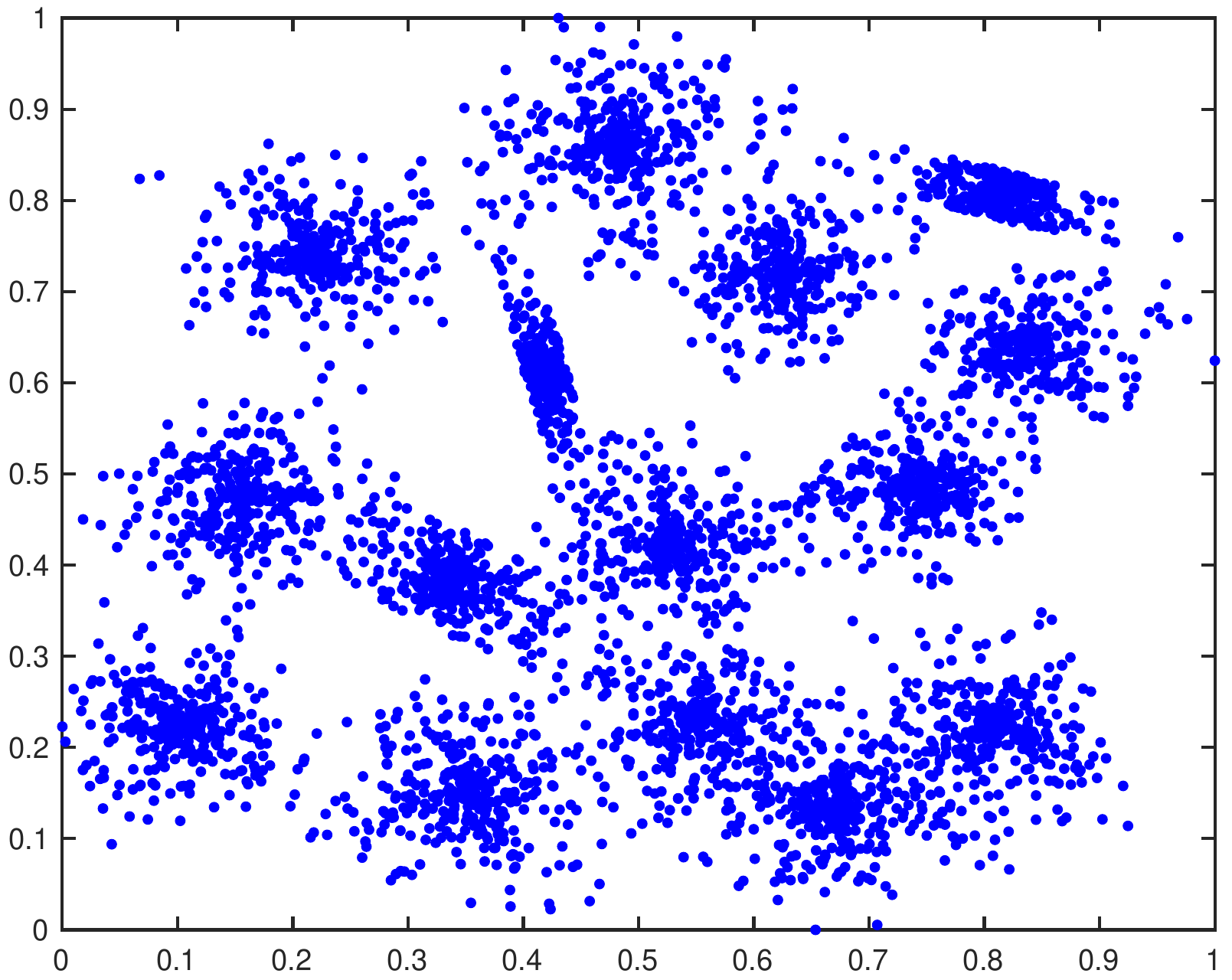}}~~
 \subfloat[D6]{\label{fig4f}\includegraphics[width=0.15\textwidth]{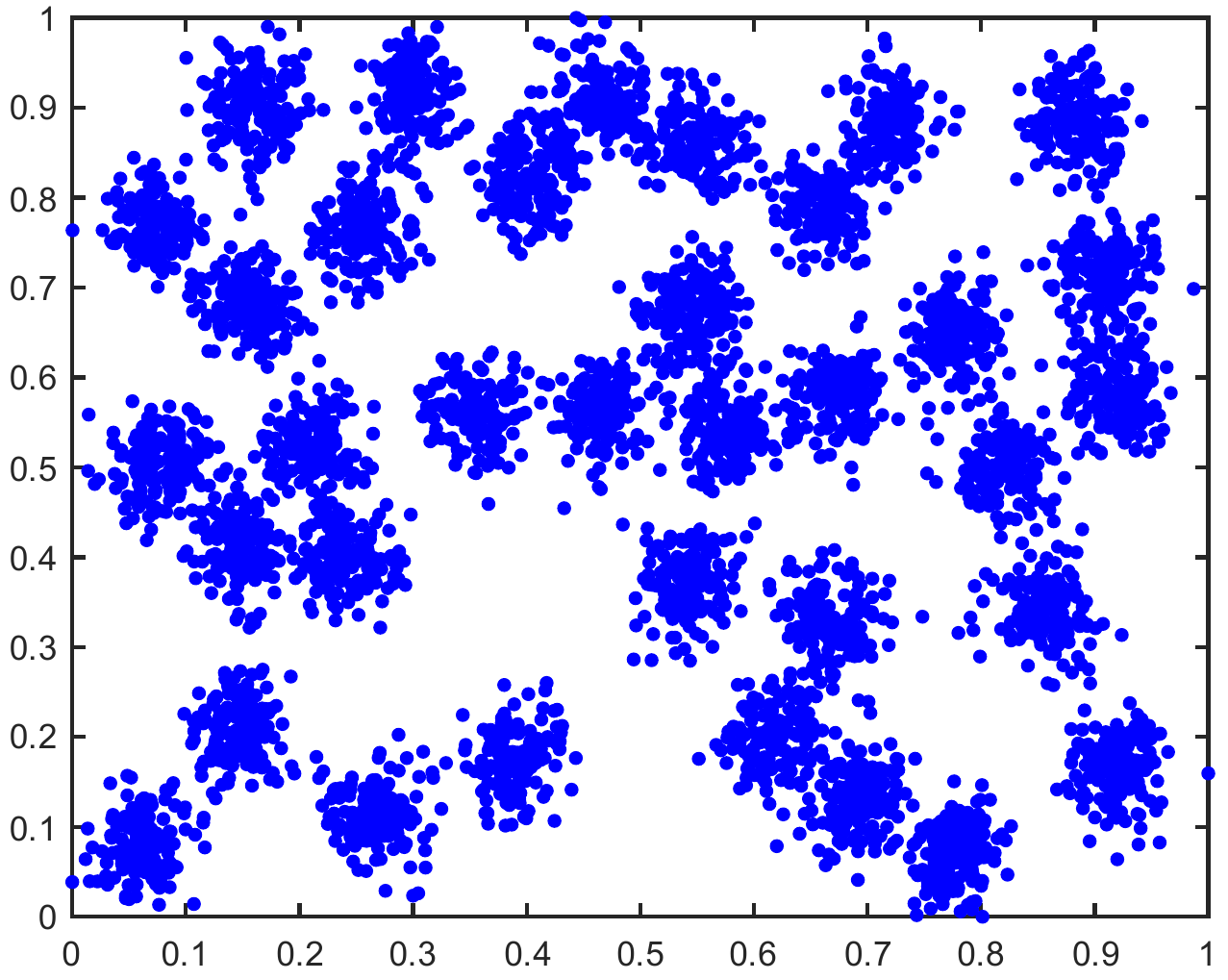}}
 \caption{Six synthetic data sets. They include unbalanced data set, non-convex data set, and convex data set with a large number of clusters.}\label{fig5}
\end{figure}

Six normalized synthetic data sets are selected for clustering in the first set of experiments, see Fig. \ref{fig5}. They include  unbalanced data set, non-convex data sets, and convex data sets with large number of clusters.  The detailed information on the data sets is given in Table \ref{table2}, where $n$ is the number of training size, $p$ is the dimensionality of samples, and $k$ is the true number of clusters. In order to have a better understanding of MCKM,  the performance of the multi-prototypes sampling (MPS) and the convex merging (CM) are shown respectively in Sections \ref{sec5-1-1} and \ref{sec5-1-2}.

\subsubsection{Performance of Multi-Prototypes Sampling}\label{sec5-1-1}
\begin{figure}[htb]
 \centering
  \subfloat[D1, $k=2$]{\label{fig5a}\includegraphics[width=0.15\textwidth]{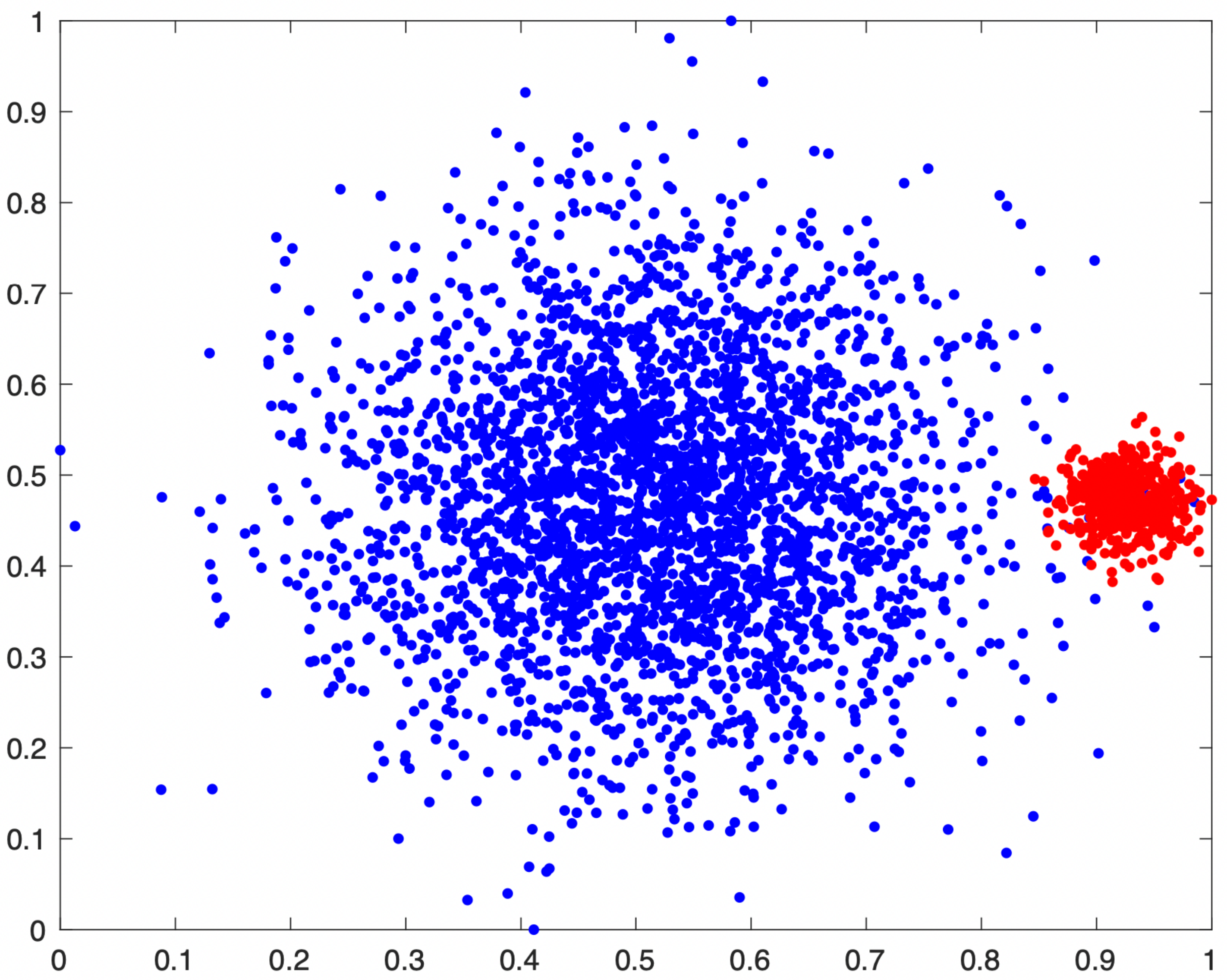}}~~
 \subfloat[D2, $k=2$]{\label{fig5b}\includegraphics[width=0.15\textwidth]{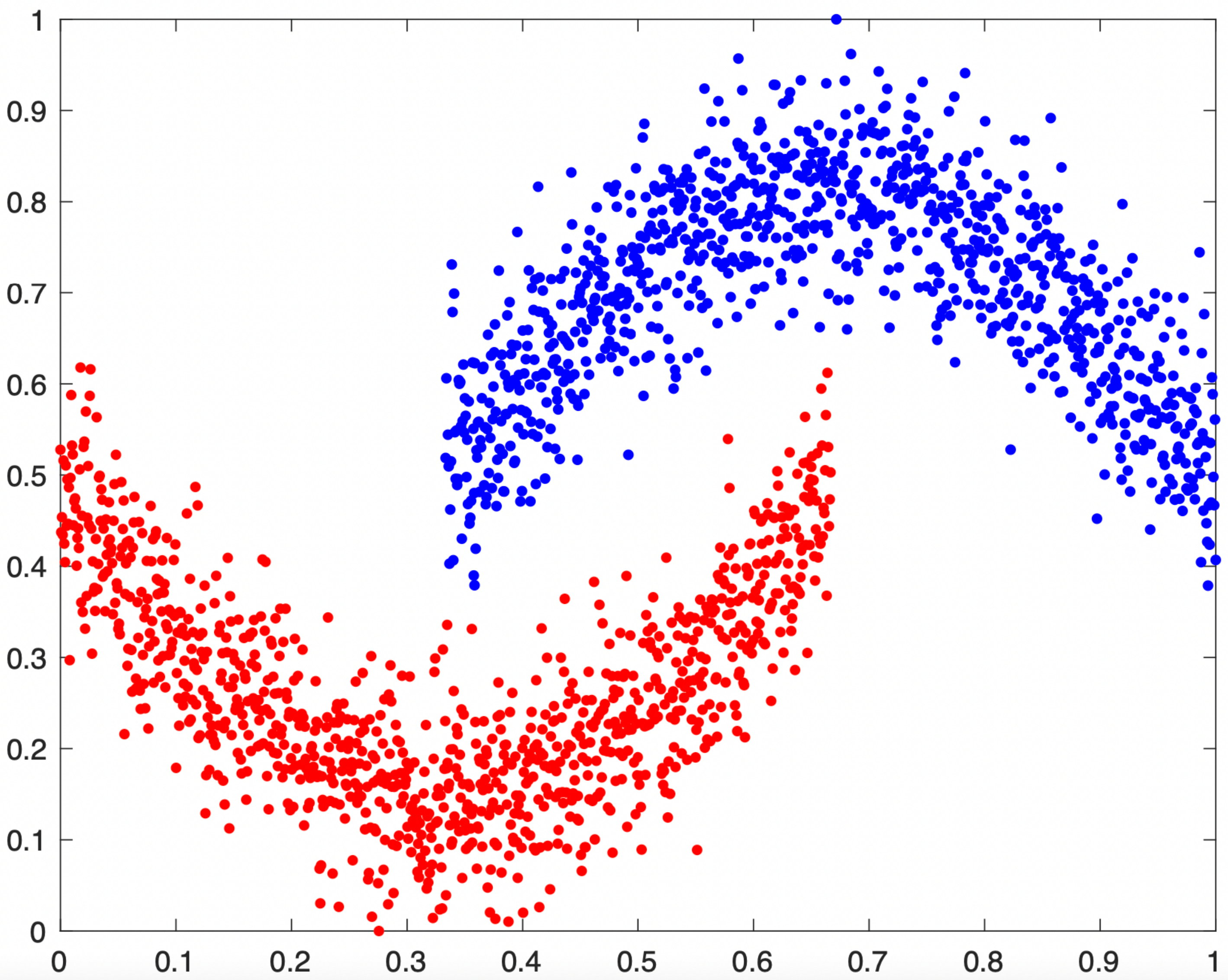}}~~\\
 \subfloat[$\rho=0.1$]{\label{fig5c}\includegraphics[width=0.15\textwidth]{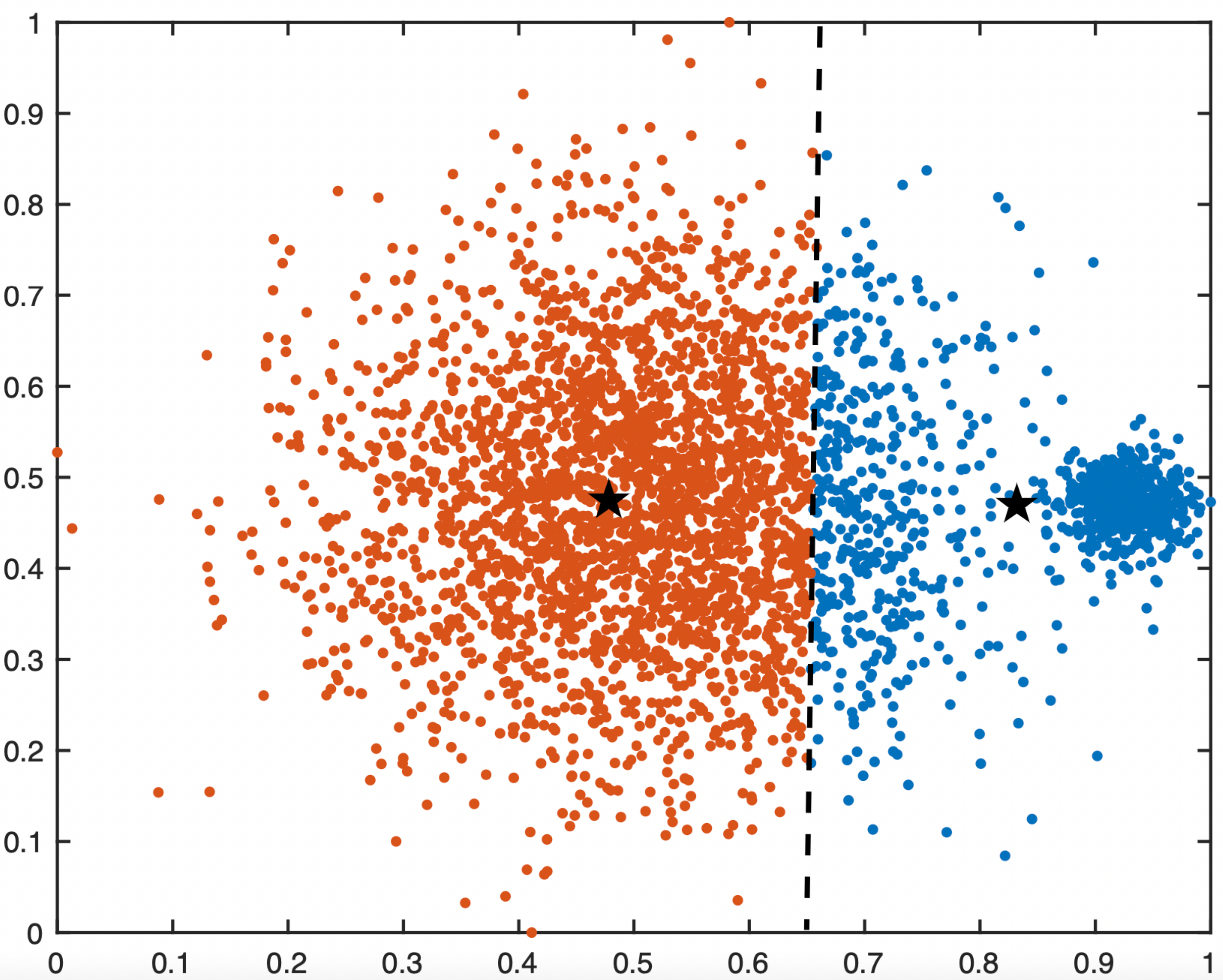}}~~
 \subfloat[$\rho=1$]{\label{fig5d}\includegraphics[width=0.15\textwidth]{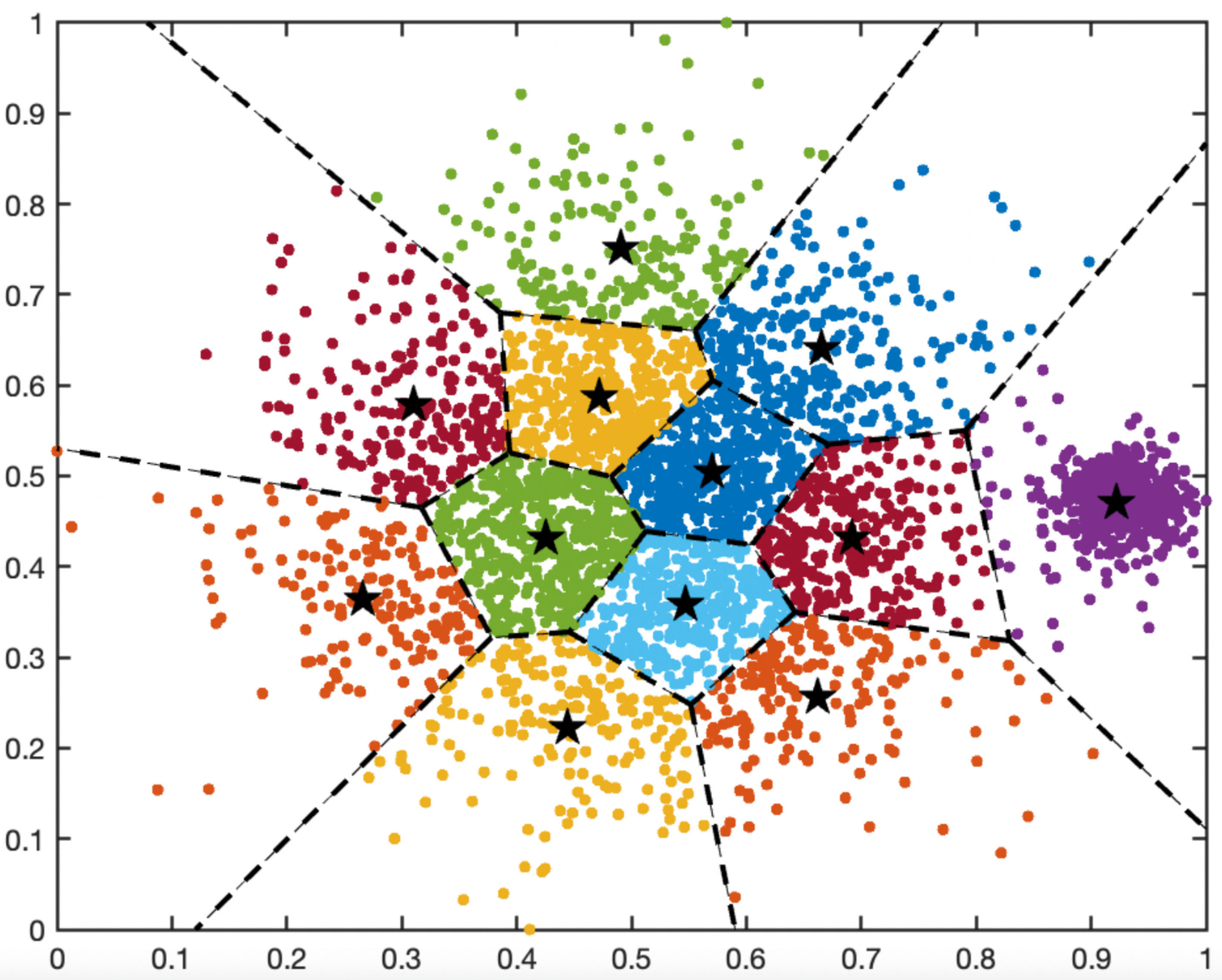}}~~
 \subfloat[$\rho=5$]{\label{fig5e}\includegraphics[width=0.15\textwidth]{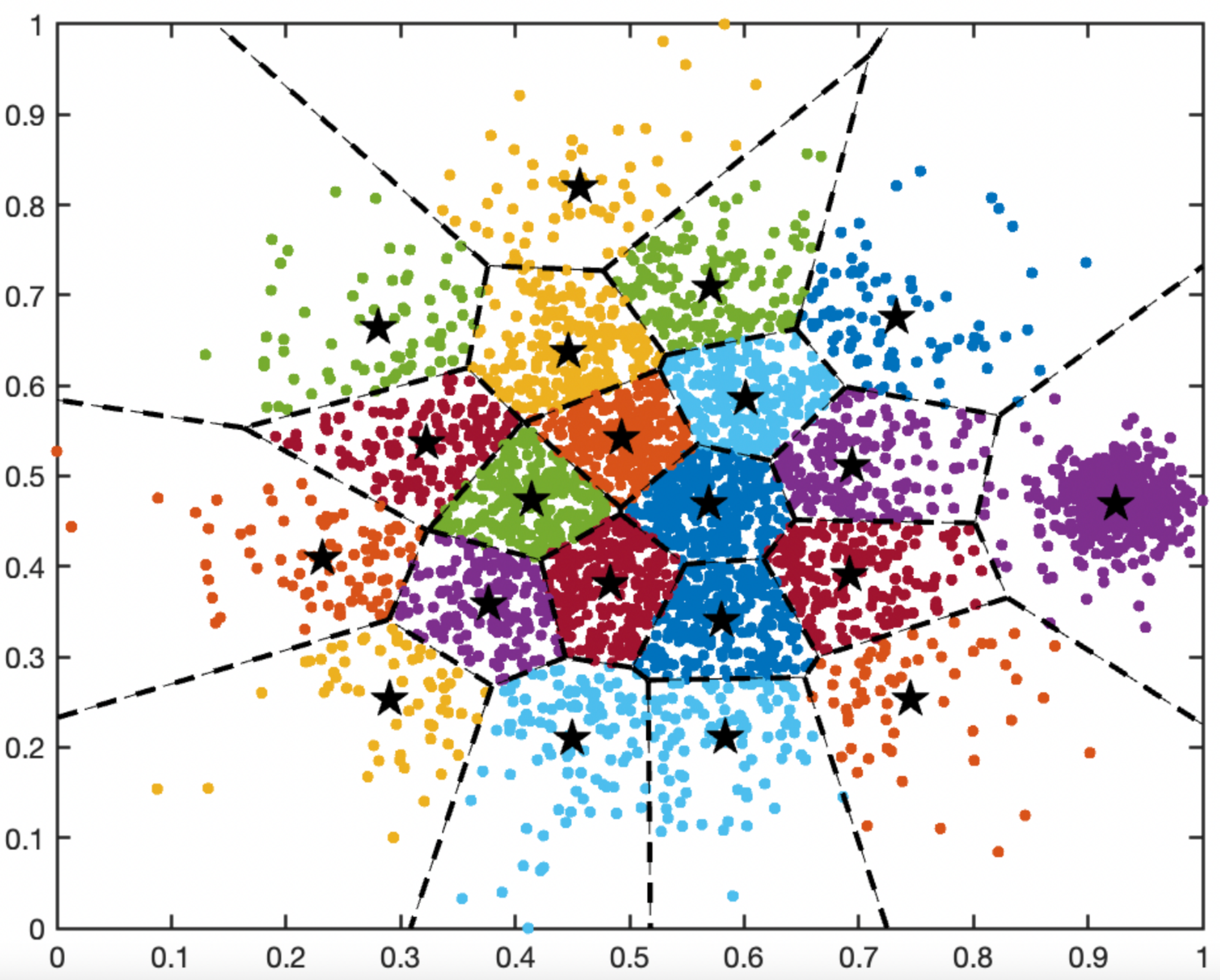}}\\
 \subfloat[$\rho=0.1$]{\label{fig5f}\includegraphics[width=0.15\textwidth]{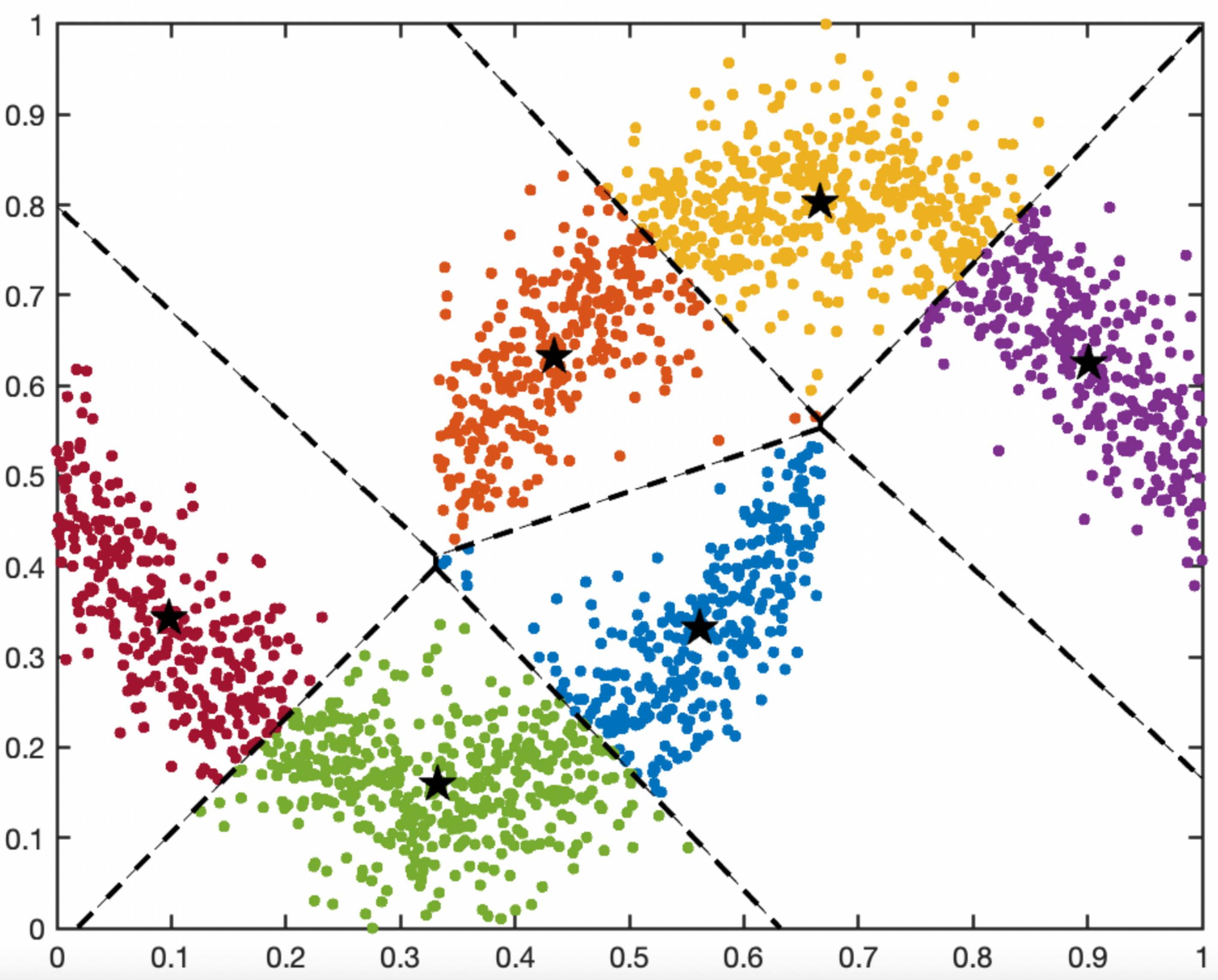}}~~
 \subfloat[$\rho=1$]{\label{fig5g}\includegraphics[width=0.15\textwidth]{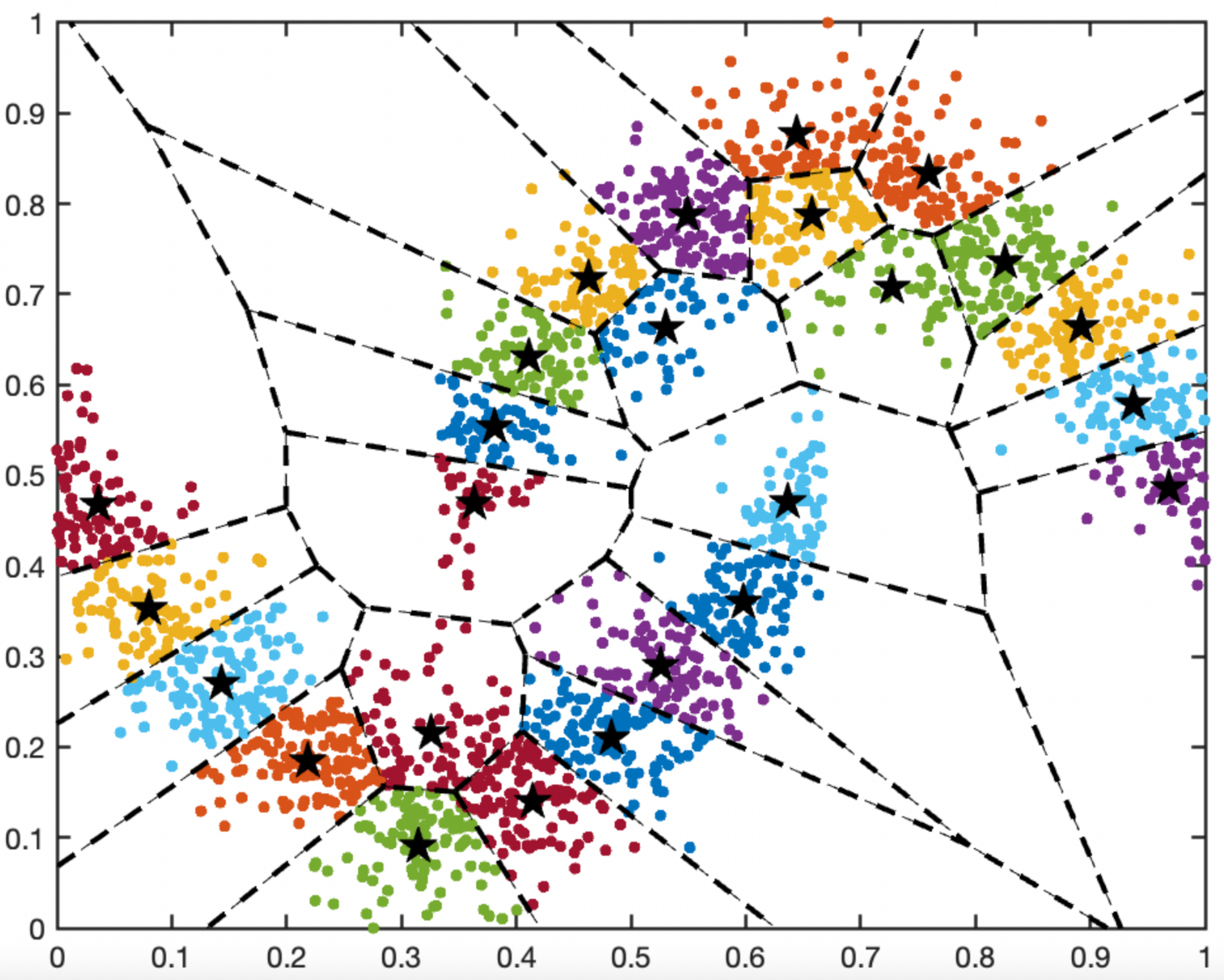}}~~
 \subfloat[$\rho=5$]{\label{fig5h}\includegraphics[width=0.15\textwidth]{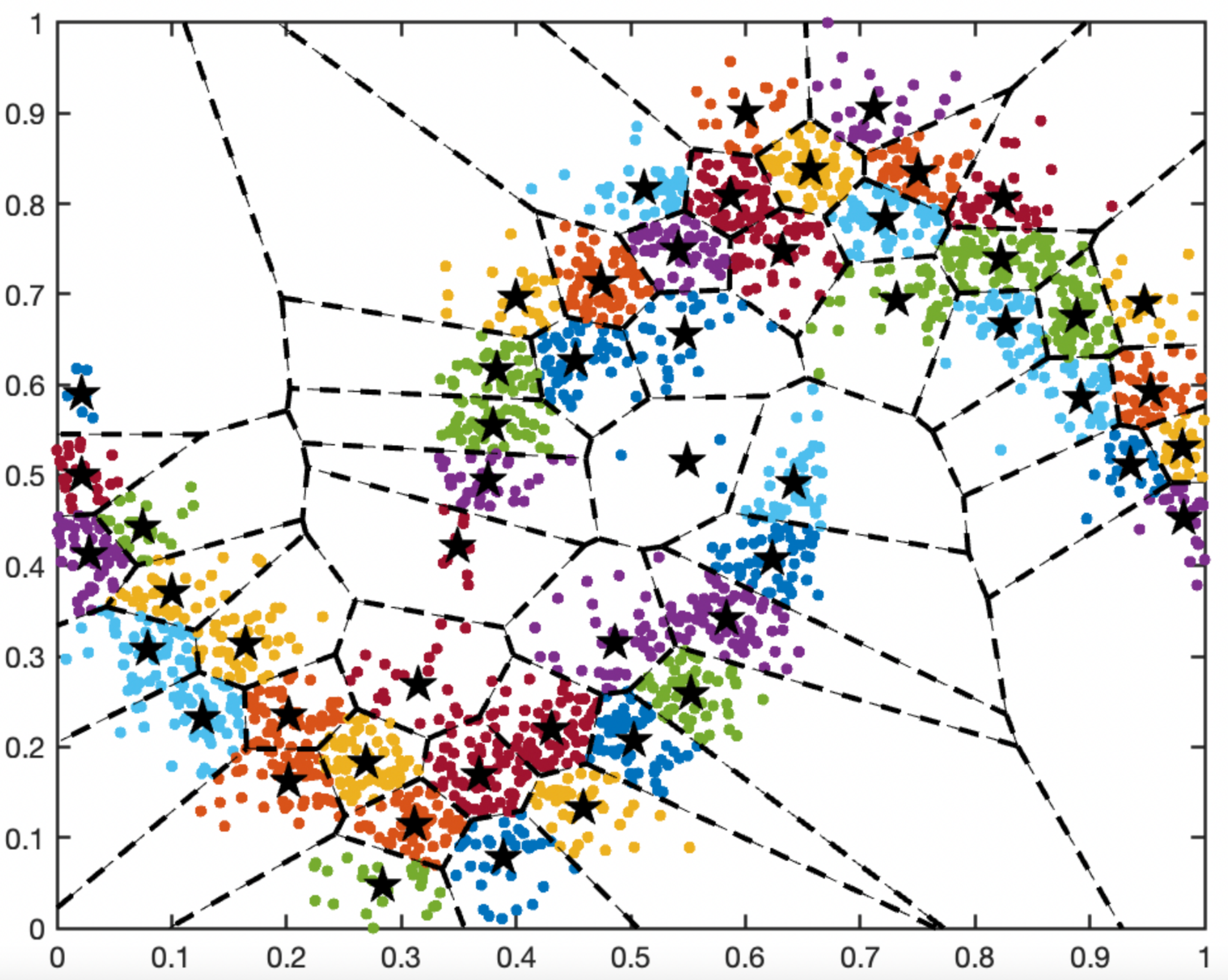}}
 \caption{{{The true clusters of D1 and D2 in \ref{fig5a} and \ref{fig5b}. The results of MPS and the corresponding Voronoi partition on D1 and D2 with $\rho=0.1, 1, 5$ in  \ref{fig5c}-\ref{fig5e} and \ref{fig5f}-\ref{fig5h}, respectively, where the black stars are the final multi-prototypes}}.}\label{fig5-1}
\end{figure}

Here we show that MPS can adapt to the arbitrary shape data sets by choosing appropriate $\rho$ in \eqref{eq_thred}.  To illustrate these, {{MPS is performed on D1 and D2 with $\rho=0.1, 1, 5$ respectively. The results and corresponding Voronoi partition are shown in Fig. \ref{fig5-1}. }}

\begin{figure}[htb]
 \centering
 \subfloat[D1, $s^{*}=12$]{\label{fig6a}\includegraphics[width=0.15\textwidth]{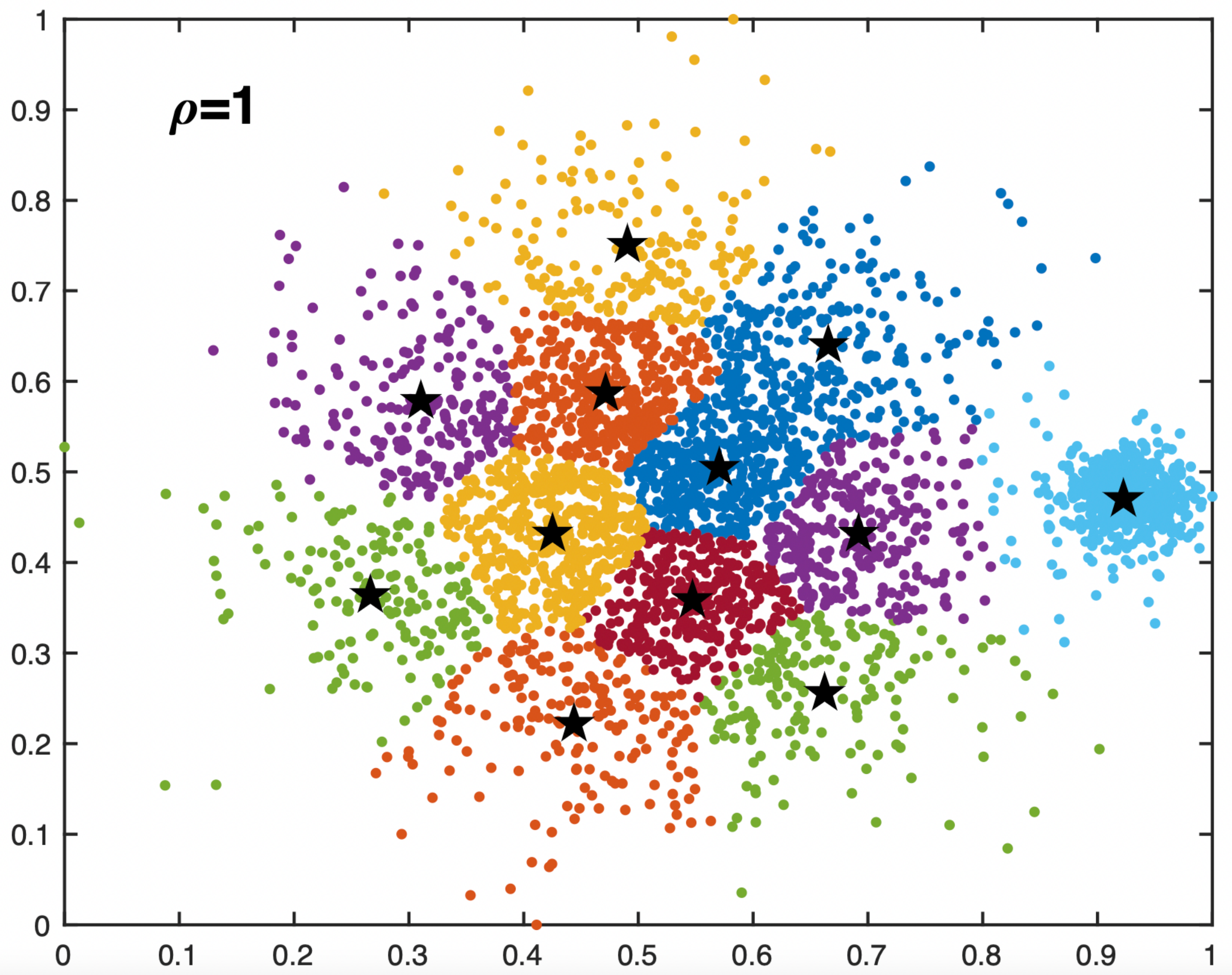}}~~
 \subfloat[D2, $s^{*}=25$]{\label{fig6b}\includegraphics[width=0.15\textwidth]{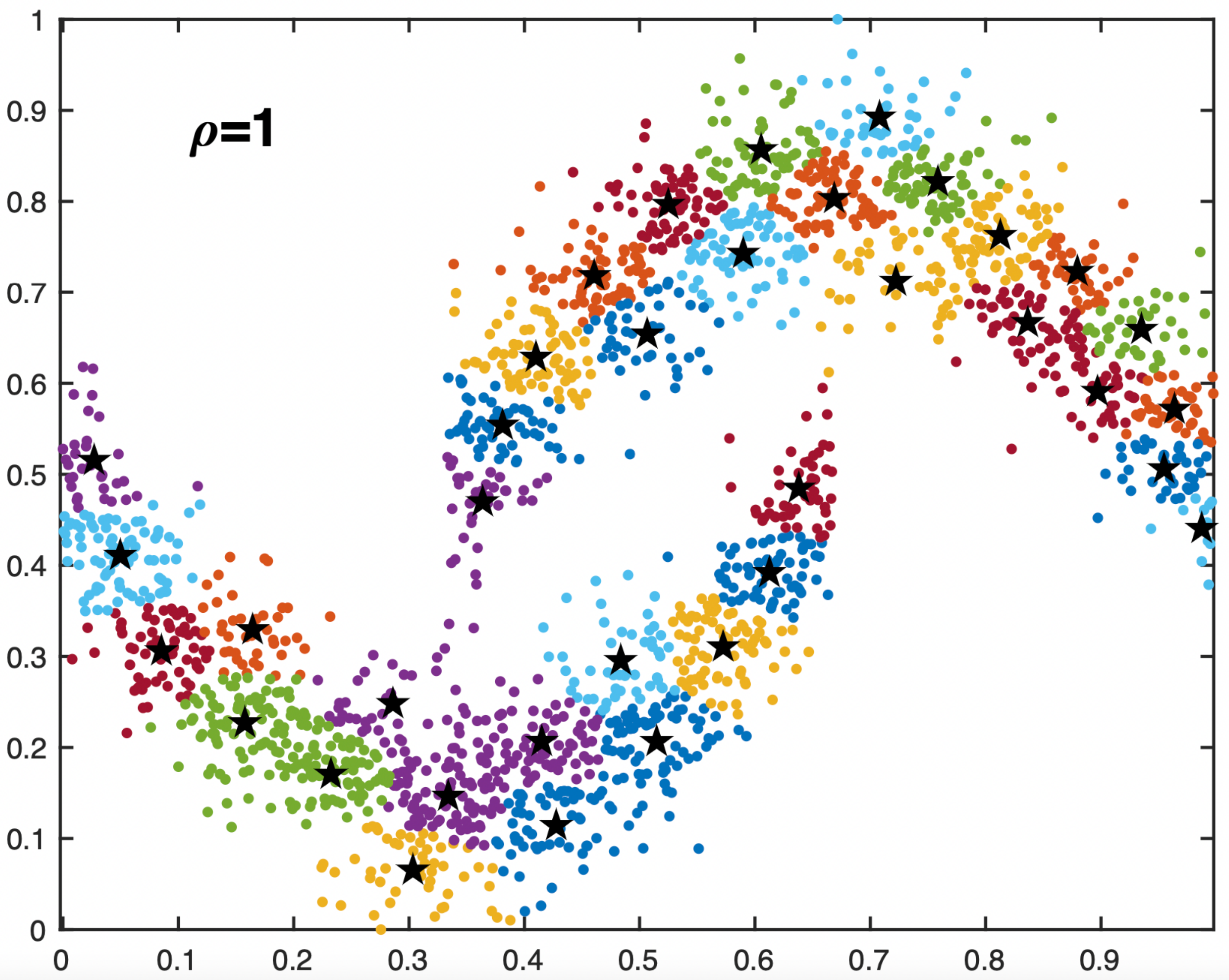}}~~
 \subfloat[D3, $s^{*}=156$]{\label{fig6c}\includegraphics[width=0.15\textwidth]{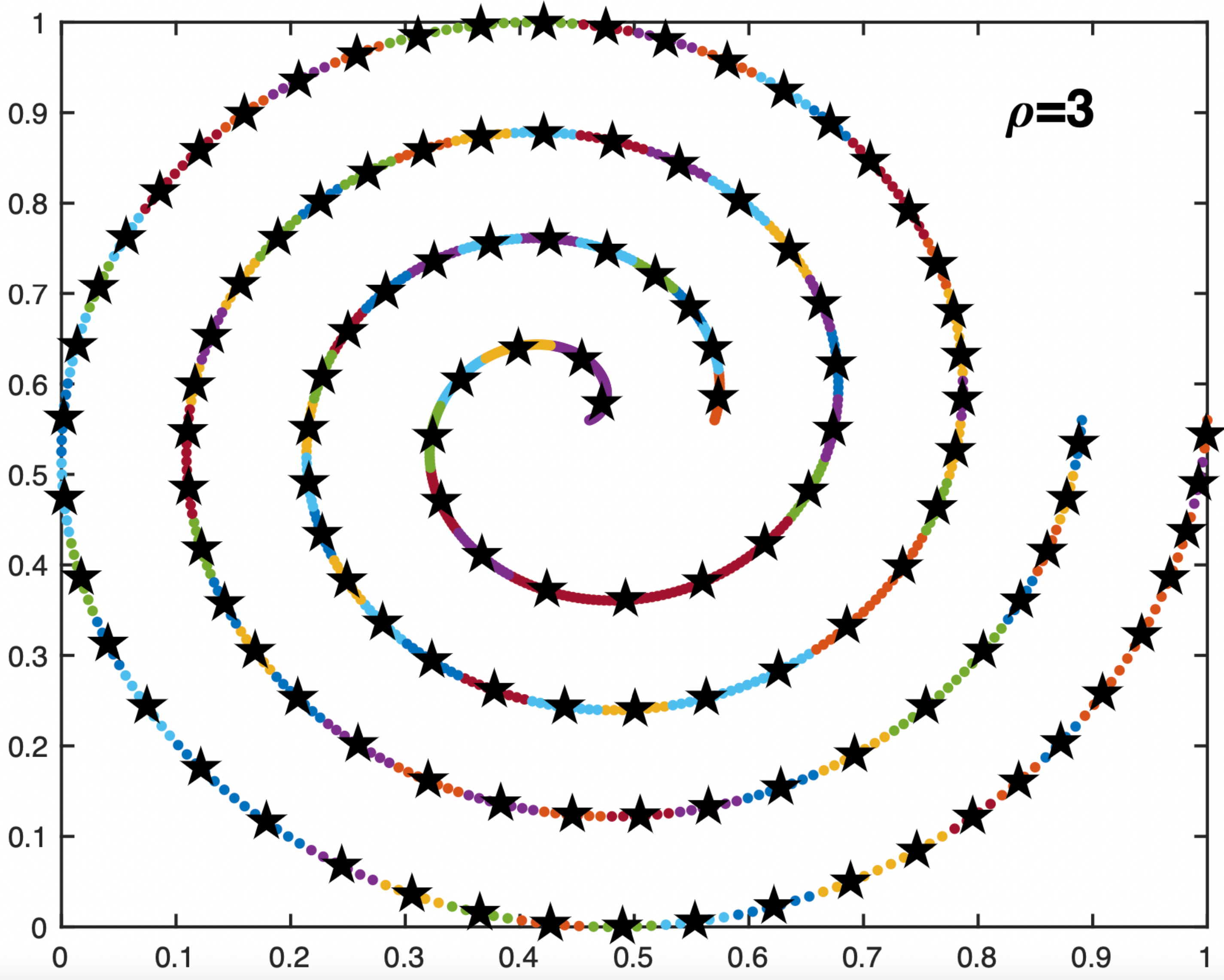}}\\
 \subfloat[D4, $s^{*}=33$]{\label{fig6e}\includegraphics[width=0.15\textwidth]{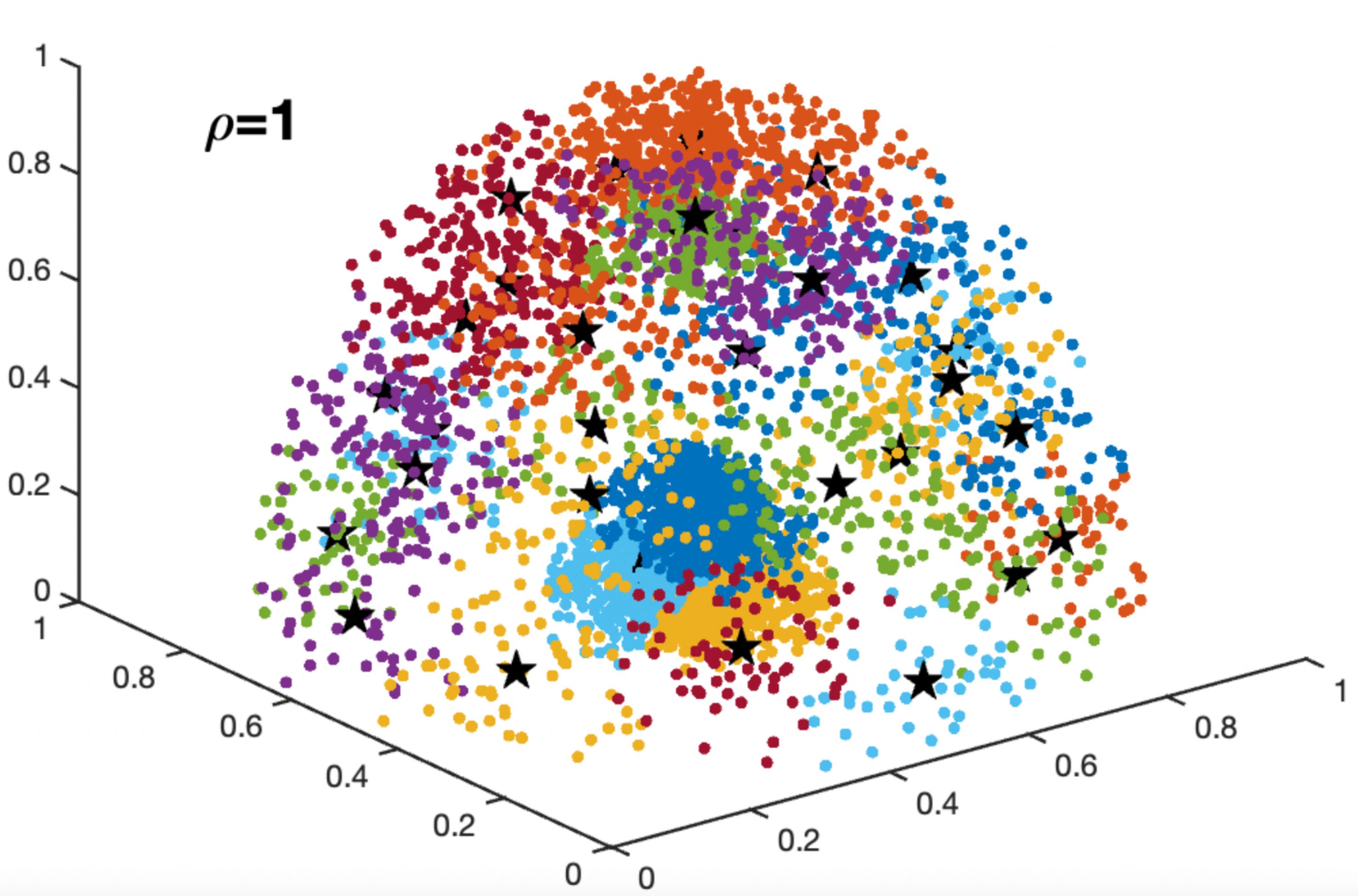}}~~
 \subfloat[D5, $s^{*}=31$]{\label{fig6f}\includegraphics[width=0.15\textwidth]{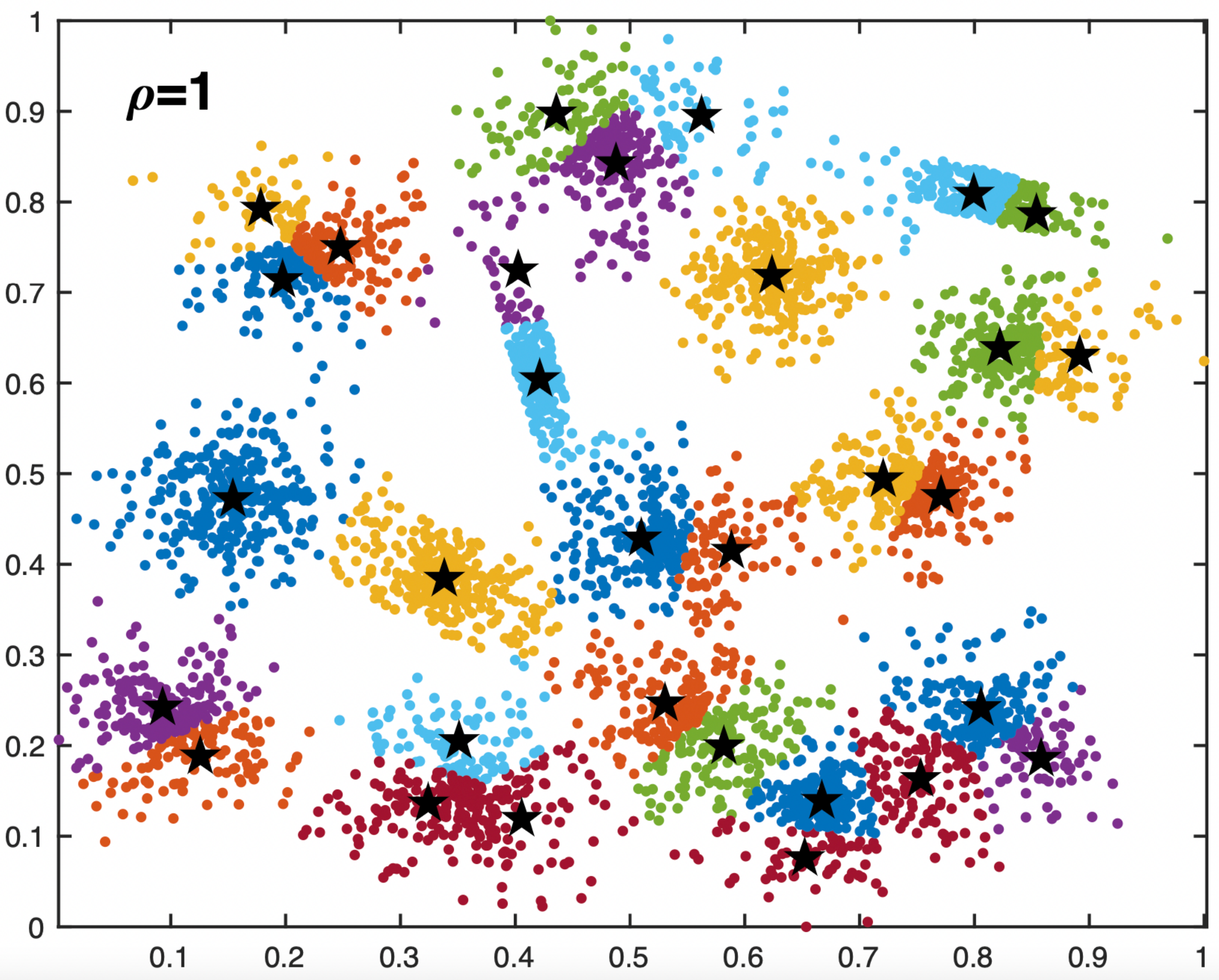}}~~
 \subfloat[D6, $s^{*}=39$]{\label{fig6h}\includegraphics[width=0.15\textwidth]{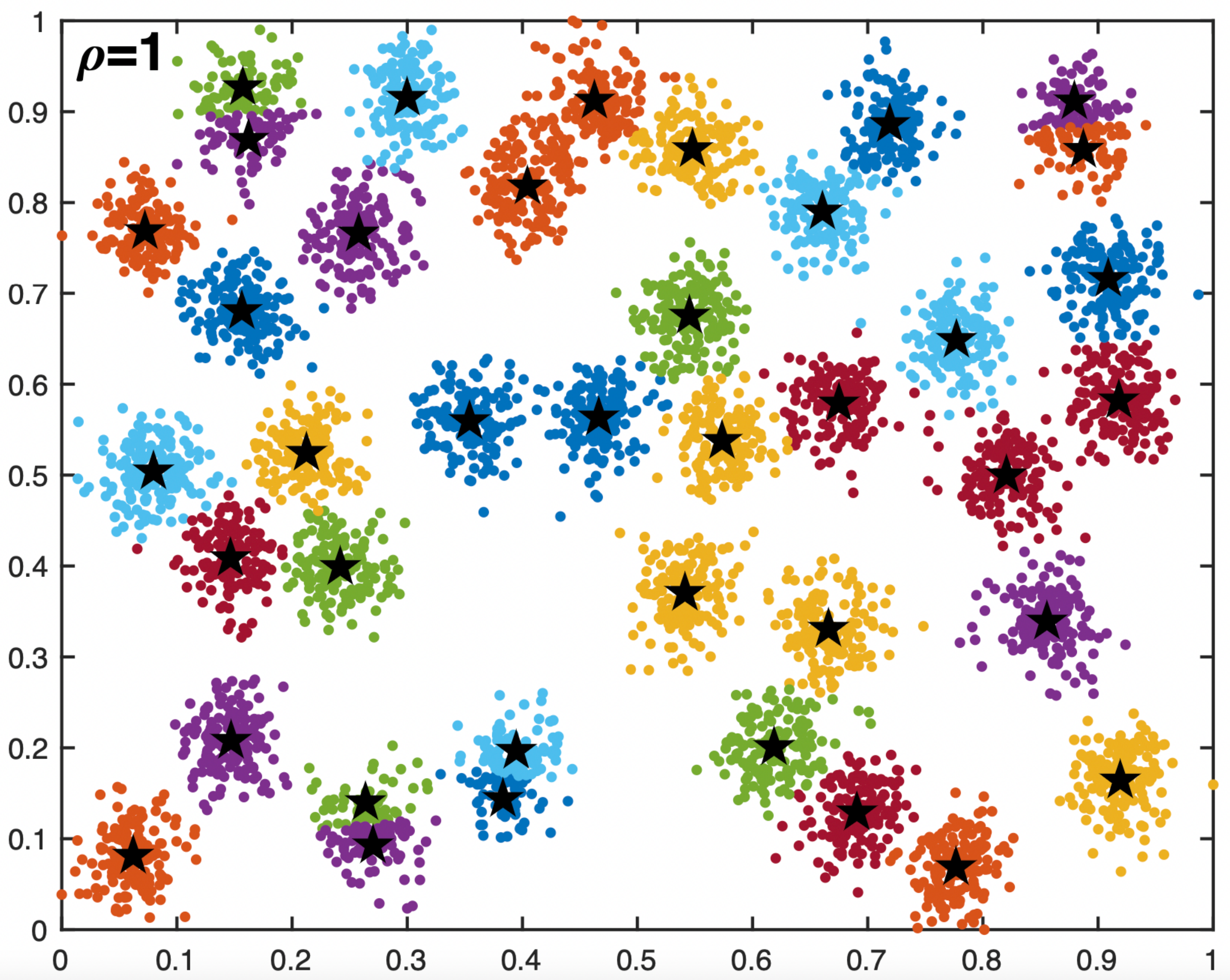}}
 \caption{{{The results of MPS on the six data sets with the  appropriate $\rho$}}, where the black stars are the final multi-prototypes and $s^{*}$ is the number of the multi-prototypes.}\label{fig6}
\end{figure}

{{The true clusters of D1 and D2 are shown in Fig. \ref{fig5a} and \ref{fig5b}. When $\rho=0.1$, we  observe from the corresponding Voronoi partition in Fig. \ref{fig5c} and \ref{fig5f} that some prototypes obtained by MPS are put in the centroid of the two true clusters. When $\rho=5$, some prototypes obtained by MPS lie in the outliers on D2, as shown in Fig. \ref{fig5h}. For D1, when $\rho=1$ and $\rho=5$, there is no under-refinement structure of the true clusters, as shown in Fig. \ref{fig5d} and \ref{fig5e}. Naturally, we use a small number of multi-prototypes for the next CM. Therefore, in MPS, $\rho=1$ is appropriate for D1 and D2.  

For the better performance of MPS on the other four data sets, we empirically set $\rho=3$ in D3, and $\rho=1$ in the rest of data sets.  Then, the results of MPS on the  the six data sets with the  appropriate $\rho$ are shown in Fig. \ref{fig6}.}}

From Fig. \ref{fig6}, it can be found that by changing $\rho$ appropriately, MPS allows the clustering results of K-Means to better adapt to the arbitrary shape data sets. In detail, for the arbitrary shape data set,  MPS with an appropriate $\rho$ can achieve that each true cluster have one or more prototypes, and none of the prototypes are put in the centroid of multiple true clusters. Based on the results of MPS, the better local minima of K-Means can be obtained by the subsequent convex merging.

\subsubsection{Performance of MCKM} \label{sec5-1-2}

\begin{figure*}[htb]
\centering
\begin{minipage}[t]{0.17\textwidth}
\centerline{\includegraphics[width=1\textwidth]{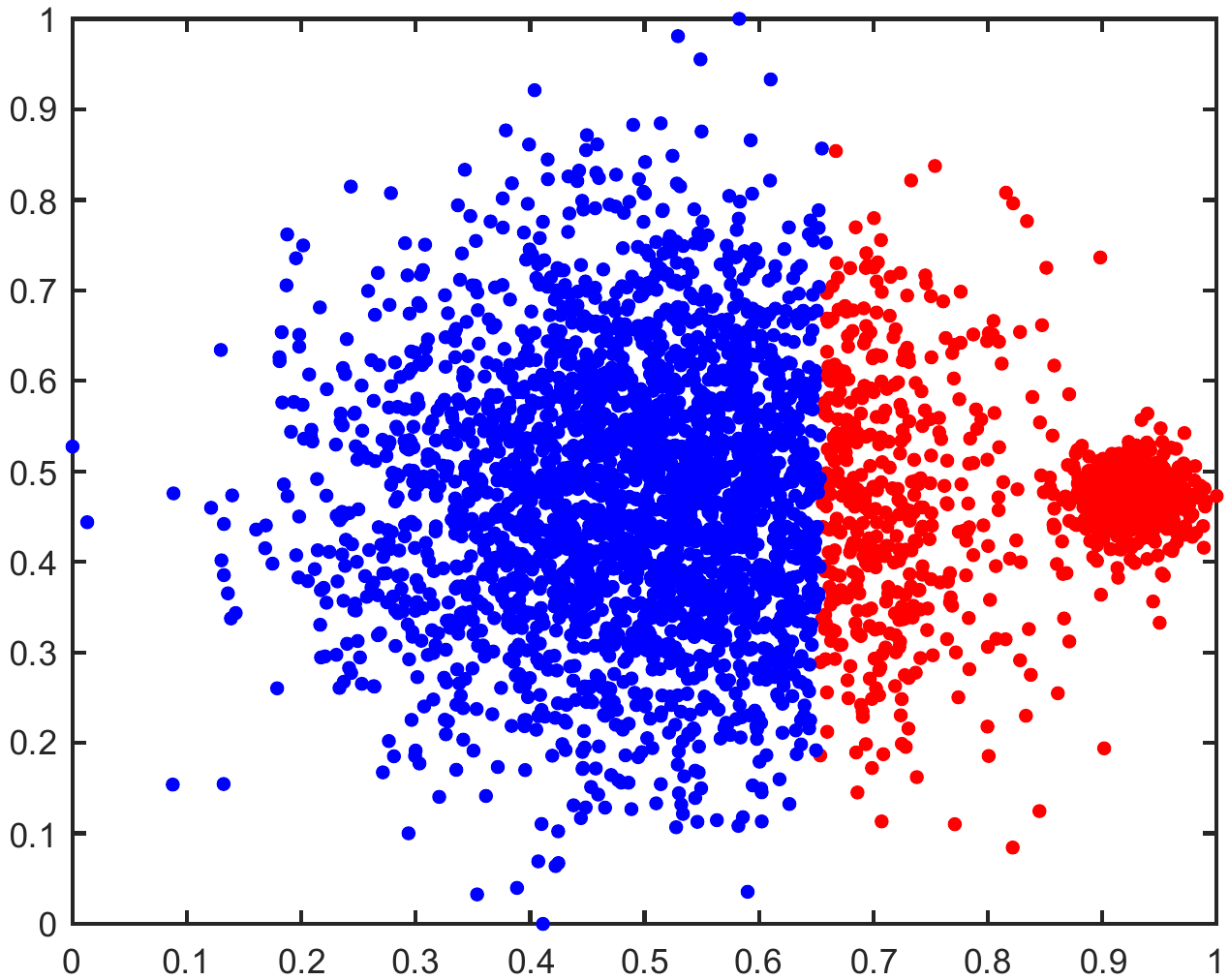}}
\vspace{3mm}
\centerline{\includegraphics[width=1\textwidth]{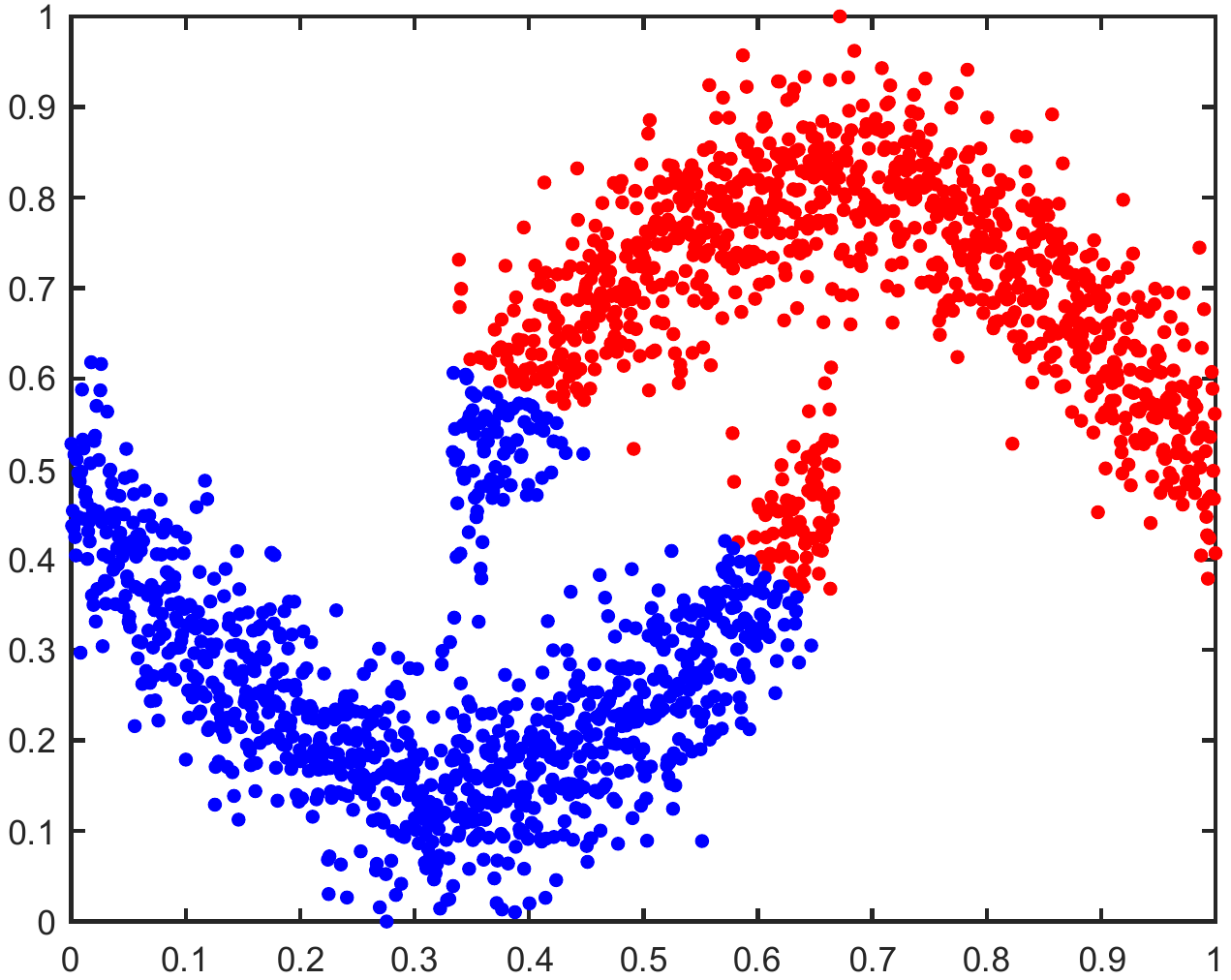}}
\vspace{3mm}
\centerline{\includegraphics[width=1\textwidth]{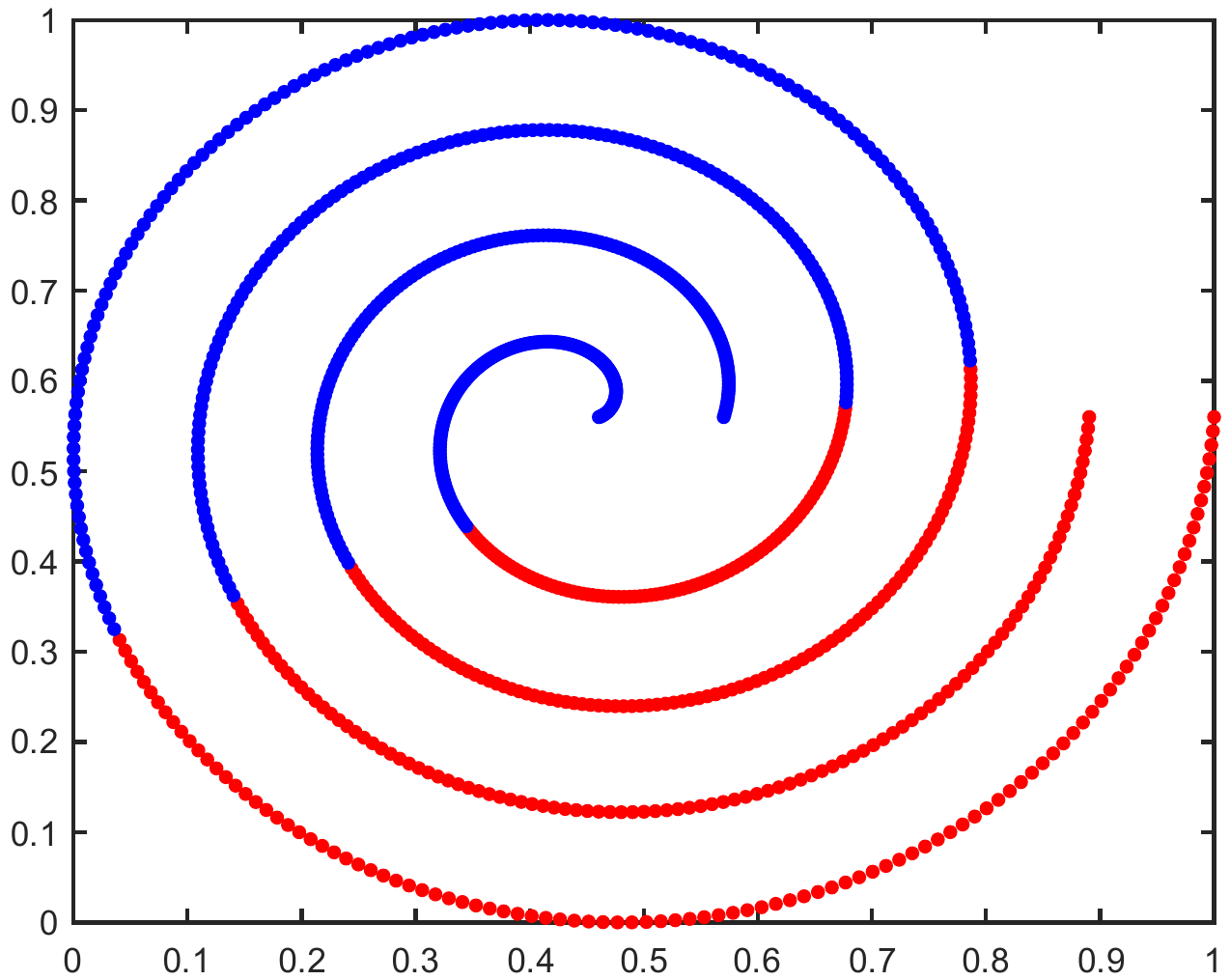}}
\vspace{3mm}
\centerline{\includegraphics[width=1\textwidth]{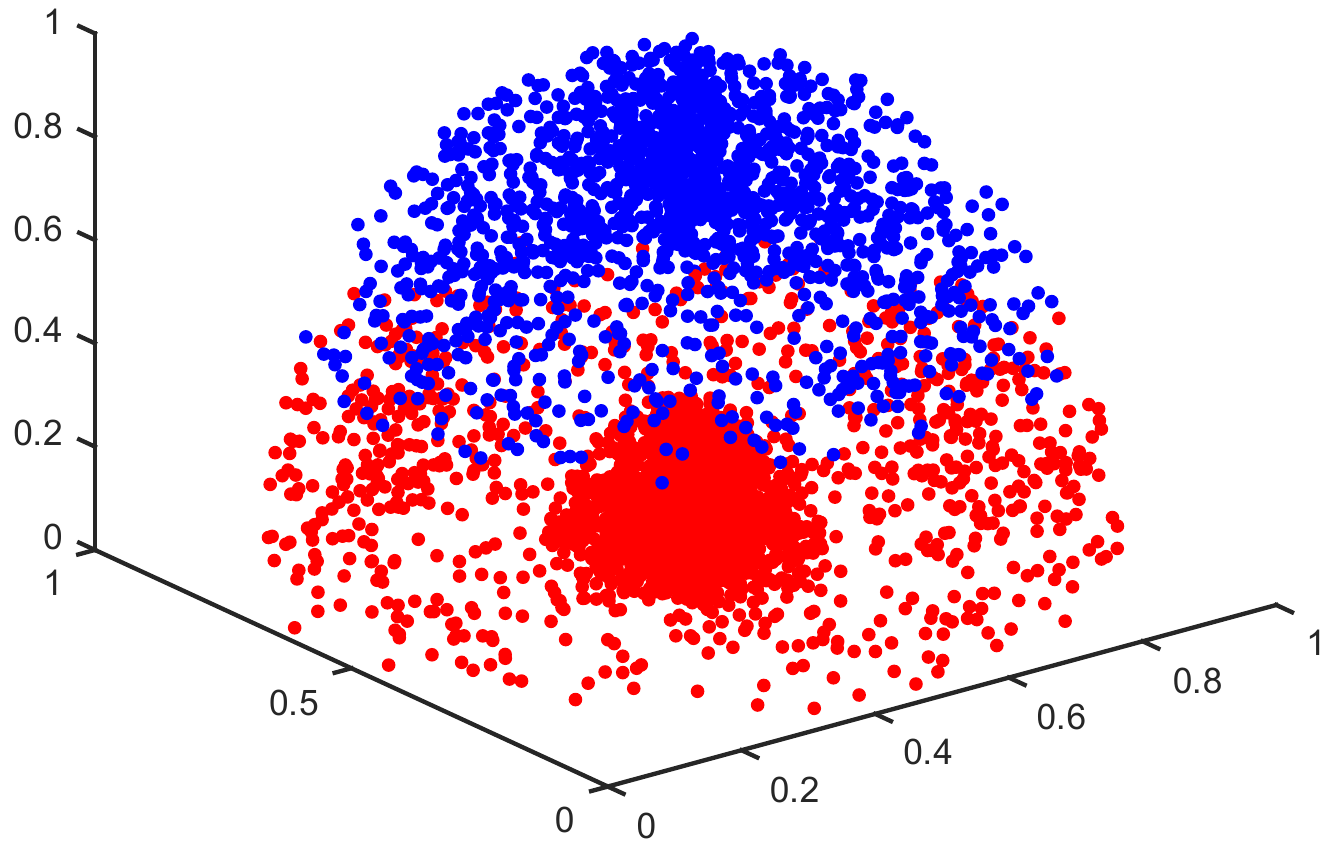}}
\vspace{3mm}
\centerline{\includegraphics[width=1\textwidth]{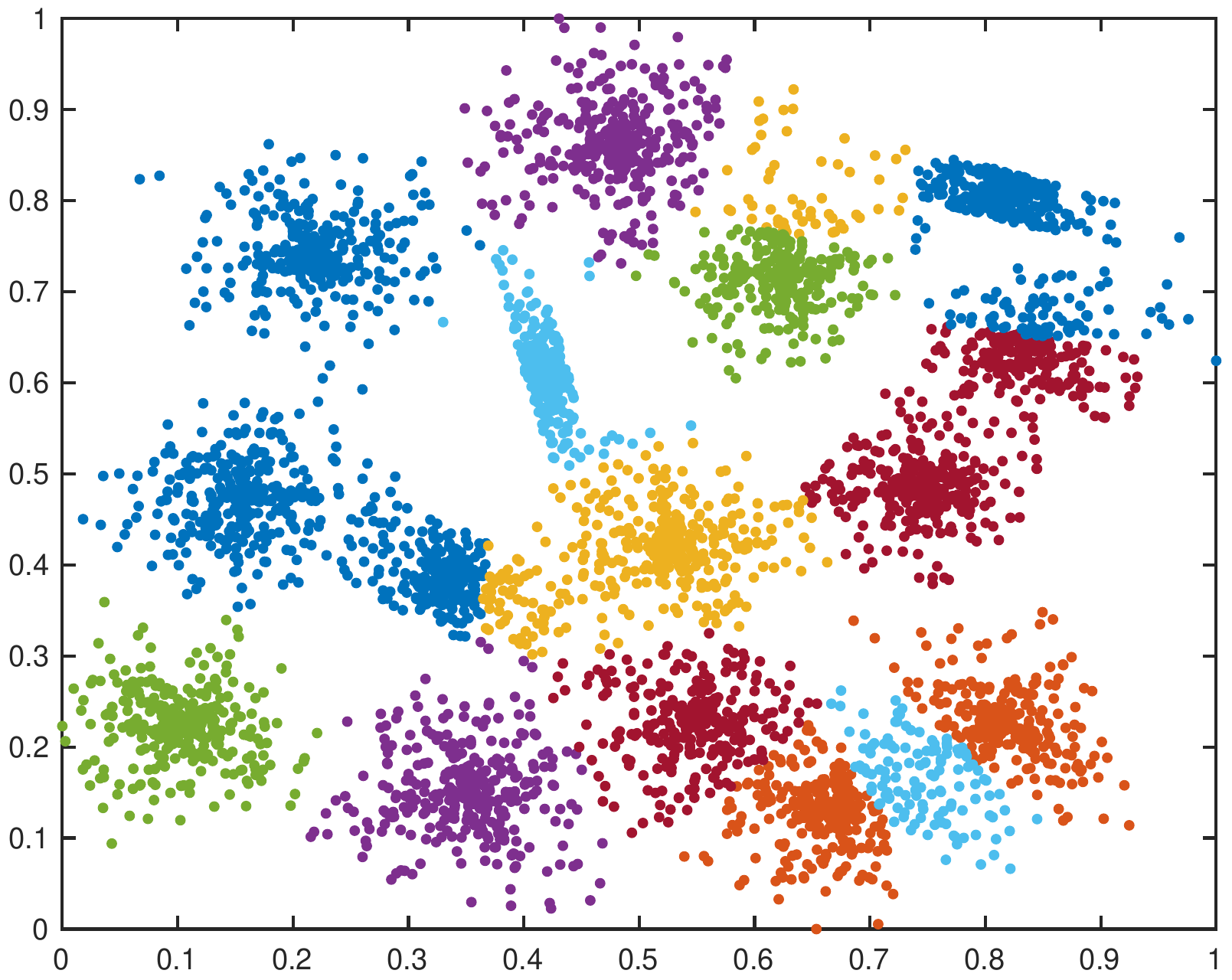}}
\vspace{3mm}
\centerline{\includegraphics[width=1\textwidth]{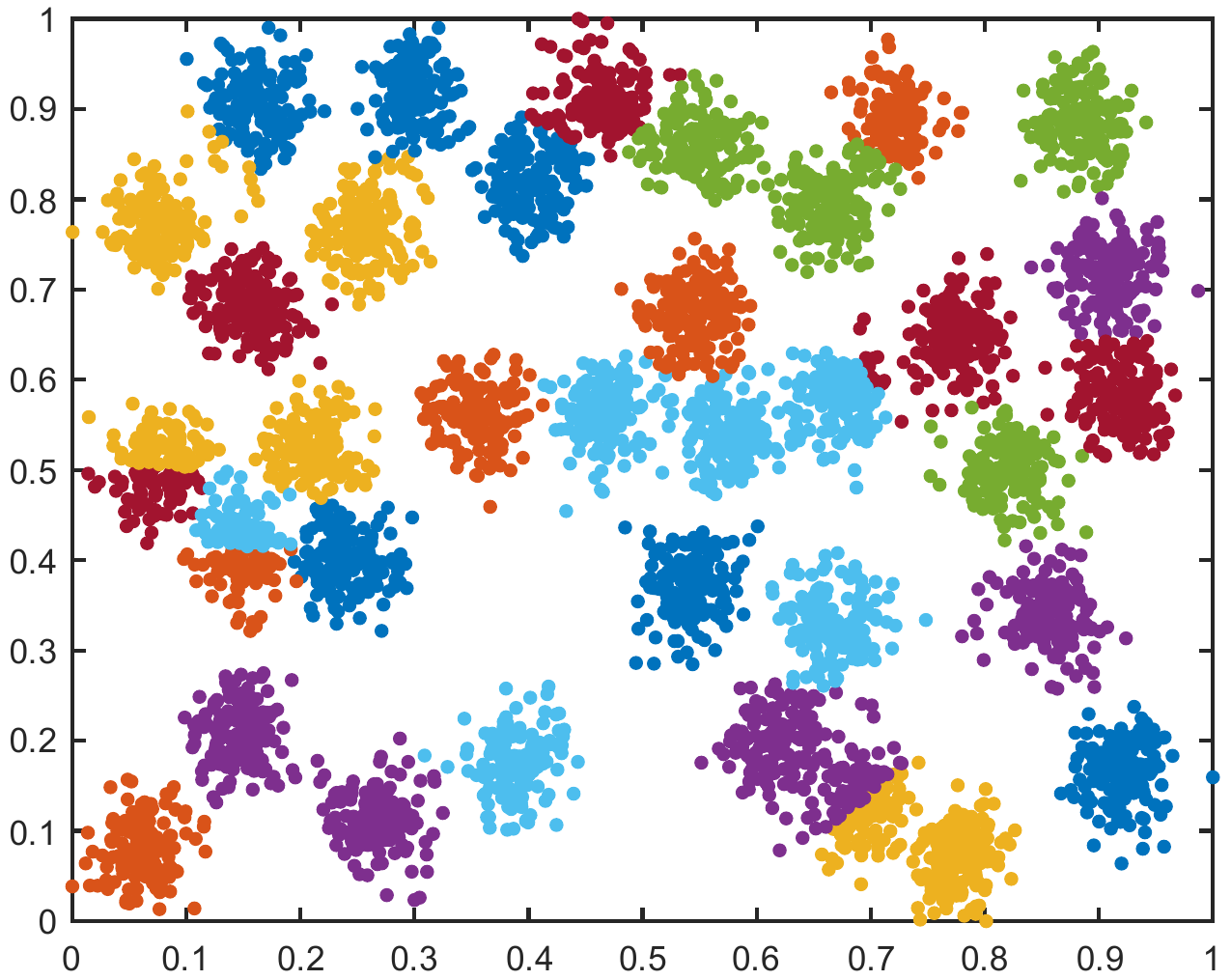}}
\centerline{(a) K-Means}
\end{minipage}~~~~~
\begin{minipage}[t]{0.17\textwidth}
\centerline{\includegraphics[width=1\textwidth]{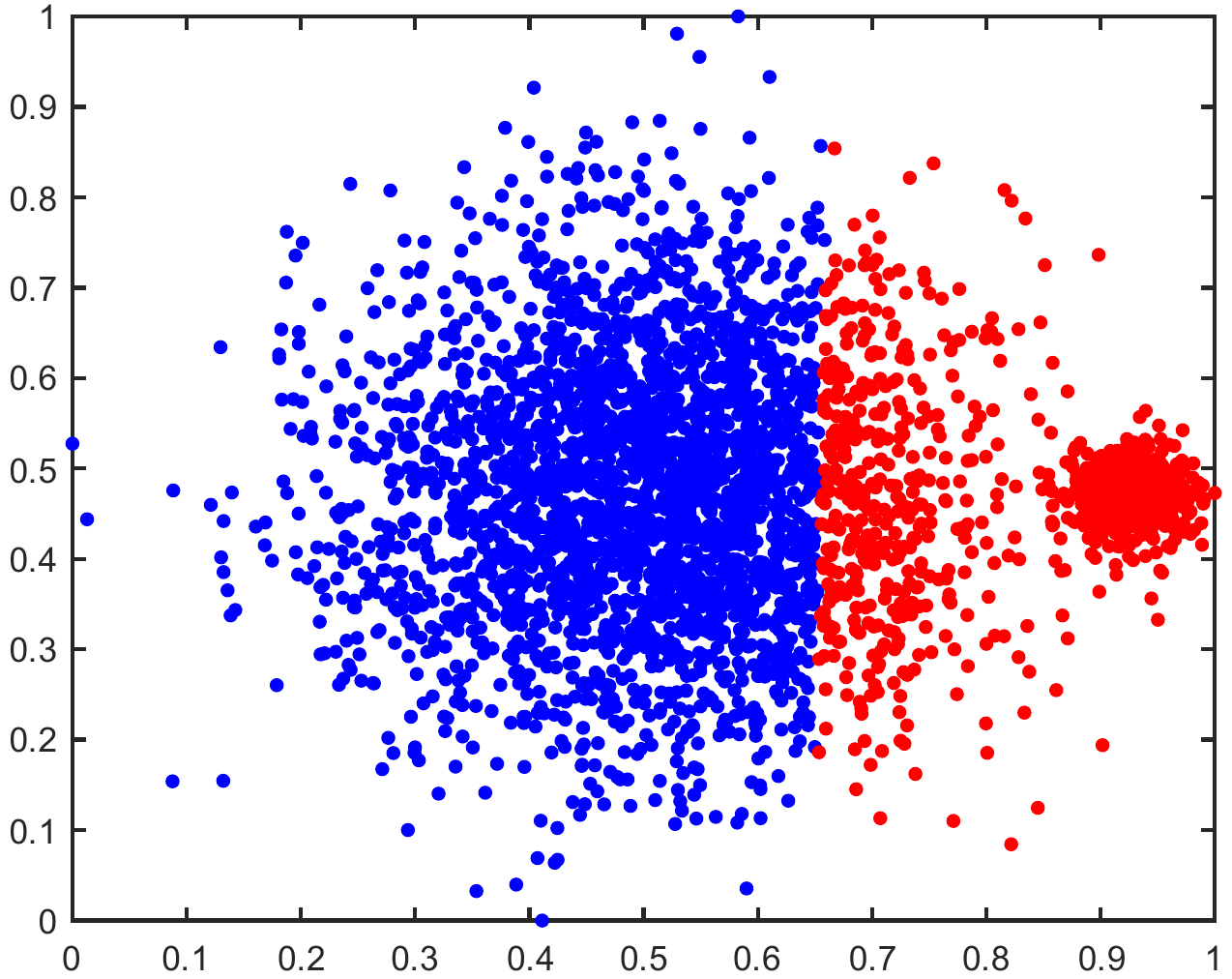}}
\vspace{3mm}
\centerline{\includegraphics[width=1\textwidth]{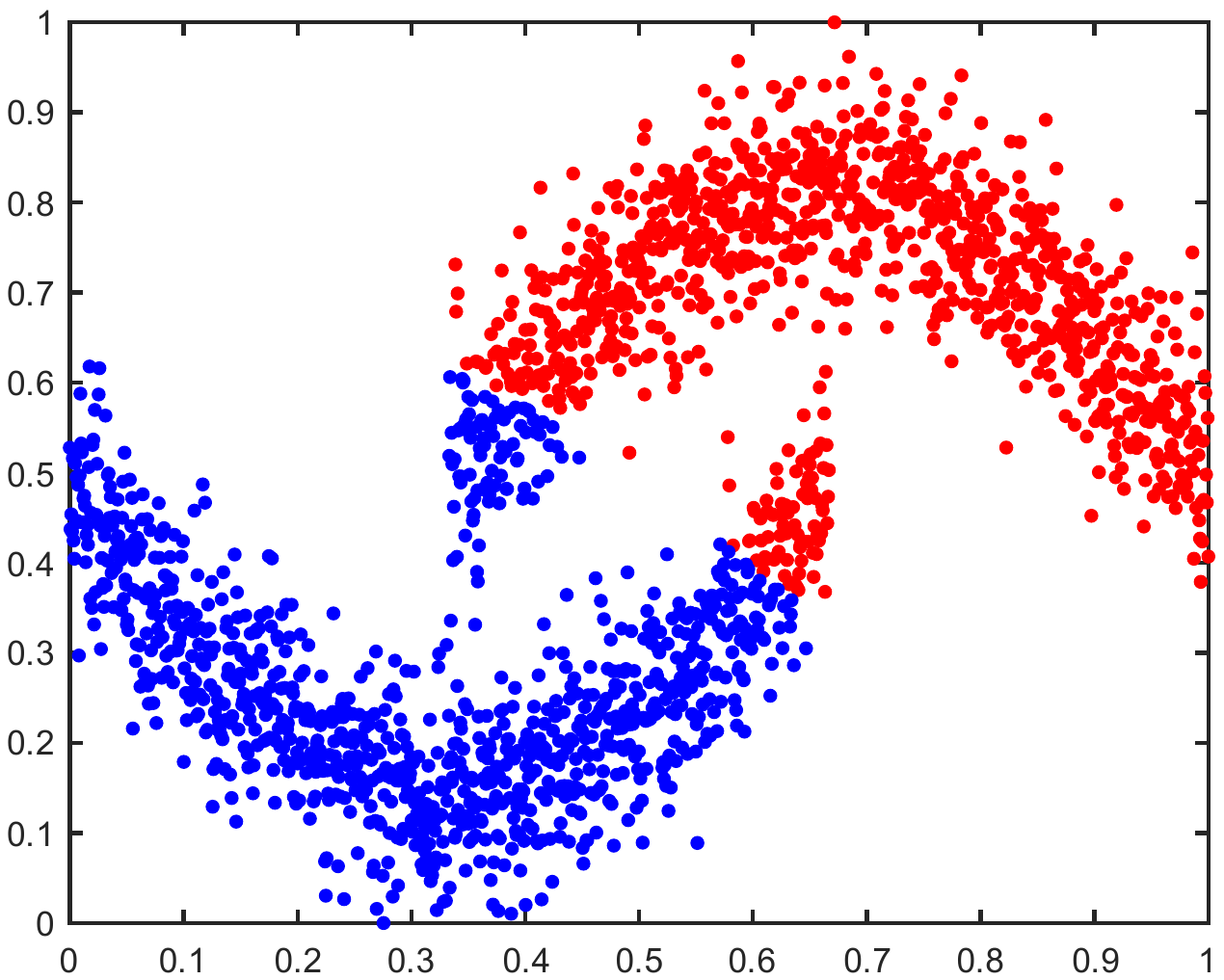}}
\vspace{3mm}
\centerline{\includegraphics[width=1\textwidth]{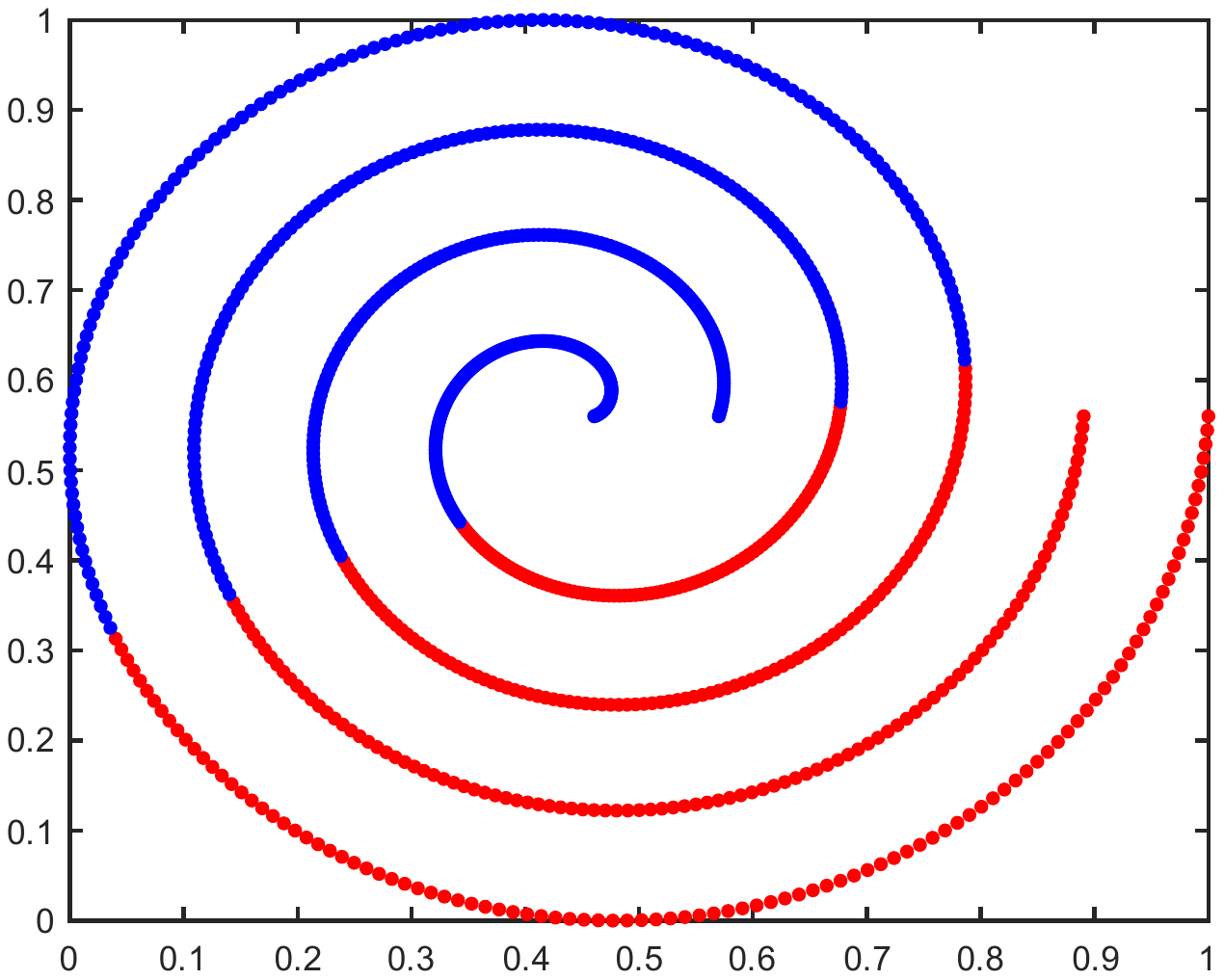}}
\vspace{3mm}
\centerline{\includegraphics[width=1\textwidth]{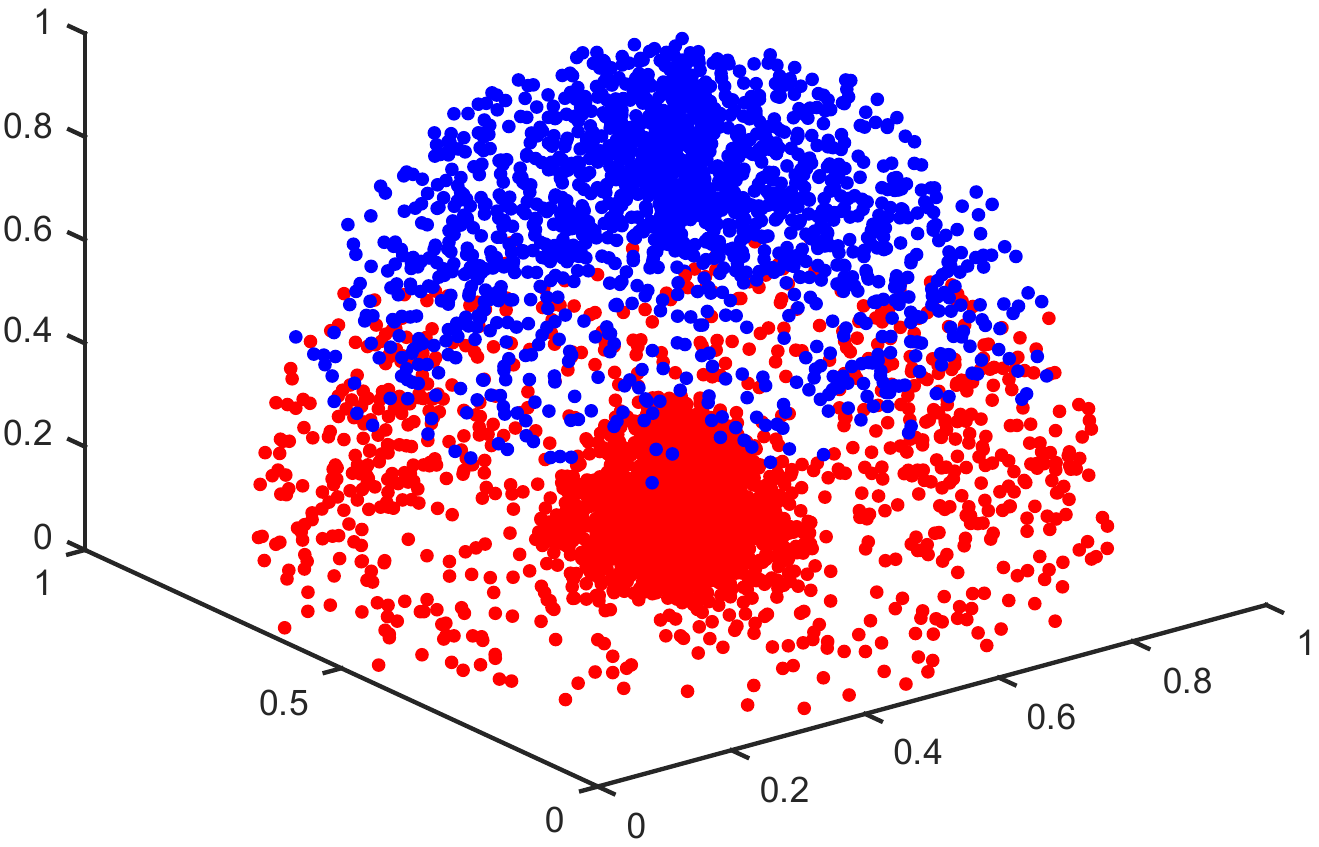}}
\vspace{3mm}
\centerline{\includegraphics[width=1\textwidth]{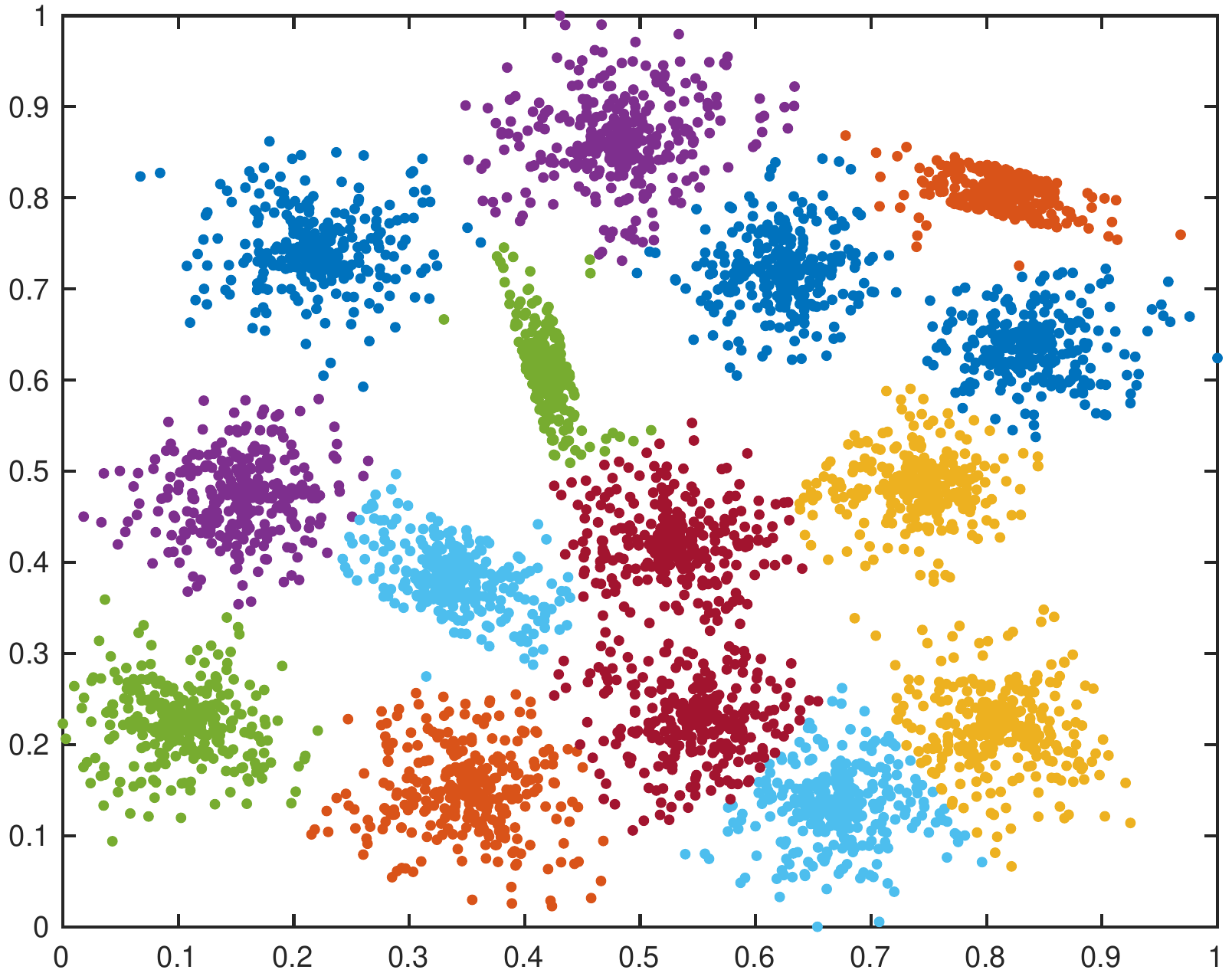}}
\vspace{3mm}
\centerline{\includegraphics[width=1\textwidth]{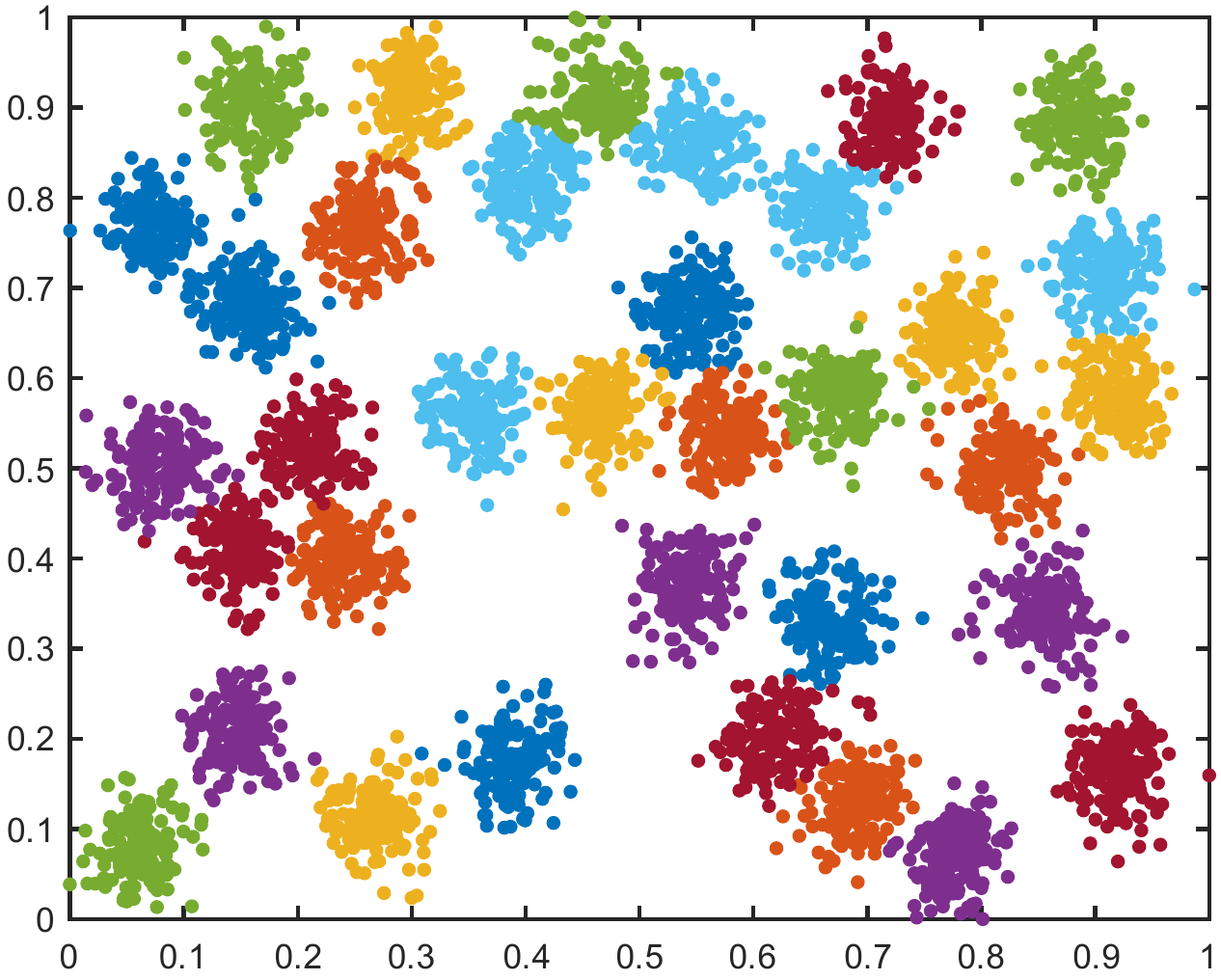}}
\centerline{(b) SMKM}
\end{minipage}~~~~~
\begin{minipage}[t]{0.17\textwidth}
\centerline{\includegraphics[width=1\textwidth]{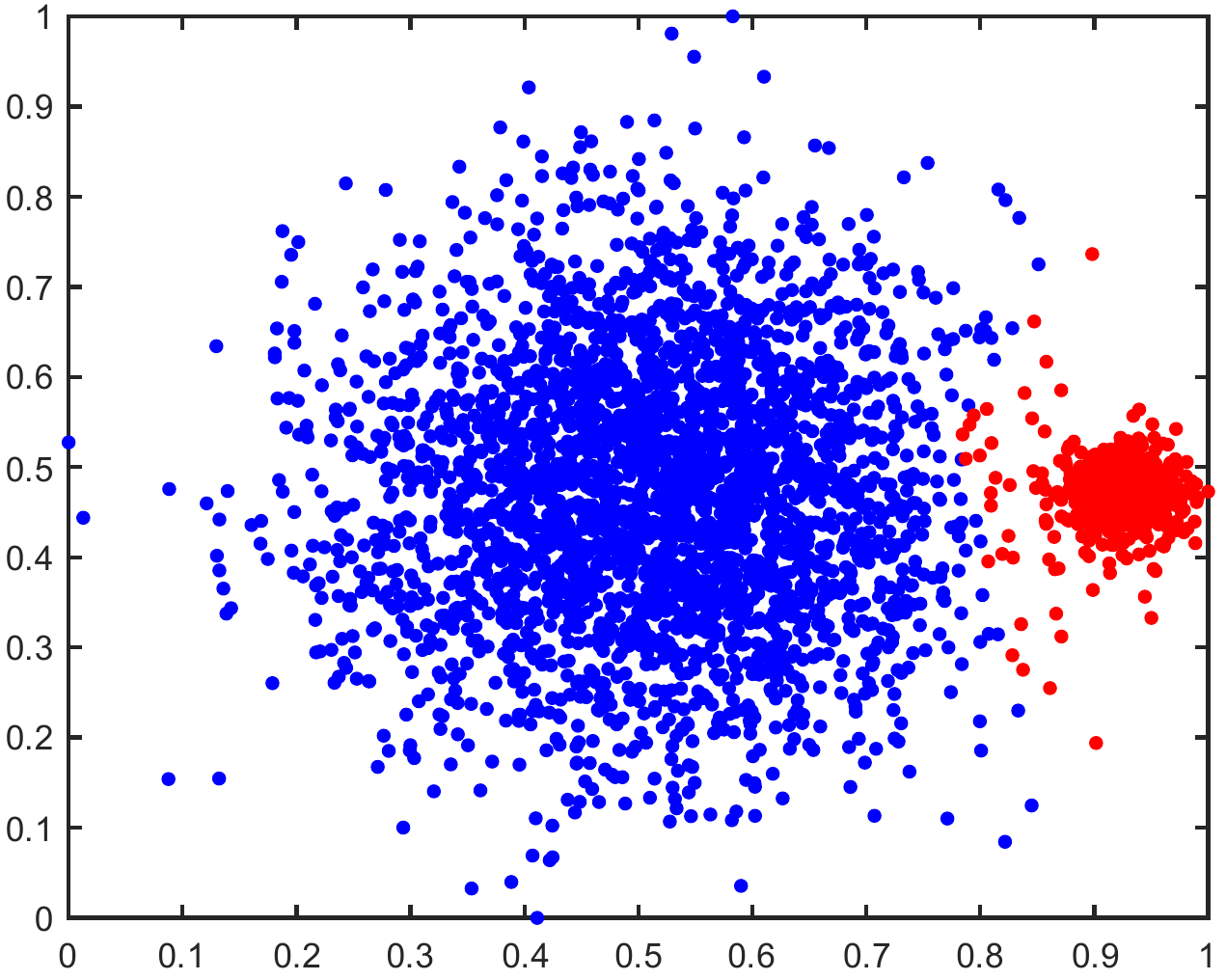}}
\vspace{3mm}
\centerline{\includegraphics[width=1\textwidth]{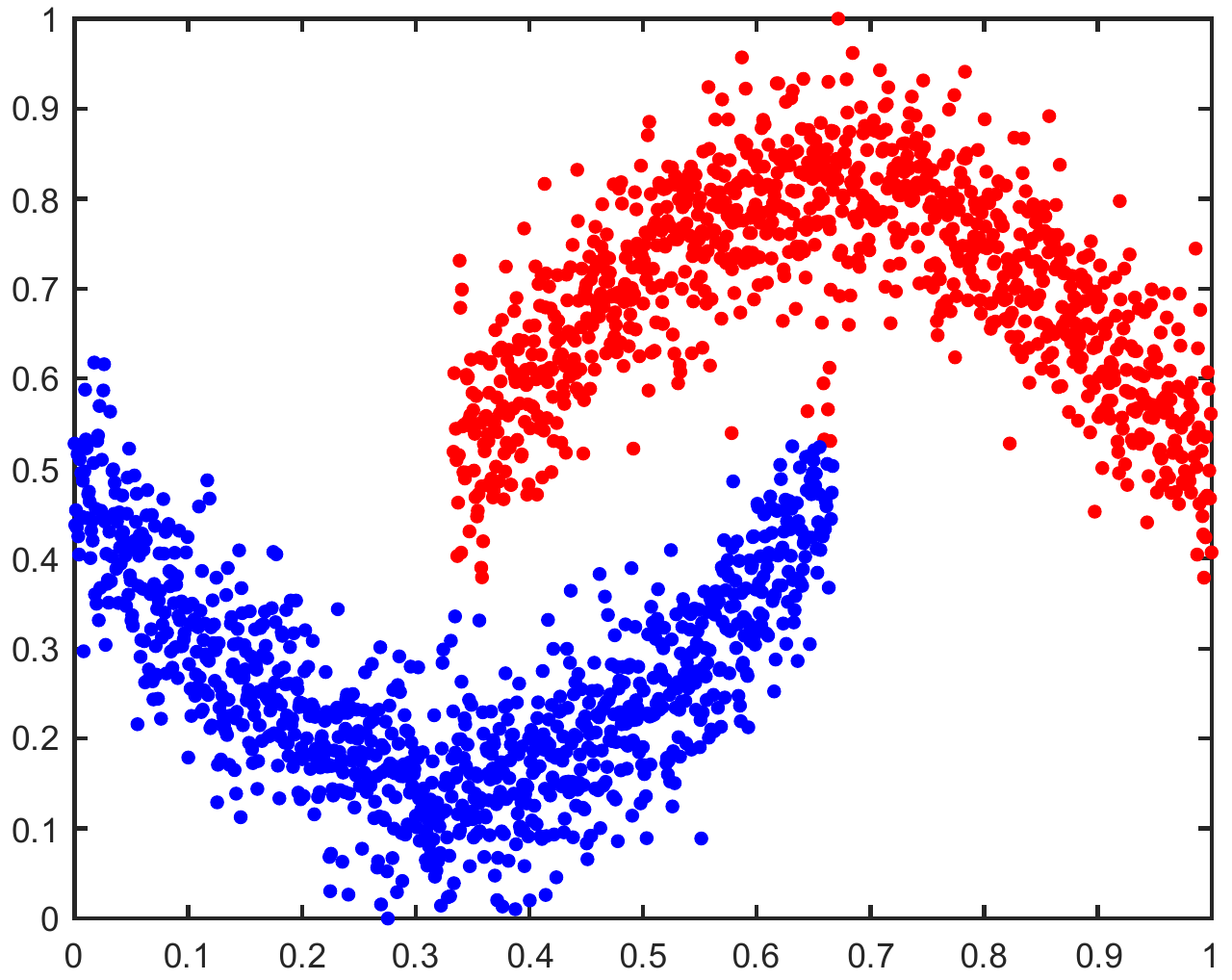}}
\vspace{3mm}
\centerline{\includegraphics[width=1\textwidth]{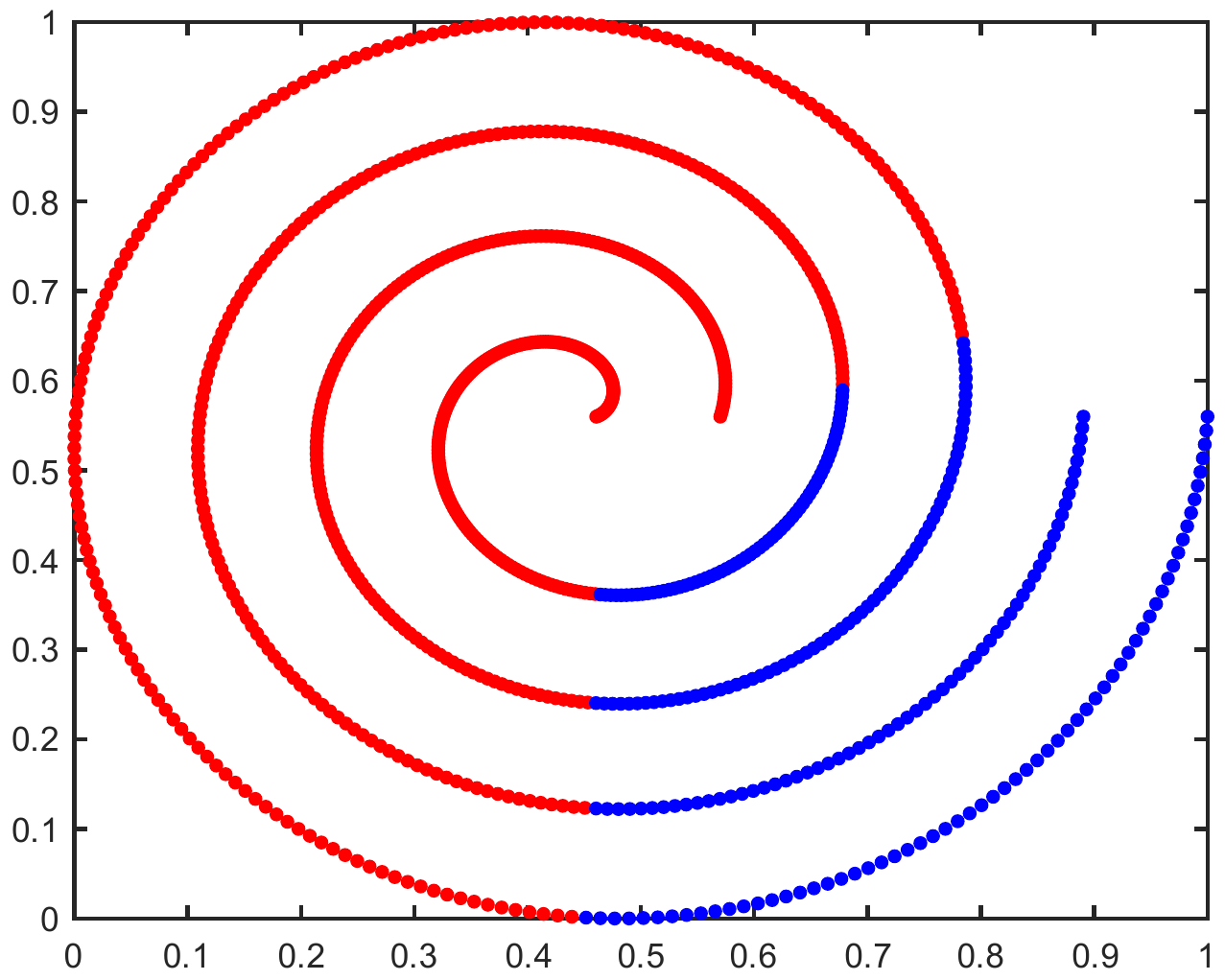}}
\vspace{3mm}
\centerline{\includegraphics[width=1\textwidth]{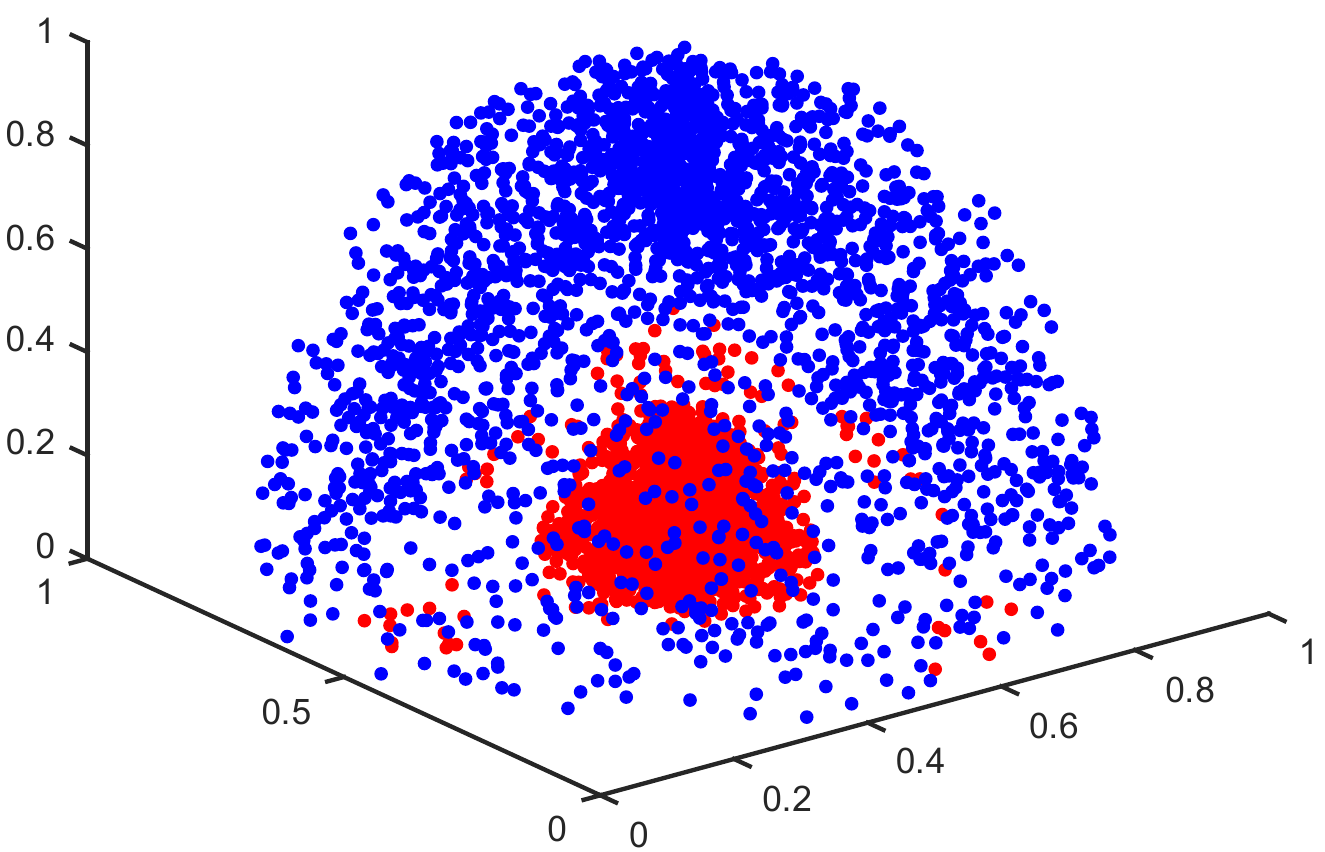}}
\vspace{3mm}
\centerline{\includegraphics[width=1\textwidth]{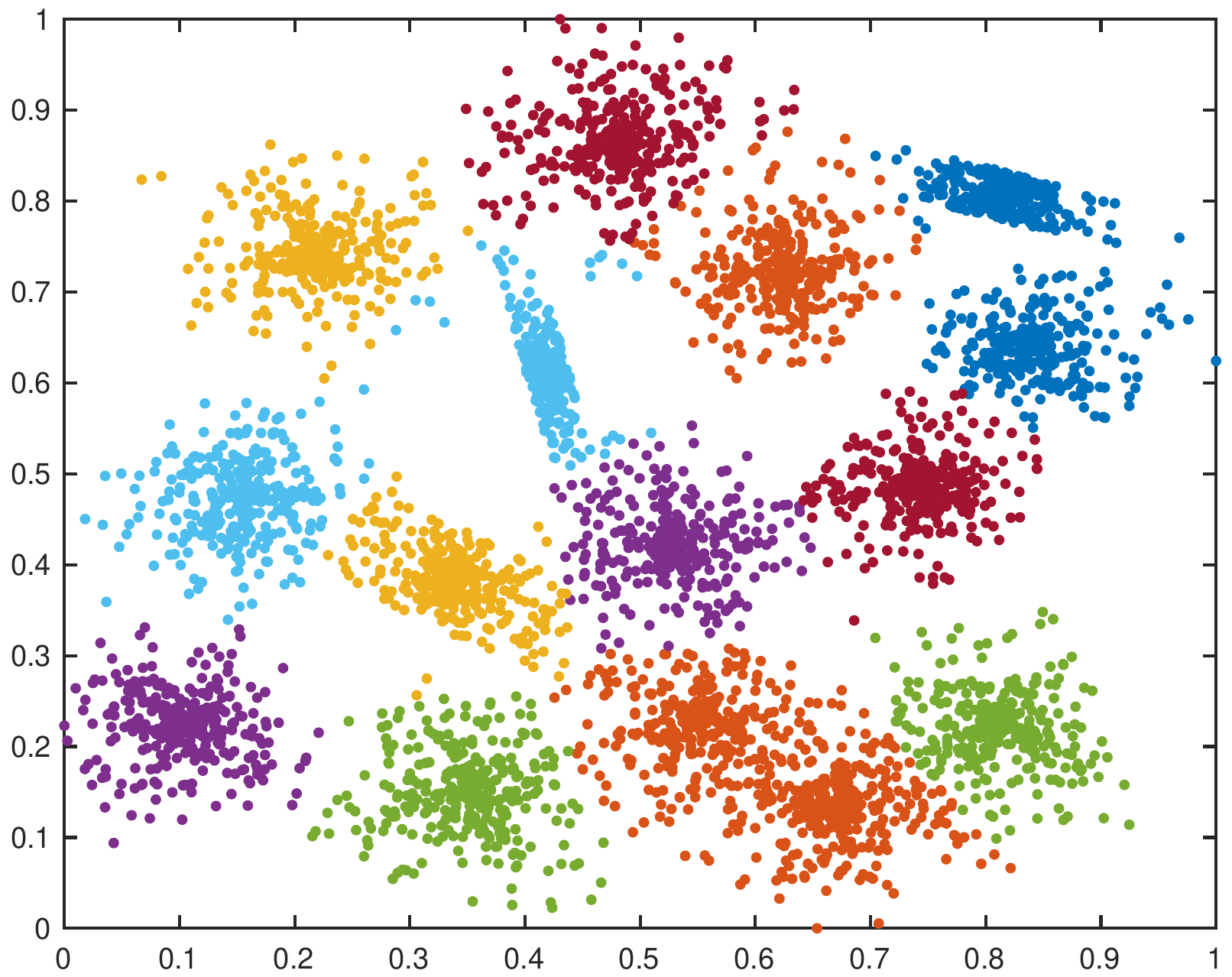}}
\vspace{3mm}
\centerline{\includegraphics[width=1\textwidth]{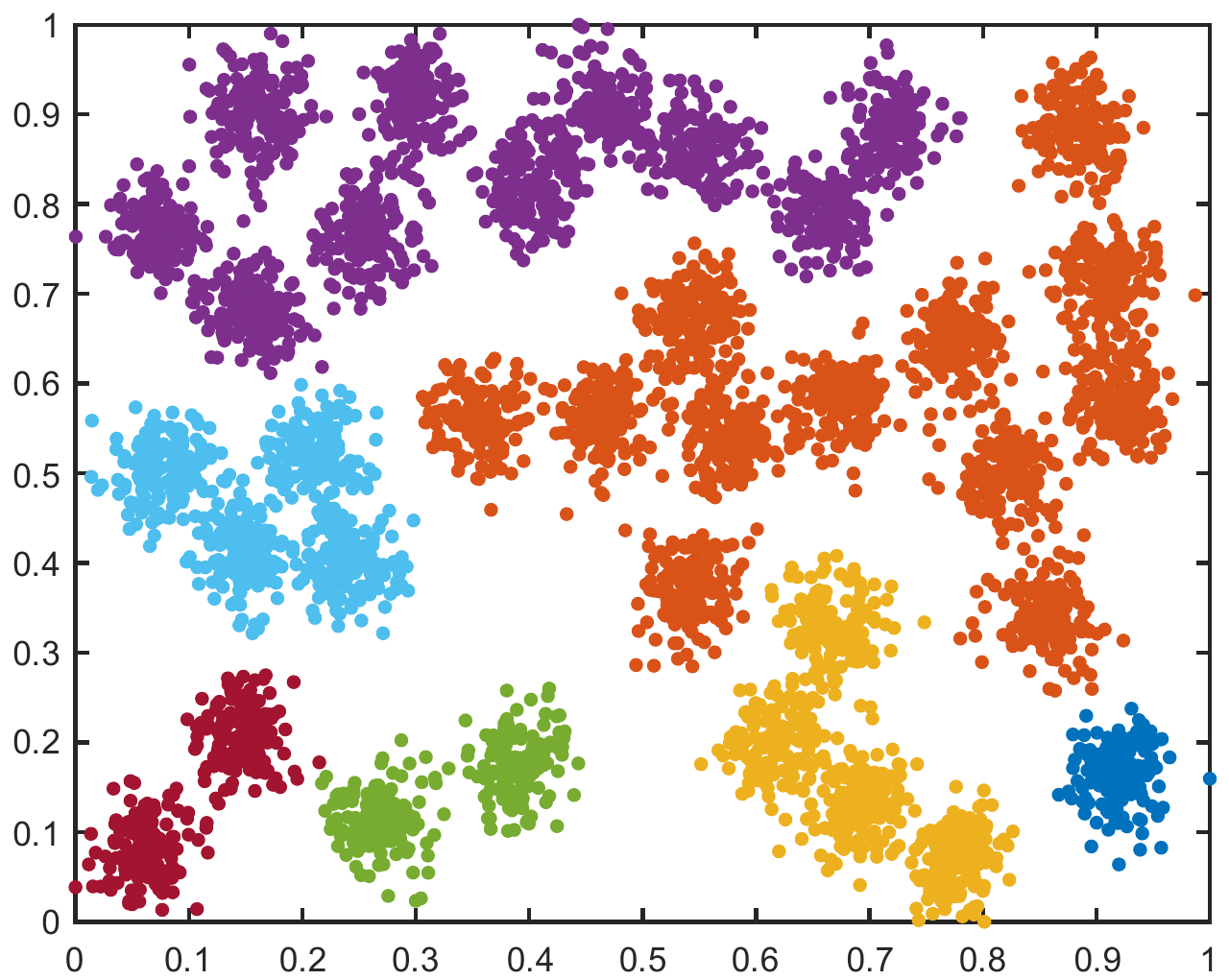}}
\centerline{(c) SMCL}
\end{minipage}~~~~~
\begin{minipage}[t]{0.17\textwidth}
\centerline{\includegraphics[width=1\textwidth]{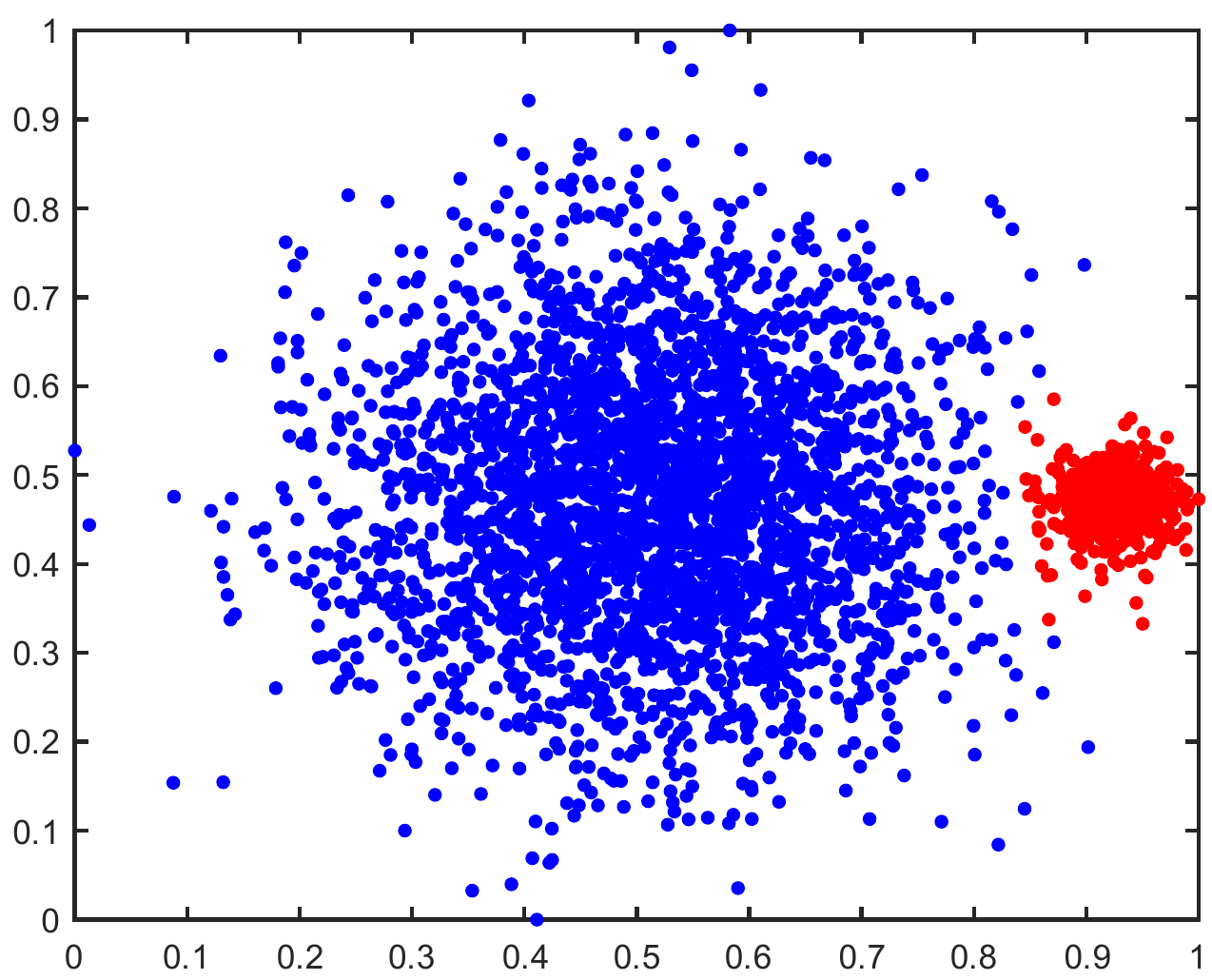}}
\vspace{3mm}
\centerline{\includegraphics[width=1\textwidth]{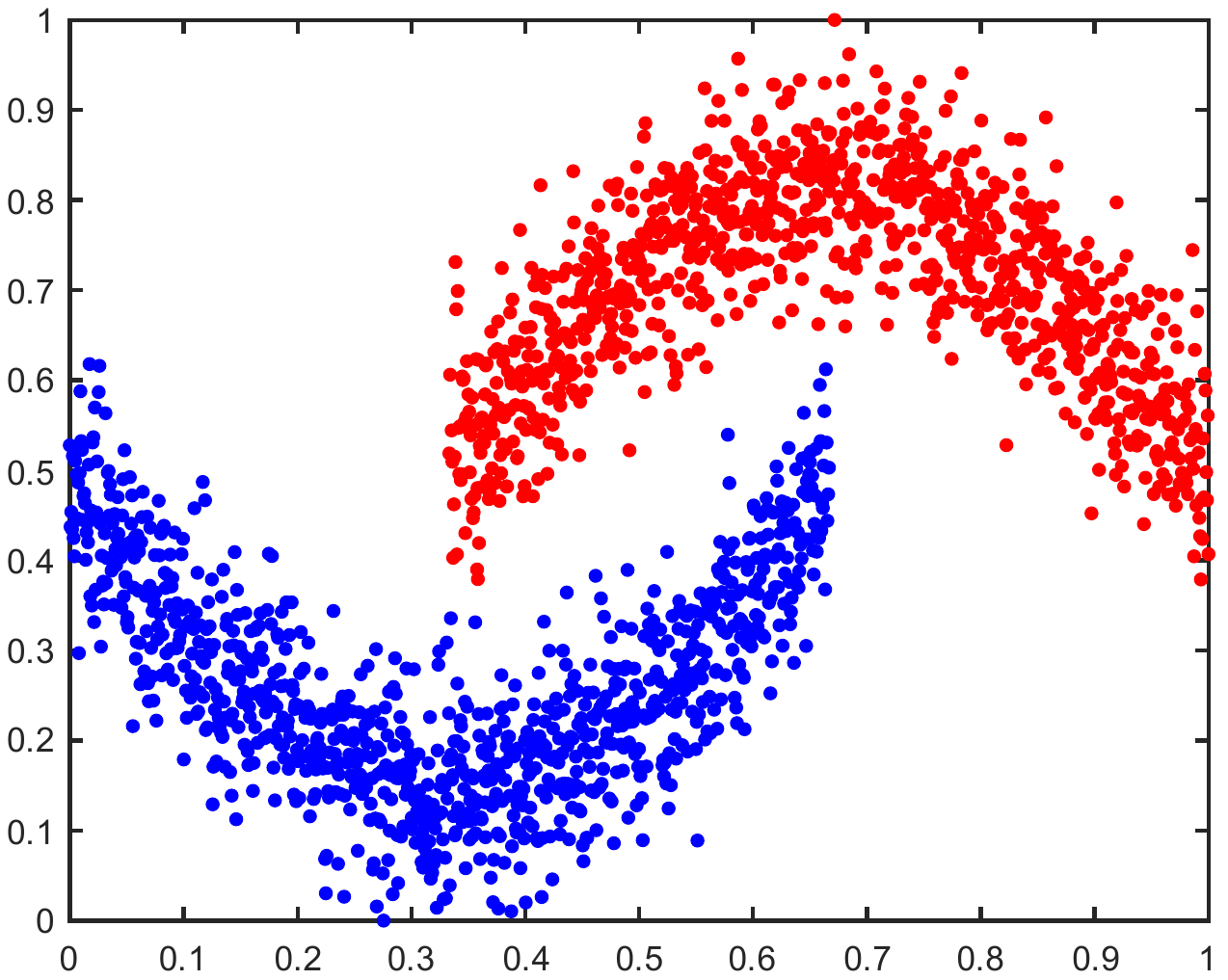}}
\vspace{3mm}
\centerline{\includegraphics[width=1\textwidth]{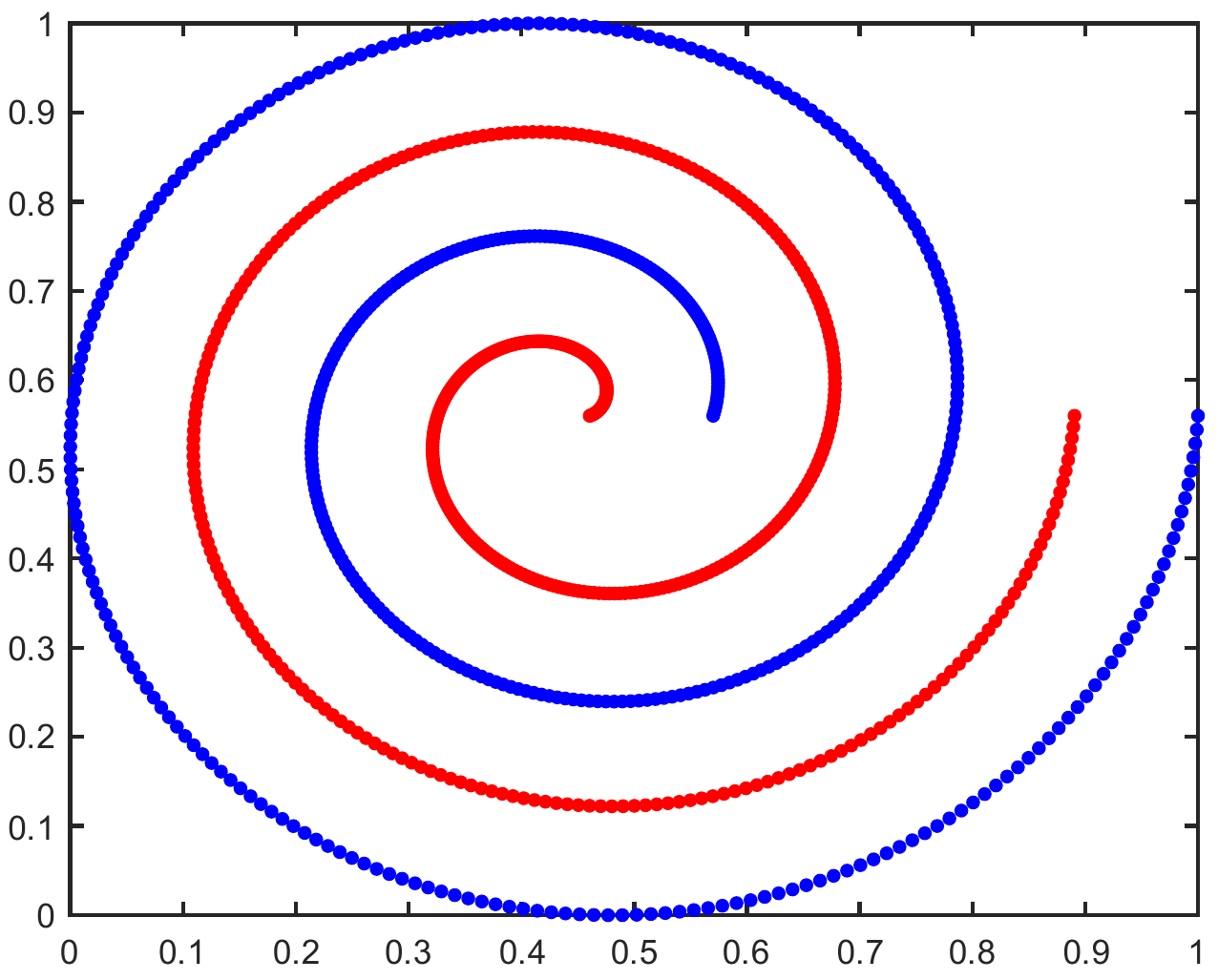}}
\vspace{3mm}
\centerline{\includegraphics[width=1\textwidth]{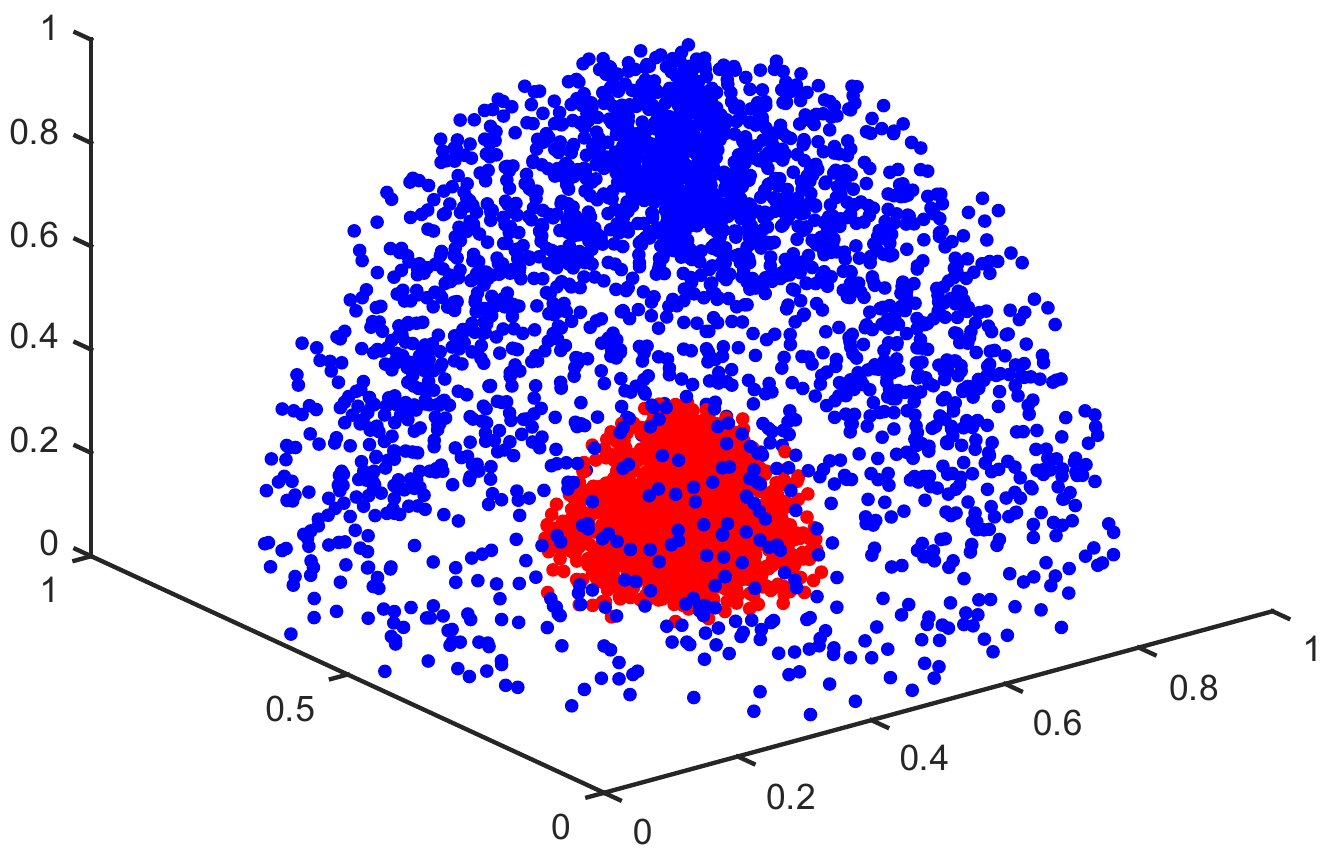}}
\vspace{3mm}
\centerline{\includegraphics[width=1\textwidth]{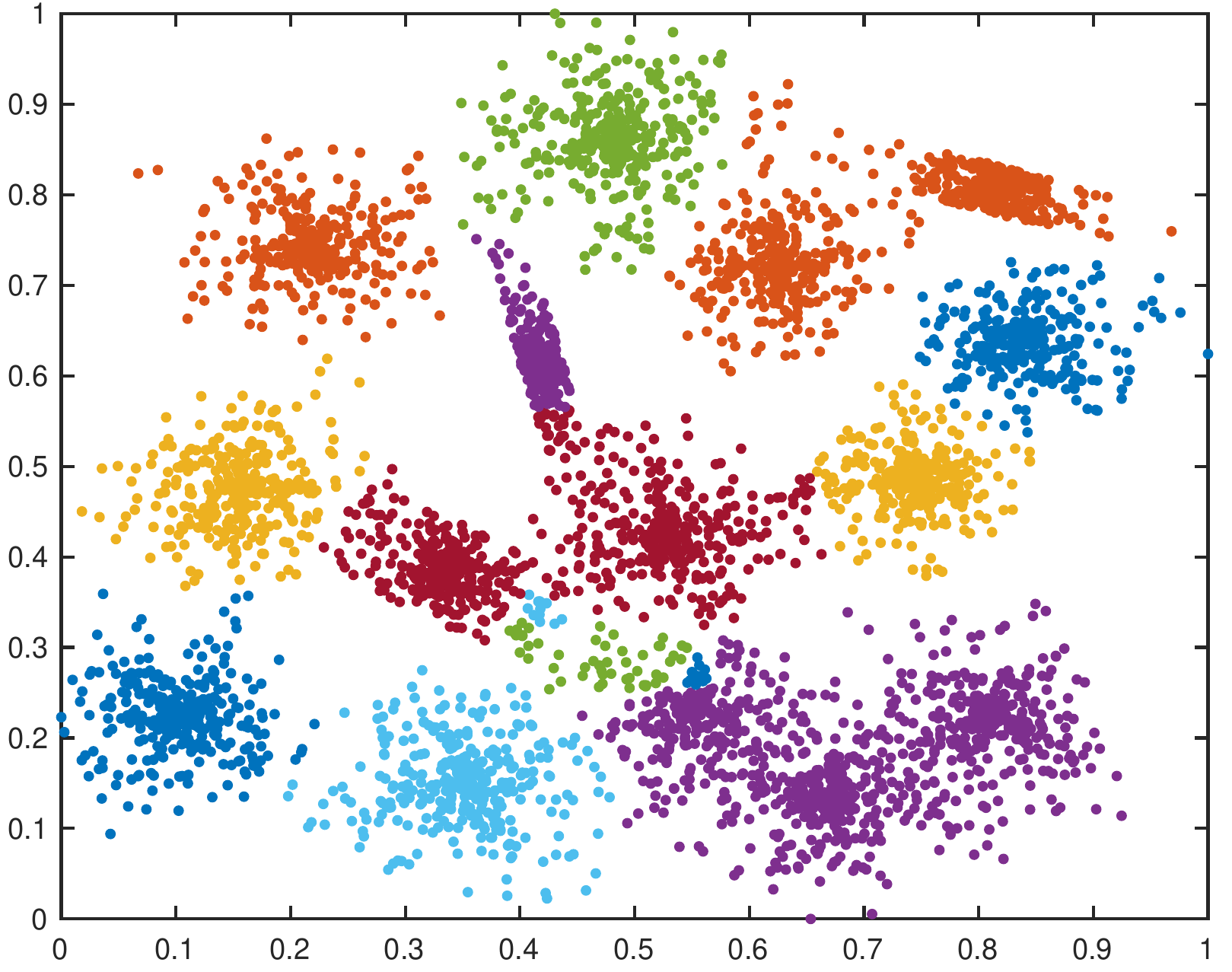}}
\vspace{3mm}
\centerline{\includegraphics[width=1\textwidth]{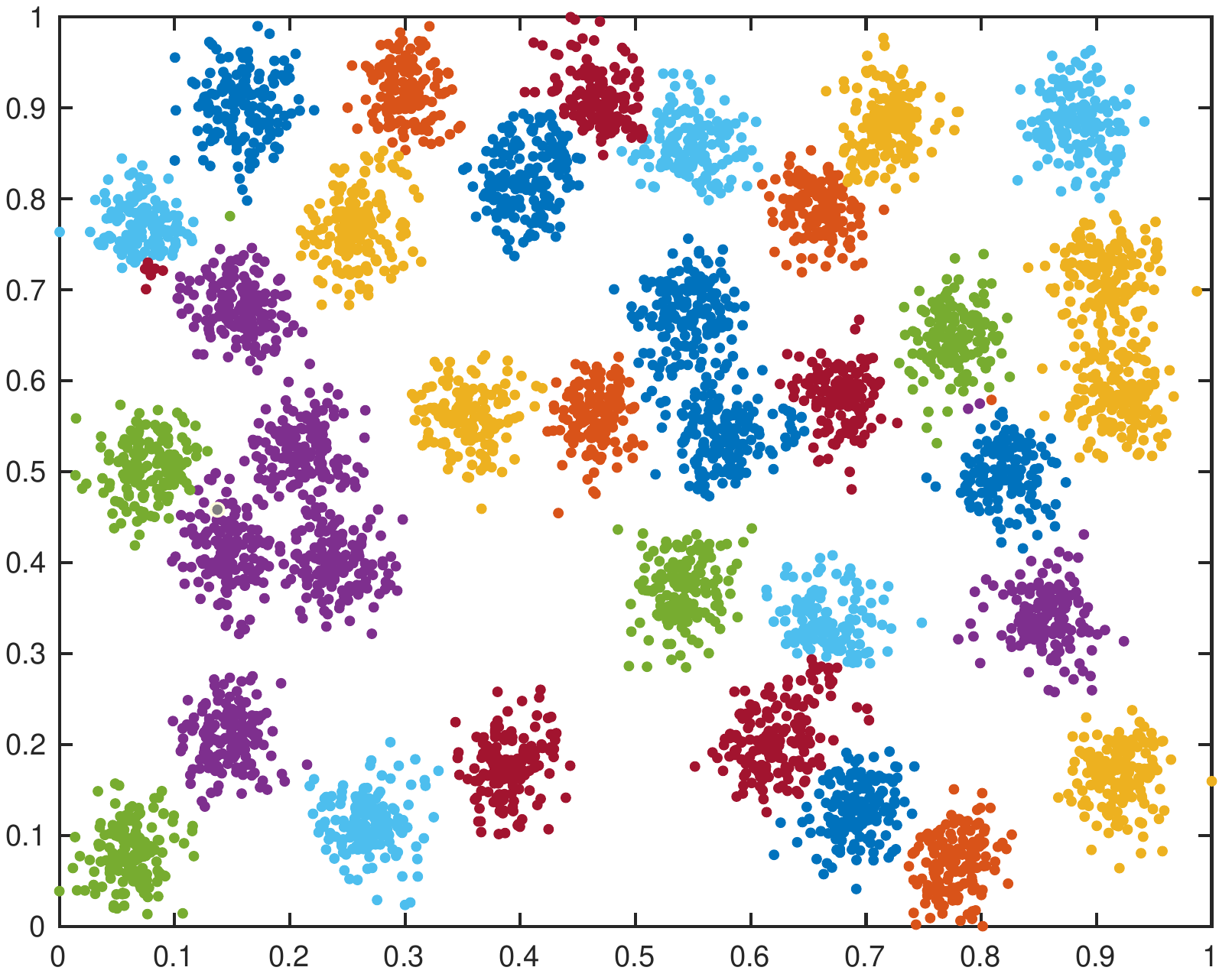}}
\centerline{(d) CC}
\end{minipage}~~~~~
\begin{minipage}[t]{0.17\textwidth}
\centerline{\includegraphics[width=1\textwidth]{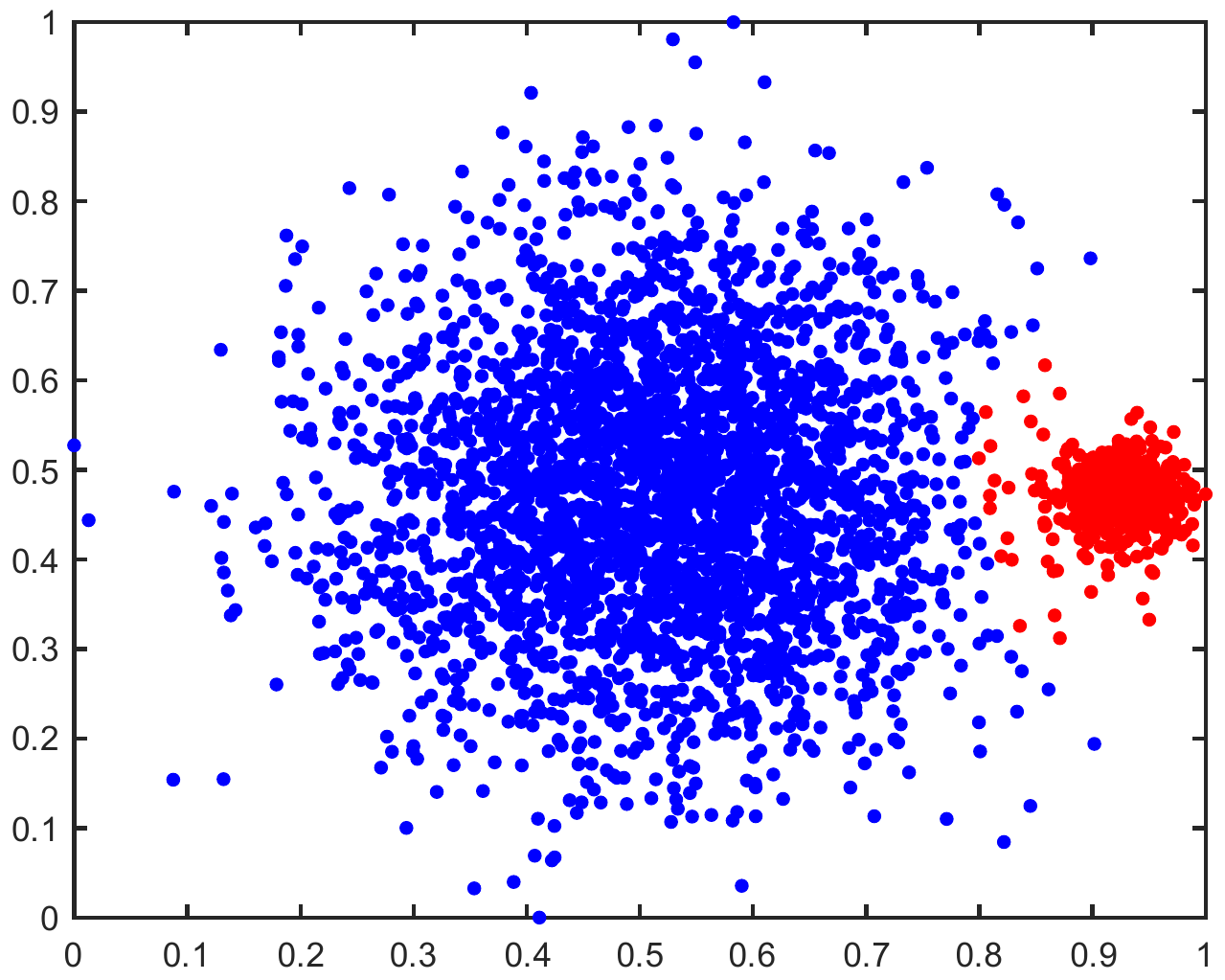}}
\vspace{3mm}
\centerline{\includegraphics[width=1\textwidth]{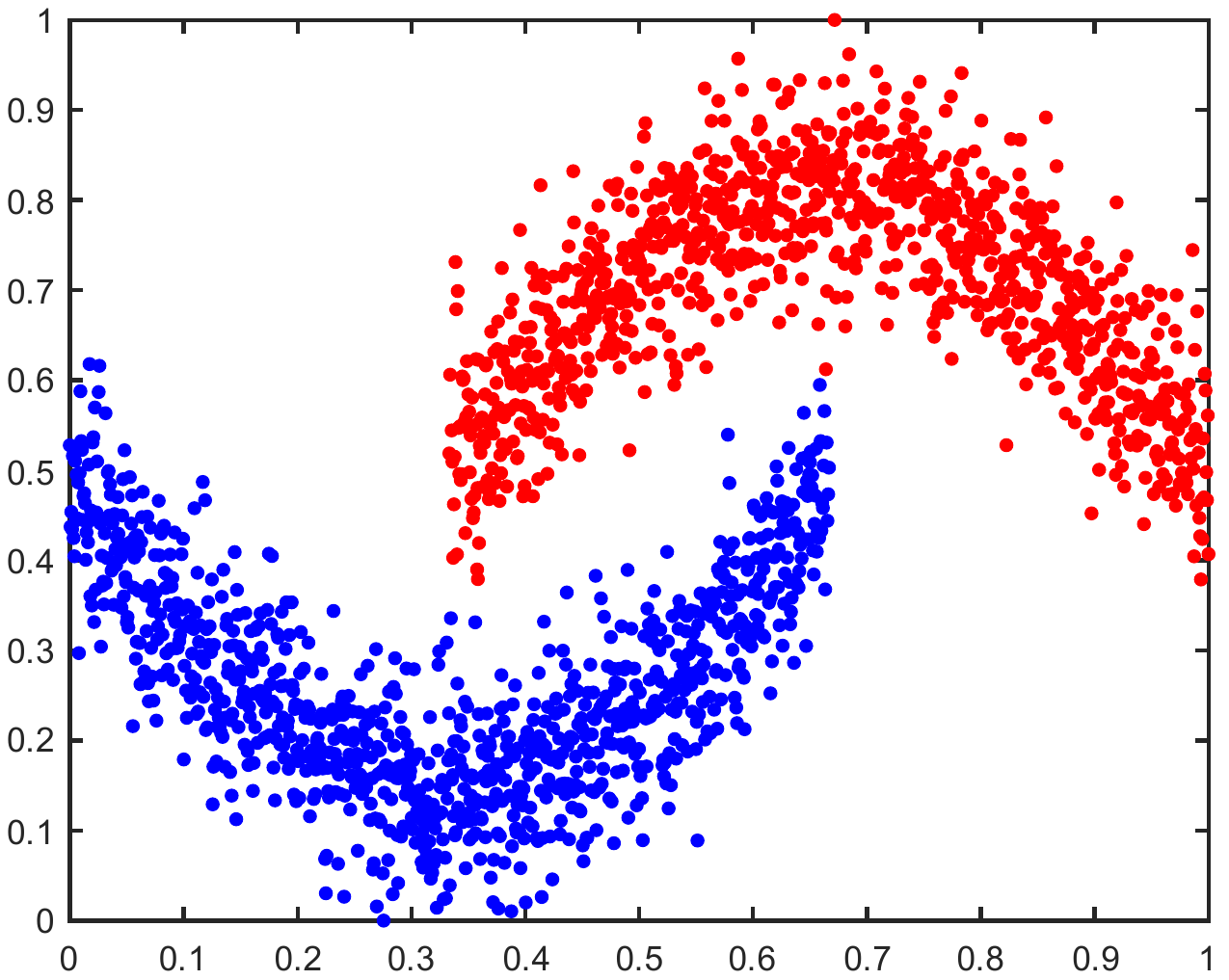}}
\vspace{3mm}
\centerline{\includegraphics[width=1\textwidth]{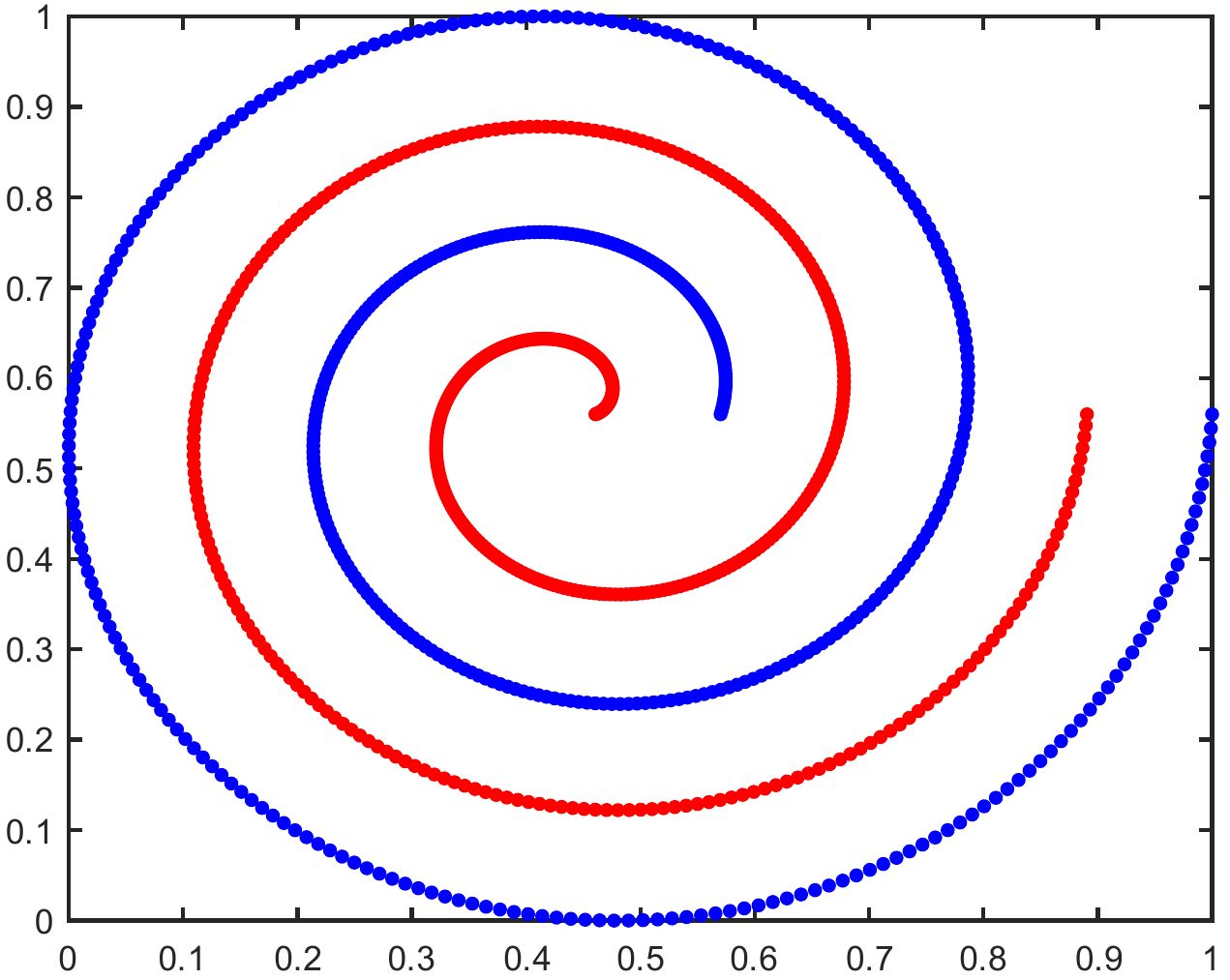}}
\vspace{3mm}
\centerline{\includegraphics[width=1\textwidth]{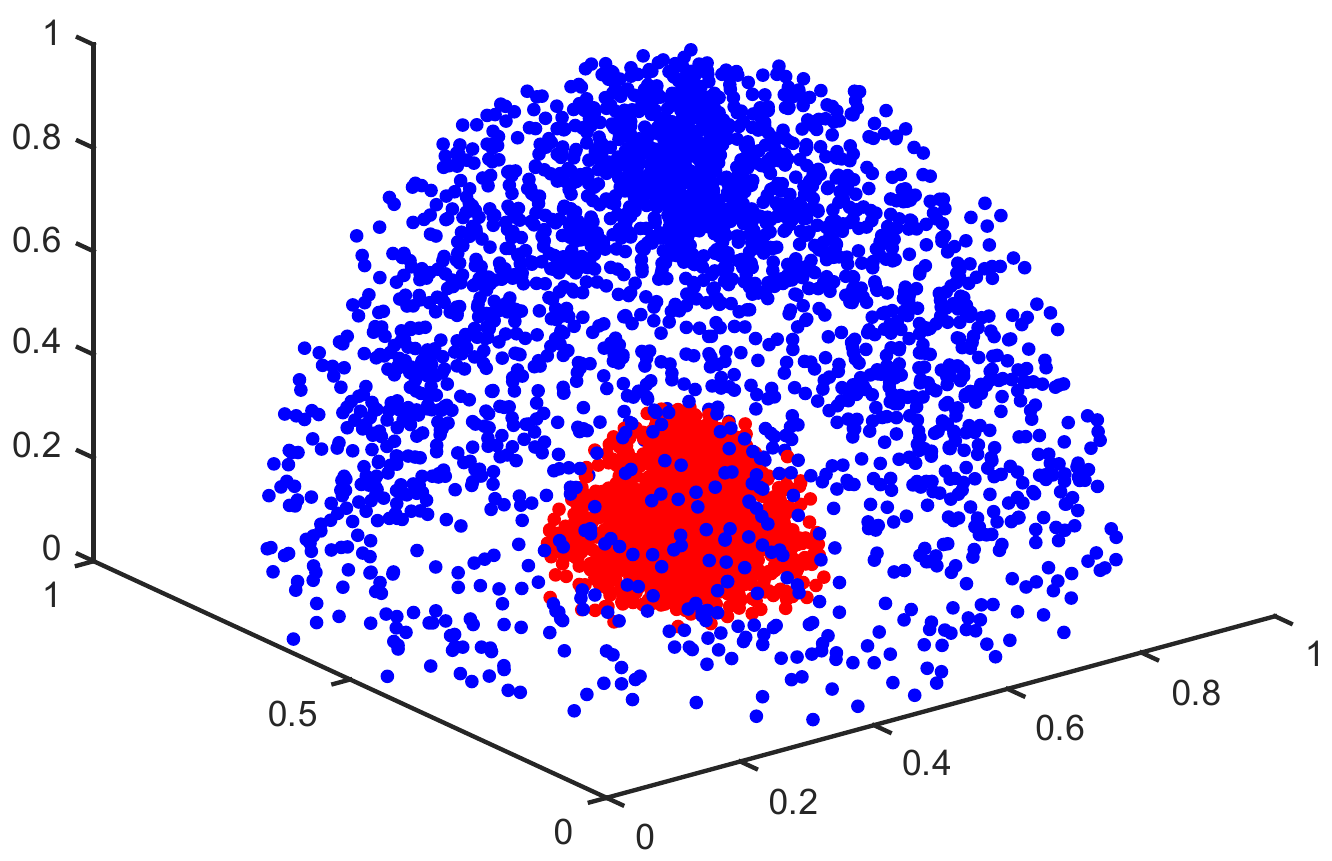}}
\vspace{3mm}
\centerline{\includegraphics[width=1\textwidth]{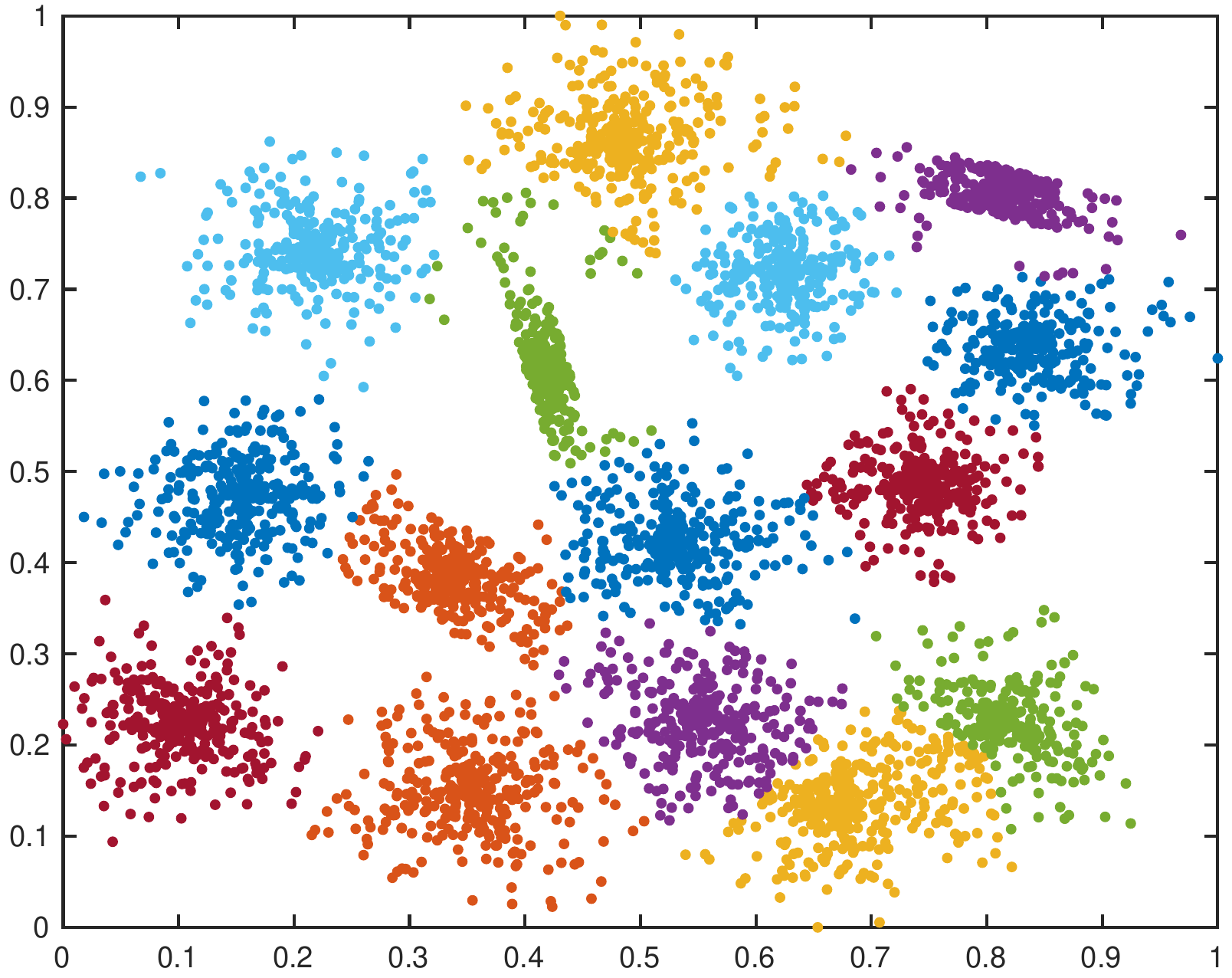}}
\vspace{3mm}
\centerline{\includegraphics[width=1\textwidth]{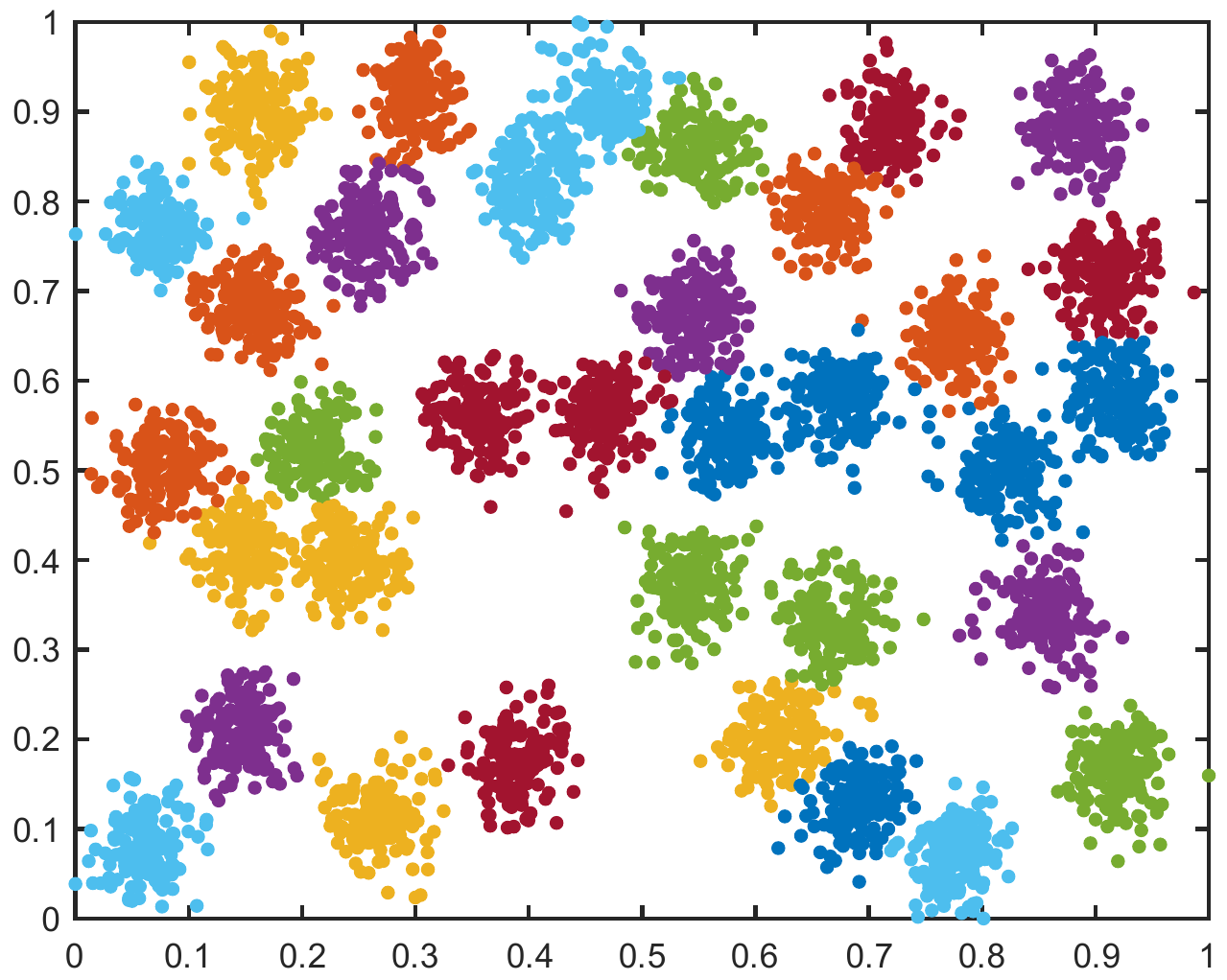}}
\centerline{(e) MCKM}
\end{minipage}
\caption{The clustering results of K-Means \cite{Least1982},  SMKM \cite{Capo2022}, SMCK \cite{lu2019}, CC \cite{Lindsten2011} and the proposed MCKM on the six synthetic data sets.}\label{fig7}
\end{figure*}

After MPS,  CM  is applied to merge the multi-prototypes to get the local minima of K-Means problem. The clustering results of MCKM (MPS+CM) and of the other four algorithms are plotted in Fig. \ref{fig7}. The metric results of $\textbf{F}^{*}$, \textbf{NMI}, and \textbf{ARI} are shown in Table \ref{table2}, where {$k^{*}$} is the number of clusters obtained by the algorithms. Suitable hyper-parameters are selected for SMCL, CC and MCKM and are listed in the second column of Table \ref{table2}. Furthermore,  the running times of the algorithms are displayed in Table \ref{table3}, where the values are averaged over 20 trials.

\begin{table}[htb]
\centering
\caption{The evaluation of the clustering results of  the different algorithms on the six synthetic data sets. The best results are shown in boldface.}
\label{table2}
\begin{tabular}{r|l|cccc}
\toprule
\multicolumn{1}{c|}{\multirow{2}{*}{Algorithms}}
&\multicolumn{1}{c|}{\multirow{1}{*}{Parameter}}
&\multicolumn{1}{c}{\multirow{1}{*}{$\textrm{F}^{*}$}}
&\multicolumn{1}{c}{\multirow{1}{*}{NMI}}
&\multicolumn{1}{c}{\multirow{1}{*}{ARI}}
&\multicolumn{1}{c}{\multirow{1}{*}{$k^{*}$}}
\\
\cmidrule(r){2-6}
&\multicolumn{5}{c}{\multirow{1}{*}{D1 ($n=3500, p=2, k=2$)}}
\\
\midrule
K-Means&$k=2$ &0.8574&0.4481&0.4999&$-$ \\
SMKM&$k=2$ &0.8574&0.4481&0.4999&$-$ \\
SMCL& $\alpha_{c}=\eta=0.005$ &0.9878&0.8748&0.9373&\textbf{2}\\
CC&$q=5$; $\gamma=5$ &\textbf{0.9939}&\textbf{0.9267}&\textbf{0.9681}&\textbf{2} \\
MCKM&$q=2$; $\gamma=0.205$ &0.9899&0.8909&0.9475&\textbf{2}\\

\midrule
&\multicolumn{5}{c}{\multirow{1}{*}{D2 ($n=2000, p=2, k=2$)}}
\\
\midrule
K-Means&$k=2$  &0.9160&0.5839&0.6921&$-$\\
SMKM&$k=2$ &0.9160&0.5839&0.6921&$-$\\
SMCL&$\alpha_{c}=\eta=0.005$ &0.9965&0.9669&0.9860&\textbf{2}\\
CC&$q=5$; $\gamma=20$  &\textbf{1.0000}&\textbf{1.0000}&\textbf{1.0000}&\textbf{2}\\
MCKM&$q=2$; $\gamma=1.7$ &0.9995&0.9943&0.9980&\textbf{2}\\

\midrule
&\multicolumn{5}{c}{\multirow{1}{*}{D3 ($n=1000, p=2, k=2$)}}
\\
\midrule
K-Means&$k=2$ &0.5180&0.0006&-0.0002&$-$ \\
SMKM&$k=2$  &0.5178&0.0006&-0.0002&$-$\\
SMCL&$\alpha_{c}=\eta=0.005$  &0.5386&0.0029&0.0027&2\\
CC&$q=5$; $\gamma=14$  &\textbf{1.0000}&\textbf{1.0000}&\textbf{1.0000}&\textbf{2}\\
MCKM&$q=2$; $\gamma=10$  &\textbf{1.0000}&\textbf{1.0000}&\textbf{1.0000}&\textbf{2}\\

\midrule
&\multicolumn{5}{c}{\multirow{1}{*}{D4 ($n=5000, p=3, k=2$)}}
\\
\midrule
K-Means&$k=2$&0.8338&0.4701&0.4318&$-$ \\
SMKM&$k=2$  &0.8338&0.4701&0.4318&$-$\\
SMCL & $\alpha_{c}=\eta=0.001$ &0.9988&0.9878&0.9952&\textbf{2}\\
CC&$q=5$; $\gamma=20$  &\textbf{1.0000}&\textbf{1.0000}&\textbf{1.0000}&\textbf{2}\\
MCKM&$q=2$; $\gamma=1.5$ &\textbf{1.0000}&\textbf{1.0000}&\textbf{1.0000}&\textbf{2} \\

\midrule
&\multicolumn{5}{c}{\multirow{1}{*}{D5 ($n=5000, p=2, k=15$)}}
\\
\midrule
K-Means &$k=15$ &0.8600&0.8823&0.7798&$-$\\
SMKM&$k=15$  &\textbf{0.9700}&\textbf{0.9465}&\textbf{0.9379}&$-$\\
SMCL &$\alpha_{c}=\eta=0.005$ &0.9312&0.9306&0.8395&14\\
CC & $q=5$; $\gamma=9.5$&0.8314&0.8879&0.7311&\textbf{15}\\
MCKM & $q=1$; $\gamma=0.1$&0.9580&0.9326&0.9148&\textbf{15}\\

\midrule
&\multicolumn{5}{c}{\multirow{1}{*}{D6 ($n=5250, p=2, k=35$)}}
\\
\midrule
K-Means &$k=35$ &0.9068&0.9489&0.8652&$-$\\
SMKM &$k=35$&0.9890&0.9838&0.9775&$-$ \\
SMCL&$\alpha_{c}=\eta=0.0001$  &0.3058&0.6329&0.0366&7\\
CC&$q=5$; $\gamma=2.1$  &0.8992&0.9544&0.9819&\textbf{35}\\
MCKM &$q=1$;  $\gamma=0.05$ &\textbf{0.9893}&\textbf{0.9841}&\textbf{0.9783}&\textbf{35}\\
\bottomrule
\end{tabular}
\end{table}
\begin{table*}[h]
\centering
\caption{Running time in seconds on the  synthetic data sets. The first two algorithms require the number of clusters be given a priori while the last three algorithms do not.  The timings are averaged over 20 trials. The standard deviations are given after the means.}
\label{table3}
\begin{tabular}{r|rrrrrr}
\toprule
\multicolumn{1}{c|}{\multirow{1}{*}{Algorithms}}
&\multicolumn{1}{c}{\multirow{1}{*}{D1}}
&\multicolumn{1}{c}{\multirow{1}{*}{D2}}
&\multicolumn{1}{c}{\multirow{1}{*}{D3}}
&\multicolumn{1}{c}{\multirow{1}{*}{D4}}
&\multicolumn{1}{c}{\multirow{1}{*}{D5}}
&\multicolumn{1}{c}{\multirow{1}{*}{D6}}
\\
\midrule
K-Means&0.007$\pm$0.001&0.006$\pm$0.001&0.010$\pm$0.001&0.006$\pm$0.001&0.016$\pm$0.001&0.025$\pm$0.001 \\
SMKM&0.021$\pm$0.001&0.016$\pm$0.001&0.022$\pm$0.001&0.025$\pm$0.001&0.131$\pm$0.001&0.280$\pm$0.002 \\
\midrule
SMCL&10.485$\pm$3.121&5.081$\pm$1.930&8.564$\pm$0.026&12.354$\pm$1.218&14.495$\pm$0.115&27.590$\pm$3.448\\
CC&0.634$\pm$0.136&0.305$\pm$0.001&0.185$\pm$0.001&0.568$\pm$0.002&3.448$\pm$0.227&1.880$\pm$0.021 \\
MCKM&0.166$\pm$0.001&0.057$\pm$0.001&0.115$\pm$0.001&0.075$\pm$0.001&0.130$\pm$0.001&0.109$\pm$0.001\\
\bottomrule
\end{tabular}
\end{table*}

From the results, we have the following findings.
\begin{itemize}
  \item From Fig. \ref{fig7} and the corresponding Table \ref{table2}, MCKM is competent for the clustering of the arbitrary shape data sets, including  unbalanced data set,   non-convex data sets, and  convex data sets with a large cluster number.  Its clustering results are comparable to those of the state-of-the-art clustering algorithms, or even better, see, for example, the results of D3, D4, and D6.

    \item  In terms of getting the number of clusters, SMCL, CC and MCKM have the same performance when the true number of clusters is small. However, when the true number of clusters is relatively large, only CC and MCKM still work well by choosing a suitable parameter $\gamma$, because CM and hence MCKM inherits the advantage of convex clustering.
 In detail,  CM and CC solve the similar convex optimization model as in \eqref{eq_7} and \eqref{eq_15}. In summary, both CC and MCKM are outstanding in getting the number of clusters.

\item From Table \ref{table3}, we see that the running time of K-Means and SMKM is much less than that of SMCL and CC. Obviously, the estimation of the number of clusters is very time-consuming for the clustering algorithms.  Comparing the three algorithms without the cluster number given a priori, i.e., SMCL, CC, and MCKM, MCKM has significantly higher efficiency.  In detail, compared with CC, we note that $n$ samples are involved in the CC convex optimization model  \eqref{eq_7} whereas  $s^{*}$ prototypes are involved in the CM convex optimization model \eqref{eq_15}, where the number of the multi-prototypes by MPS $s^{*}\ll n$. Overall, the running time of MCKM is less than that of CC. This is consistent with  the complexity analysis in Subsections \ref{sec4-1} and \ref{sec4-2}. In particular, MCKM is even more efficient than SMKM on convex data sets with a large number of clusters, see D5 and D6. Therefore, both MPS and CM in MCKM are very efficient.
    \end{itemize}

\subsection{Experiments on Real-world Data Sets}
The second set of experiments are performed on real-world data sets selected from UCI Machine Learning Repository\footnote{\url{https://archive.ics.uci.edu/ml/index.php}}. The detailed information on the data sets, the clustering perforances and the hyper-parameters used are given in  Table \ref{table4}.   In MCKM, the constant $\rho= 1, 0.8, 1.6, 1, 2$ are  chosen empirically for MPS on  HTRU2, Iris, Wine, X8D5K, and Statlog  respectively. The running time of the algorithms are displayed in Table \ref{table5}, where the values are averaged over 20 trials.

From the results, we obtain the following findings.
\begin{itemize}
  \item In terms of the clustering results,  MCKM outperforms the other algorithms on almost all real-world data sets.
\item  In terms of evaluating the cluster number, CC and MCKM still perform better than the other algorithms.
\item In terms of the running time of the algorithms, although MCKM is not as efficient as the other algorithms where the cluster number are given a priori, i.e., K-Means and SMKM, it is the best among the algorithms that do not require the cluster number. In particular, the running time of MCKM is about 38$\%$ of  that of CC on average.
\end{itemize}

In conclusion,  based on the structure of spurious local minima of the K-Means problem, MCMK provides an explainable two-stage approach for recovering a better local minima, in which oversampling is performed  by MPS, and then the multi-prototypes are merged by CM. Moreover, the experiments on synthetic and real-world data sets show that MCKM is an outstanding and efficient clustering algorithm for recovering a better local minima.

\begin{table}[htb]
\centering
\caption{The evaluation of the clustering results of the different algorithms on the five real-world data sets. The best results are shown in boldface.}
\label{table4}
\begin{tabular}{r|l|cccc}
\toprule
\multicolumn{1}{c|}{\multirow{2}{*}{Algorithms}}
&\multicolumn{1}{c|}{\multirow{1}{*}{Parameter}}
&\multicolumn{1}{c}{\multirow{1}{*}{$\textrm{F}^{*}$}}
&\multicolumn{1}{c}{\multirow{1}{*}{NMI}}
&\multicolumn{1}{c}{\multirow{1}{*}{ARI}}
&\multicolumn{1}{c}{\multirow{1}{*}{$k^{*}$}}
\\
\cmidrule(r){2-6}
&\multicolumn{5}{c}{\multirow{1}{*}{HTRU2 ($n=17898, p=8, k=2$)}}
\\
\midrule
K-Means&$k=2$ &0.9121  &0.3396  &0.5318&$-$ \\
SMKM&$k=2$ &0.9121&0.3395&0.5317&$-$ \\
SMCL & $\alpha_{c}=\eta=0.001$&0.9050&0.2754&0.4753&\textbf{2}\\
CC&$q=8$; $\gamma=5$ &0.9228&0.3571&0.5484&\textbf{2} \\
    MCKM& $q=2$; $\gamma=2$ &\textbf{0.9637}&\textbf{0.5195}&\textbf{0.6784}&\textbf{2}\\
\midrule

&\multicolumn{5}{c}{\multirow{1}{*}{Iris ($n=150, p=4, k=3$)}}
\\
\midrule
K-Means&$k=3$  &0.8227   &0.6873   &0.6255&$-$\\
SMKM &$k=3$&0.8873&0.7392&0.7148&$-$ \\
SMCL& $\alpha_{c}=\eta=0.001$ &0.7778&0.7337&0.3705&2\\
CC&$q=5$; $\gamma=1$ &0.8955&\textbf{0.7701}&0.7312&\textbf{3} \\
MCKM& $q=2$;  $\gamma=0.5$ &\textbf{0.9008}&0.7578&\textbf{0.7430}&\textbf{3} \\

\midrule
&\multicolumn{5}{c}{\multirow{1}{*}{Wine ($n=178, p=13, k=3$)}}
\\
\midrule
K-Means&$k=3$ &0.9509&0.8356&0.8545&$-$ \\
SMKM &$k=3$&0.9495&0.8357&0.8484&$-$ \\
SMCL & $\alpha_{c}=\eta=0.001$&0.8179&0.6787&0.4804&2 \\
CC &$q=5$; $\gamma=1.5$ &0.9444&0.8252&0.8368&\textbf{3}\\
MCKM&$q=2$;  $\gamma=2$ &\textbf{0.9721}&\textbf{0.8926}&\textbf{0.9149}&\textbf{3} \\

\midrule
&\multicolumn{5}{c}{\multirow{1}{*}{X8D5K ($n=1000, p=8, k=5$)}}
\\
\midrule
K-Means&$k=5$ &0.9401&0.9587&0.9152&$-$\\
SMKM &$k=5$ &\textbf{1.0000}&\textbf{1.0000}&\textbf{1.0000}&$-$\\
SMCL & $\alpha_{c}=\eta=0.001$&\textbf{1.0000}&\textbf{1.0000}&\textbf{1.0000}&\textbf{5}\\
CC &$q=5$; $\gamma=1$ &\textbf{1.0000}&\textbf{1.0000}&\textbf{1.0000}&\textbf{5}\\
MCKM&$q=2$; $\gamma=1$&\textbf{1.0000}&\textbf{1.0000}&\textbf{1.0000}&\textbf{5} \\

\midrule
&\multicolumn{5}{c}{\multirow{1}{*}{Statlog ($n=4435, p=36, k=6$)}}
\\
\midrule
K-Means &$k=6$ &0.6911&0.6147&0.5305&$-$\\
SMKM &$k=6$&0.6910&0.6148&0.5307&$-$ \\
SMCL & $\alpha_{c}=\eta=0.001$&0.4674&0.2610&0.0227&2 \\
CC &$q=5$; $\gamma=13.15$&0.6955&0.5532&0.4294&\textbf{6}\\
MCKM&$q=2$; $\gamma=4$&\textbf{0.8279}&\textbf{0.6477}&\textbf{0.6175}&\textbf{6}\\
\bottomrule
\end{tabular}
\end{table}
\begin{table*}[htb]
\centering
\caption{Running time in seconds on the real-world data sets. The first two algorithms require the number of clusters be given a priori while the last three algorithms do not. The timings are averaged over 20 trials. The standard deviations are given after the means.}
\label{table5}
\begin{tabular}{r|rrrrr}
\toprule
\multicolumn{1}{c|}{\multirow{1}{*}{Algorithms}}
&\multicolumn{1}{c}{\multirow{1}{*}{HTRU2}}
&\multicolumn{1}{c}{\multirow{1}{*}{Iris}}
&\multicolumn{1}{c}{\multirow{1}{*}{Wine}}
&\multicolumn{1}{c}{\multirow{1}{*}{X8D5K}}
&\multicolumn{1}{c}{\multirow{1}{*}{Statlog}}
\\
\midrule
K-Means&0.035$\pm$0.002&0.004$\pm$0.001&0.003$\pm$0.001&0.005$\pm$0.001&0.019$\pm$0.001 \\
SMKM&0.085$\pm$0.001&0.020$\pm$0.001&0.016$\pm$0.001&0.029$\pm$0.001&0.049$\pm$0.001 \\
\midrule
SMCL&131.727$\pm$5.836&0.110$\pm$0.001&0.170$\pm$0.001&0.863$\pm$0.001&70.035$\pm$8.425\\
CC&202.409$\pm$1.582&0.180$\pm$0.001&0.116$\pm$0.001&0.190$\pm$0.001&55.175$\pm$0.667 \\
MCKM&0.402$\pm$0.004&0.110$\pm$0.001&0.107$\pm$0.001&0.068$\pm$0.001&0.279$\pm$0.001\\
\bottomrule
\end{tabular}
\end{table*}

\subsection{Performance of the Approximation to the Global Minima of K-Means Problem}
In the third set of experiments, we verify the approximation capability of MCKM on the global minima of K-Means problem. The corresponding K-Means errors of the chosen algorithms on all data sets are compared with the  optimal errors of the corresponding data sets. Referring to \cite{dasgupta2008hardness},  K-Means cost function can equivalently be rewritten as:\\
\begin{equation}\label{eq_other}
\begin{aligned}
 J_{\mathbf{X}}=&\sum_{i=1}^k \frac{1}{2|\C_{i}|} \sum_{j, j' \in \C_{i}} \|\x_{j}-\x_{j'}\|^2.
    \end{aligned}
\end{equation}

For a data set $\mathbf{X}$, the optimal error $J_{\mathbf{X}}^{*}$ can be calculated using the partition of the cluster from the true label. The corresponding error $J_{\mathbf{X}}$ for an algorithm can be calculated based on the partition of the cluster determined by the algorithm.  Table \ref{table6} shows the approximation capability of the algorithms by using   $|J_{\mathbf{X}}-J_{\mathbf{X}}^{*}|$. The best results are shown in boldface.

\begin{table*}[ht]
\centering
\caption{Performance of the approximation of the chosen algorithms on all data sets measured by  $|J_{\mathbf{X}}-J_{\mathbf{X}}^{*}|$. The best results are shown in boldface.}
\label{table6}
\begin{tabular}{r|rrrrrrrrrrr}
\toprule
\multicolumn{1}{c|}{\multirow{1}{*}{Data sets}}
&\multicolumn{1}{c}{\multirow{1}{*}{D1}}
&\multicolumn{1}{c}{\multirow{1}{*}{D2}}
&\multicolumn{1}{c}{\multirow{1}{*}{D3}}
&\multicolumn{1}{c}{\multirow{1}{*}{D4}}
&\multicolumn{1}{c}{\multirow{1}{*}{D5}}
&\multicolumn{1}{c}{\multirow{1}{*}{D6}}
&\multicolumn{1}{c}{\multirow{1}{*}{HTRU2}}
&\multicolumn{1}{c}{\multirow{1}{*}{Iris}}
&\multicolumn{1}{c}{\multirow{1}{*}{Wine}}
&\multicolumn{1}{c}{\multirow{1}{*}{X8D5K}}
&\multicolumn{1}{c}{\multirow{1}{*}{Statlog}}
\\
\midrule
$J_{\bf{X}}^{*}$ &60.5422&51.2649&54.8162&221.6124&8.0299&3.8056&778.7111&3.9087&24.9993&28.2419&974.7959\\
\midrule
K-Means&6.9862&5.5864&21.8904&60.6594&3.6511&1.7755&172.2072&5.0353&4.9308&5.2891&343.7963\\
SMKM&6.9862&5.5864&21.8904&60.6594&0.5653&0.0337&177.6846&0.4096&0.5192&\textbf{0}&348.9786 \\
SMCL&2.0607&0.2956&2.8287&0.0697&1.6028&93.1209&162.1329&2.1631&39.7478&\textbf{0}&1113.925 \\
CC&\textbf{1.4758}&\textbf{0.1841}&\textbf{0}&\textbf{0}&8.9249&2.5693&174.2149&0.3648&0.3584&\textbf{0}&359.1041 \\
MCKM&2.0926&0.2956&\textbf{0}&\textbf{0}&\textbf{0.1814}&\textbf{0.0313} &\textbf{41.3547}&\textbf{0.3037}&\textbf{0.3316}&\textbf{0}&\textbf{138.0041} \\
\bottomrule
\end{tabular}
\end{table*}

From Table \ref{table6}, we conclude that MCKM approximate better than the other algorithms in almost all data sets. This is attributed to MPS's better adaptation to  the arbitrary shape data sets and CM's superior merging mechanism for the multi-prototypes.

\section{Conclusion}\label{sec6}
In this paper, multi-prototypes convex merging based K-Means clustering algorithm (MCKM) is proposed to recover a better local minima of K-Means problem without the cluster number given first. In the proposed algorithm,  a multi-prototypes sampling (MPS) is used to select the appropriate number of multi-prototypes with better adaptation to data distribution. A theoretical proof is given to guarantee that  MPS can achieve a constant factor approximation to the global minima of K-Means problem. Then, a convex merging (CM) technique is developed to formulate the merging of the multi-prototypes task as a convex optimization problem. Specifically, CM obtains the optimal merging and estimate the correct cluster number.
 Experimental results have verified MCKM's effectiveness and efficiency on synthetic and real-world data sets.  For future work, MPS  could be explored to achieve a better approximate of the upper bounds. Another interesting possibility is to implement  the adaptive selection technique of the regularization  parameter $\gamma$ in CM.

\bibliographystyle{IEEEtran}
 \bibliography{Reference}
 
 \clearpage
\begin{appendices}

\section{Proof of \textbf{Theorem} \ref{theorem3}}\label{sec7-1}
 Here, we prove that  the multi-prototypes obtained by MPS can achieve a constant factor approximation to the optimal cost of K-Means problem.

\begin{proof}
Let $\vv(\x)$ be the prototype to which $\x$ belongs and $\vv^{*}(\x)$ be the optimal prototype to which $\x$ belongs.
Assume that MPS has chosen $s$ samples, $1\le s\le n$, as the prototypes $\V$, and we continue to choose the next prototype $\x^{(s+1)}$ from $\mathbf{X}$. The probability of being selected is precisely $D(\x^{(s+1)})^2/\sum_{\x \in \mathbf{X}} D(\x)^2$.
After adding the prototype $\x^{(s+1)}$, any sample $\x$ will contribute $\min(D(\x), \|\x-\x^{(s+1)}\|)^2$ to the objective function. Therefore, \\
\begin{align*}
E[J_{\mathbf{X}}]=\sum_{\x^{(s+1)} \in \mathbf{X}} \frac{D(\x^{(s+1)})^2}{\sum_{\x \in \mathbf{X}} D(\x)^2}\sum_{\x \in \mathbf{X}} \min(D(\x), \|\x-\x^{(s+1)}\|)^2.
\end{align*}

According to the termination condition of MPS, since $\x^{(s+1)}$  is selected, we have:
\begin{align*}
\frac{R(s)-R(s+1)}{R(s)} \geq \varepsilon \Leftrightarrow (1-\varepsilon) R(s) \geq R(s+1).
\end{align*}
Based on \eqref{eq_9}, we have $E[J_{\mathbf{X}}]  \leq (1-\varepsilon)\sum_{\x \in \mathbf{X}} D(\x)^2$. By the power-mean inequality $\|\x-\vv(\x)\|^2 \leq 2\|\x-\vv^{*}(\x)\|^2+2\|\vv^{*}(\x)-\vv(\x)\|^2$, we have
\begin{align*}
E[J_{\mathbf{X}}] & \leq (1-\varepsilon)\sum_{\x \in \mathbf{X}} D(\x)^2, \\
& \leq 2(1-\varepsilon)\sum_{\x \in \mathbf{X}} (\|\x-\vv^{*}(\x)\|^2+\|\vv^{*}(\x)-\vv(\x)\|^2),\\
&=2(1-\varepsilon)({J_{\mathbf{X}}}^{opt}+\sum_{\x \in \mathbf{X}} \|\vv^{*}(\x)-\vv(\x)\|^2).
\end{align*}

Assume that MPS continues to run and terminates after the algorithm has sampled $s^{*}$ prototypes. Because MPS adopts $D^2$ sampling method, we have for any $\x$, $D(\x^{(s^{*}+1)})^2 \geq D(\x)^2$. Accordingly, we have:
\begin{align*}
D(\x^{(s^{*}+1)}) \geq D(\x) \geq  \left|\|\vv(\x)-\vv^{*}(\x)\|-\|\x-\vv^{*}(\x)\|\right|.
\end{align*}
Then, define $\mathbf{X}_{a}=\{\x| \|\vv(\x)-\vv^{*}(\x)\| \geq \|\x-\vv^{*}(\x)\|\}$ and $\mathbf{X}_{b}=\{\mathbf{X} \setminus \mathbf{X}_{a}\}$. Let $n_{a}=|\mathbf{X}_{a}|$ represents the cardinality of  the set $\mathbf{X}_{a}$. Combining the above derivations, we have:
\begin{align*}
E[J_{\mathbf{X}}] &\leq 2(1-\varepsilon)({J_{\mathbf{X}}}^{opt}+\sum_{\x \in \mathbf{X}} \|\vv^{*}(\x)-\vv(\x)\|^2)\\
&\leq 2(1-\varepsilon)({J_{\mathbf{X}}}^{opt}+\sum_{\x \in \mathbf{X}_{b}} \|\x-\vv^{*}(\x)\|^2\\
&+\sum_{\x \in \mathbf{X}_{a}}\left[\|\x-\vv^{*}(\x)\|+D(\x^{(s^{*}+1)})\right]^2).
\end{align*}
Since MPS terminates at $s^*$ steps, we have:
\begin{align*}
\frac{R(s^{*})-R(s^{*}+1)}{R(s^{*})} \leq \varepsilon \Leftrightarrow (1-\varepsilon) R(s^{*}) \leq R(s^{*}+1)
\end{align*}
which is equivalent to:
\begin{align*}
\sum_{\x \in \mathbf{X}} \{D(\x)^2-\min(D(\x), \|\x-\x^{(s^{*}+1)}\|)^2\} \leq \varepsilon \sum_{\x \in \mathbf{X}} D(\x)^2.
\end{align*}

Here, $\mathbf{X}$ is divided into three parts according to the following rules: 1) $\x \in \mathbf{X}_{1}$, if $\min(D(\x), \|\x-\x^{(s^{*}+1)}\|)^2=D(\x)^2$; 2) $\x \in \mathbf{X}_{2}$, if $\min(D(\x), \|\x-\x^{(s^{*}+1)}\|)^2=\|\x-\x^{(s^{*}+1)}\|)^2$; 3) $\mathbf{X}_{3}=\{\x^{(s^{*}+1)}\}$. Evidentially,
\begin{align*}
&\sum_{\x \in \mathbf{X}} \{D(\x)^2-\min(D(\x), \|\x-\x^{(s^{*}+1)}\|)^2\}\\
&=0+\sum_{\x \in \mathbf{X_{2}}}\{D(\x)^2-\|\x-\x^{(s^{*}+1)}\|^2\}+D(\x^{(s^{*}+1)})^2.
\end{align*}
Therefore, we have:
\begin{align*}
D(\x^{(s^{*}+1)})^2 \leq \varepsilon \sum_{\x \in \mathbf{X}} D(\x)^2.
\end{align*}
The last step is summarized as follows:
\begin{align*}
E[J_{\mathbf{X}}] \leq 2(1-\varepsilon)(3{J_{\mathbf{X}}}^{opt}+2 \varepsilon n_{a} \sum_{\x \in \mathbf{X}} D(\x)^2).
\end{align*}

The proof process is complete.
\label{proof1}
\end{proof}

\section{The ADMM for solving CM.}\label{sec7-2}

The objective of CM \eqref{eq_15} is recast as the equivalent constrained problem:\\
\begin{equation}\label{eq_16}
\begin{aligned}
 \min_{\bm{\mu}_{1},..., \bm{\mu}_{s^{*}} \in \Real^{p}} & \frac{1}{2} \sum_{i=1}^{s^{*}} \|\bm{\mu}_{i}-\vv_{i}\|^2+\gamma \sum_{l \in E} w_{l}\|\bm{y}_{l}\|_{2},\\
 s.t. & \quad \bm{\mu}_{l_{1}}-\bm{\mu}_{l_{2}}-\bm{y}_{l}=0,
    \end{aligned}
\end{equation}
where $l=(l_{1},l_{2})$ with $l_{1} < l_{2}$, and $\bm{y}_{l}=\bm{\mu}_{l_{1}}-\bm{\mu}_{l_{2}}$ is introduced to simplify the penalty terms.
For the constrained optimization problem \eqref{eq_16}, the augmented Lagrangian is given by:
\begin{equation}\label{eq_17}
\begin{aligned}
\mathcal{L}_{\nu}&(\bm{\mu}, \bm{y}, \bm{\lambda})=\frac{1}{2} \sum_{i=1}^{s^{*}} \|\bm{\mu}_{i}-\vv_{i}\|^2+\gamma \sum_{l \in E} w_{l}\|\bm{y}_{l}\|_{2}\\
&+\sum_{l \in E}\langle \bm{\lambda}_{l}, \bm{y}_{l}-\bm{\mu}_{l_{1}}+\bm{\mu}_{l_{2}}\rangle+ \frac{\nu}{2} \sum_{l \in E}\|\bm{y}_{l}-\bm{\mu}_{l_{1}}+\bm{\mu}_{l_{2}}\|_{2}^{2}.
    \end{aligned}
\end{equation}
ADMM minimizes the augmented Lagrangian by the following iterative process:
\begin{equation}\label{eq_18}
\begin{aligned}
\bm{\mu}^{t+1}&=\arg \min_{\bm{\mu}} \mathcal{L}_{\nu}(\bm{\mu}, \bm{y}^{t}, \bm{\lambda}^{t});\\
\bm{y}^{t+1}&=\arg \min_{\bm{y}} \mathcal{L}_{\nu}(\bm{\mu}^{t+1}, \bm{y}, \bm{\lambda}^{t});\\
\bm{\lambda}_{l}^{t+1}&=\bm{\lambda}_{l}^{t}+ \nu (\bm{y}_{l}^{t+1}-\bm{\mu}_{l_{1}}^{t+1}+\bm{\mu}_{l_{2}}^{t+1}), l \in E,
\end{aligned}
\end{equation}
where $t$ is the iteration number. According to the above analysis and derivation \cite{Splitting2015}, ADMM for solving \eqref{eq_15} is summarized in Algorithm \ref{alg5}.
\newpage

\begin{algorithm}[ht]
 \caption{ADMM for Solving \eqref{eq_15}}
 \label{alg5}
 \begin{algorithmic}[1]
        \REQUIRE The multi-prototypes $\V_{\textrm{MPS}}$, the number of the multi-prototypes $s^{*}$, the number of neighboring samples $q$, a positive constant $\kappa$, the tuning parameter $\gamma$,  the termination $\eta$, $\bm{\lambda}^{0}$, and new variables $\bm{y}^{0}$;
  \ENSURE The optimal solutions, $\bm{\mu}^{*}_{1},..., \bm{\mu}^{*}_{s^{*}}$.
        \STATE set the iteration number $t=0$
        \STATE $\bar{\V}_{\textrm{MPS}}$ is the average column of $\V_{\textrm{MPS}}$;
        \STATE compute $E$ and $W$ based on  Eq. \eqref{eq 16};
        \STATE $\sigma_{l}=\frac{\gamma w_{l}}{\nu}, l \in E$;
        \REPEAT
        \STATE compute $\bm{Z}$ with $\bm{z}_{i}=\vv_{i}+\sum_{l_{1}=i}(\bm{\lambda}_{l}^{m}+\nu \bm{y}_{l}^{m})-\sum_{l_{2}=i}(\bm{\lambda}_{l}^{m}+\nu \bm{y}_{l}^{m}), i=1,2,..., s^{*}$;
        \STATE update $\bm{\mu}^{t+1}=\frac{1}{1+c^{*}\nu}\bm{Z}+\frac{c^{*}\nu}{1+c^{*}\nu}\bar{\V}_{\textrm{MPS}}$;
        \STATE update $\bm{y}^{t+1}$ with $\bm{y}_{l}^{t+1}=\textrm{prox}_{\sigma_{l}\|\cdot\|}(\bm{\mu}_{l_{1}}^{t+1}-\bm{\mu}_{l_{2}}^{t+1}-\nu^{-1}\bm{\lambda}_{l}^{t}), l \in E$;
        \STATE update $\bm{\lambda}^{t+1}$ with $\bm{\lambda}_{l}^{t+1}=\bm{\lambda}_{l}^{t}+\nu (\bm{y}_{l}^{t+1}-\bm{\mu}_{l_{1}}^{t+1}+\bm{\mu}_{l_{2}}^{t+1}), l \in E$;
        \STATE $t=t+1$;
        \UNTIL{Stopping criterion is met}
        \STATE Obtain the optimal solutions, $\bm{\mu}^{*}_{1},..., \bm{\mu}^{*}_{s^{*}}$.
 \end{algorithmic}
\end{algorithm}
    \end{appendices}

\end{document}